\documentclass[10pt,twocolumn,letterpaper]{article}

\usepackage{cvpr}
\usepackage{times}
\usepackage{epsfig}
\usepackage{graphicx}
\usepackage{amsmath}
\usepackage{amssymb}

\usepackage[font=small]{subcaption}
\usepackage[font=small]{caption}
\usepackage{booktabs}
\usepackage{authblk}
\usepackage{url}
\usepackage[titletoc]{appendix}

\pdfpageattr {/Group << /S /Transparency /I true /CS /DeviceRGB>>}

\usepackage[breaklinks=true,bookmarks=false]{hyperref}

\cvprfinalcopy 

\def\cvprPaperID{6227} 

\ifcvprfinal\fi

\begin{document}

\title{Learning to Shadow Hand-drawn Sketches}


\author[1]{Qingyuan Zheng\thanks{Equal contribution.}}

\newcommand\CoAuthorMark{\footnotemark[\arabic{footnote}]}
\author[2]{Zhuoru Li\protect\CoAuthorMark}
\author[1]{Adam Bargteil}


\affil[1]{University of Maryland, Baltimore County}
\affil[2]{Project HAT}
\affil[ ]{\tt \small \{qing3, adamb\}@umbc.edu, hatsuame@gmail.com}

\maketitle
\thispagestyle{empty}

\begin{abstract}
  We present a fully automatic method to generate detailed and accurate artistic shadows from pairs of line drawing sketches and lighting directions. We also contribute a new dataset of one thousand examples of pairs of line drawings and shadows that are tagged with lighting directions. Remarkably, the generated shadows quickly communicate the underlying 3D structure of the sketched scene. Consequently, the shadows generated by our approach can be used directly or as an excellent starting point for artists. We demonstrate that the deep learning network we propose takes a hand-drawn sketch, builds a 3D model in latent space, and renders the resulting shadows. The generated shadows respect the hand-drawn lines and underlying 3D space and contain sophisticated and accurate details, such as self-shadowing effects.  Moreover, the generated shadows contain artistic effects, such as rim lighting or halos appearing from back lighting, that would be achievable with traditional 3D rendering methods.
\end{abstract}

\section{Introduction}

Shadows are an essential element in both traditional and digital painting.~Across artistic media and formats, most paintings are first sketched with lines and shadows before applying color.~In both the Impressionism and Neo-classicism era, artists would paint oil paintings after they rapidly drew shadowed sketches of their subjects. They recorded what they saw and expressed their vision in sketches and shadows and used these as direct references for their paintings \cite{arnheim1965art}. 

\begin{figure}
	\centering
	\includegraphics[width=\linewidth]{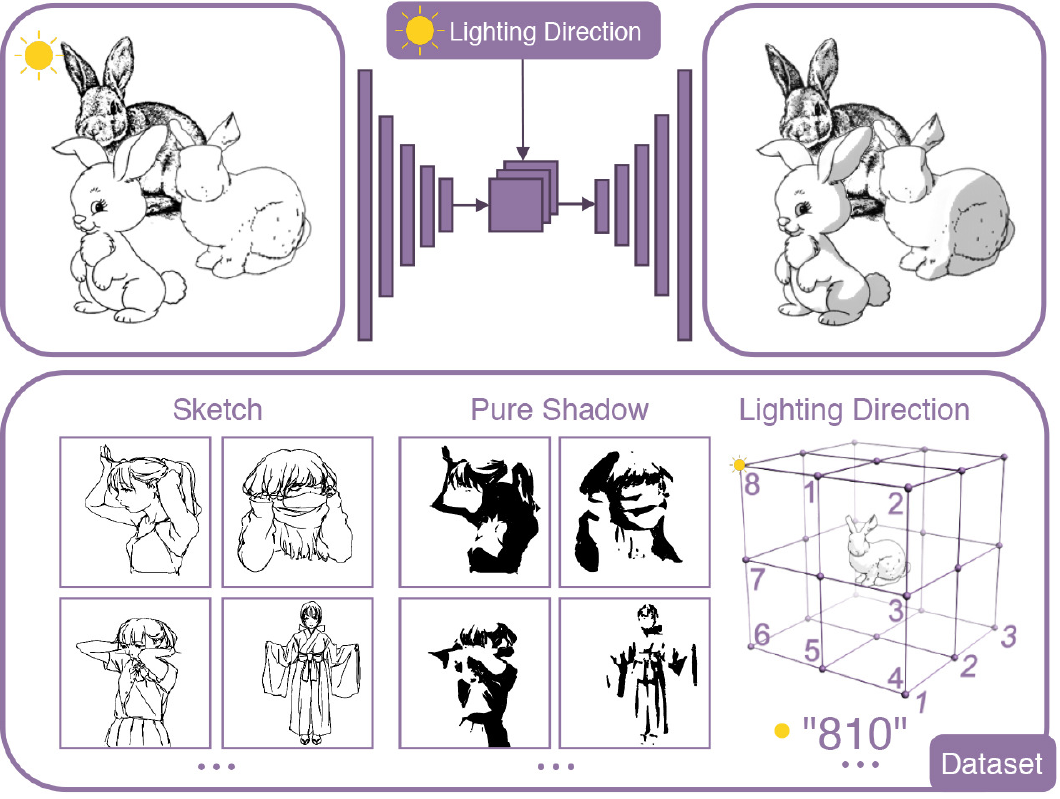}
	\captionof{figure}{Top: our shadowing system takes in a line drawing and a lighting direction label, and outputs the shadow. Bottom: our training set includes triplets of hand-drawn sketches, shadows, and lighting directions. Pairs of sketches and shadow images are taken from artists' websites and manually tagged with lighting directions with the help of professional artists. The cube shows how we denote the 26 lighting directions (see Section~\ref{section:data}). \textcopyright Toshi, Clement Sauve}
	\label{fig:flowchart}
\end{figure}

In the modern painting era, particularly for digital illustration and cel animation, shadows play an important role in depicting objects' shapes and the relationships between 2D lines and 3D space, thereby affecting the audience's recognition of the scene as whole. Illustration is a time-consuming process; illustrators frequently spend several hours drawing an appealing picture, iteratively adjusting the form and structure of the characters many times. In addition to this work, the illustrators also need to iteratively adjust and refine the shadows, either after completing the sketch or while iterating the sketching process. Drawing shadows is particularly challenging for 2D sketches that cannot be observed in the real world, because there is no 3D reference model to reason about; only the artist's imagination. In principal, the more details the structural lines contain, the more difficult it is to draw the resulting shadows. Hence adjusting the shadows can be time consuming, especially for inexperienced illustrators.

In this paper, we describe a real-time method to generate plausible shadows from an input sketch and specified lighting direction.  These shadows can be used directly, or if higher quality is desired can be used as a starting point for the artists to modify. Notably, our approach does not generate shadowed sketches directly; instead it generates a separate image of the shadow that may be composited with the sketch.  This feature is important as the artist can load the sketch and the shadow into separate image layers and edit them independently.  

Our work uses the deep learning methodology to learn a non-linear function which ``understands'' the 3D spatial relationships implied by a 2D sketch and render the binary shadows (Figure~\ref{fig:flowchart} top). The raw output from our neural network is binary shadows, which may be modified by artists in a separate layer independent of line drawings. There is no additional post-processing and the images in our paper are simple composites of the raw network outputs and the input line drawings. If soft shadows are desired, artists may use the second intermediate output from our network (Figure~\ref{fig:gans} $s_{2}$). Our network also produces consistent shadows from continuously varying lighting directions (Section~\ref{section:artistic}), even though we train from a discrete set of lighting directions. 

Given a line drawing and a lighting direction, our model automatically generates an image where the line drawing is enhanced with detailed and accurate hard shadows; no additional user input is required. We focus on 2D animation style images (\eg Japanese comic, Inker~\cite{inker}) and the training data is composed of artistic hand-drawn line drawing in the shape of animation characters, mecha, and mechanical objects. We also demonstrate that our model generalizes to line drawing of different objects such as buildings, clothes, and animals.

The term ``artistic shadow'' in our work refers to binary shadows that largely obey physics but also have artistic features such as less shadowing of characters' faces and rim lighting when characters are back lit.

The main contributions of our work: 
\begin{itemize}
	\item We created a new dataset that contains 1,160 cases of hand-drawn line drawings and shadows tagged with lighting directions.
	\item We propose a network that ``understands'' the structure and 3D spatial relationships implied by line drawings and produces highly-detailed and accurate shadows.
	\item An end-to-end application that can generate binary or soft shadows from arbitrary lighting directions given a 2D line drawing and designated lighting direction. 
\end{itemize}

In Section~\ref{section:3}, we will decribe the design of our generative and discriminator networks, and our loss functions. In Section~\ref{section:4}, we compare our results quantitatively and qualitatively to baseline network architectures  
pix2pix \cite{isola2017image} and U-net \cite{ronneberger2015u}. We also compare to the related approaches Sketch2Normal \cite{su2018interactive} 
and DeepNormal \cite{hudon2018deep} applied to our shadow generation problem. Our comparisons include a small user study to assess the perceptual accuracy of our approach.  Finally, we demonstrate the necessity of each part of our proposed network through an ablation study and metrics analysis. \footnote{Project page is at \url{https://cal.cs.umbc.edu/Papers/Zheng-2020-Shade/}.}

\section{Related Work}

\textbf{Non-photorealistic rendering in Computer Graphics.}~The previous work on stylized shadows \cite{petrovic2000shadows, decoro2007stylized} for cel animation highlights that shadows play an important role in human perception of cel animation. In particular, shadows provide a sense of depth to the various layers of character, foreground, and background. Lumo \cite{johnston2002lumo} approximates surface normals directly from line drawings for cel animation to incorprate subtle environmental illumination. Todo \etal \cite{todo2013lit, todo2007locally} proposed a method to generate artistic shadows in 3D scenes that mimics the aesthetics of Japanese 2D animation. Ink-and-Ray \cite{sykora2014ink} combined a hand-drawn character with a small set of simple annotations to generate bas-relief sculptures of stylized shadows. Recently, Hudon \etal \cite{hudon20182d} proposed a semi-automatic method of cel shading that produces binary shadows based on hand-drawn objects without 3D reconstruction. 

\textbf{Image translation and colorization.}~In recent years, the research on Generative Adversarial Networks (GANs) \cite{goodfellow2014generative, mirza2014conditional} in image translation \cite{isola2017image} has generated impressive synthetic images that were perceived to be the same as the originals. Pix2pix \cite{isola2017image} deployed the U-net \cite{ronneberger2015u} architecture in their Generator network and demonstrated that for the application of image translation U-net's performance is improved when skip connections are included. CycleGAN \cite{zhu2017unpaired} introduced a method to learn the mapping from an input image to a stylized output image in the absence of paired examples. Reearch on colorizing realistic gray scale images \cite{cheng2015deep, zhang2016colorful, IizukaSIGGRAPH2016, zhang2017real} demonstrated the feasibility of colorizing images using GANs and U-net \cite{ronneberger2015u} architectures. 

\textbf{Deep learning in line drawings.}~Researcher that considers line drawings include line drawing colorization \cite{yonetsuji2017paintschainer, kim2019tag2pix, zhang2018two, furusawa2017comicolorization, frans2017outline}, sketch simplification \cite{simo2016learning, simo2018mastering}, smart inker \cite{simo2018real}, line extraction \cite{li2017deep}, line stylization \cite{li2019im2pencil} and computing normal maps from sketches \cite{su2018interactive, hudon2018deep}. Tag2Pix \cite{kim2019tag2pix} seeks to use GANs that concatenate Squeeze and Excitation \cite{hu2018senet} to colorize line drawing.  Sketch simplification \cite{simo2016learning, simo2018mastering} cleans up draft sketches, through such operations as removing dual lines and connecting intermittent lines. Smart inker \cite{simo2018real} improves on sketch simplification by including additional user input. Users can draw strokes indicating where they would like to add or erase lines, then the neural network will output a simplified sketch in real-time.  Line extraction \cite{li2017deep} extracts pure lines from manga (comics) and demonstrates that simple downscaling and upscaling residual blocks with skip connections have superior performance. Kalogerakis \etal \cite{kalogerakis2012learning} proposed a machine learning method to create hatch-shading style illustrations. Li \etal \cite{li2019im2pencil} proposed a two-branch deep learning model to transform the line drawings and photo to pencil drawings.

\textbf{Relighting.}~Deep learning has also been applied to relighting realistic scenes. Xu \etal \cite{xu2018deep} proposed a method for relighting from an arbitrary directional light given images from five different directional light sources. Sun \etal \cite{sun2019single} proposed a method for relighting portraits given a single input, such as a selfie.  The training datasets are captured by a multi-camera rig.  This work differs from ours in that they focus on relighting realistic images 
while we focus on artistic shadowing of hand-drawn sketches.

\textbf{Line drawings to normal maps.}~Sketch2normal \cite{su2018interactive} and DeepNormal \cite{hudon2018deep} use deep learning to compute normal maps from line drawings. Their training datasets are rendered from 3D models with realistic rendering. Sketch2Normal trains on line drawings of four-legged animals with some annotations. DeepNormal takes as input line drawings with a mask for the object. They solve a different, arguably harder, problem.  However, the computed normal maps can be used to render shadows and we compare this approach to our direct shadow computation in Section~\ref{section:4}. Given color input images, Gao and colleagues~\cite{gao2018illumination} predict normal maps and then generate shadows.

\section{Learning Where to Draw Shadows}
\label{section:3}
In this section we describe our data preparation, our representation of the lighting directions, the design of our generator and discriminator networks, and our loss functions. 

\subsection{Data Preparation}
\label{section:data}
We collect our (sketch, shadow) pairs from website posts by artists. With help from professional artists, each (sketch, shadow) pair is manually tagged with a lighting direction. After pre-processing the sketches with thresholding and morphological anti-aliasing, the line drawings are normalized to obtain a consistent line width of 0.3 px in {\em cairosvg} standard \cite{cairosvg}. To standardize the hand-drawn sketch to the same line width, we use a small deep learning model similar to smart inker \cite{simo2018real} to pre-process input data. Our dataset contains 1,160 cases of hand-drawn line drawings. Each line drawing matches one specific hand-drawn shadow as ground truth and one lighting direction. 

In contrast to 3D computer animation, which contains many light sources and realistic light transport, 2D animation tends to have a single lighting direction and include some non-physical shadows in a scene. 

We observed that artists tend to choose from a relatively small set of specific lighting directions, especially in comics and 2D animation. For this reason, we define 26 lighting directions formed by the $2\times2$ cube in Figure~\ref{fig:flowchart}. We found that it was intuitive to allow users to choose from eight lighting directions clockwise around the 2D object and one of three depths (in-front, in-plane, and behind) to specify the light source.  We also allow the user to choose two special locations: directly in front and directly behind.  This results in $8\times 3+2=26$ lighting directions.  The user specifies the light position with a three-digit string.  The first digit corresponds to the lighting direction (1-8), the second to the plane (1-3), and the third is '0' except for the special directions, which are ``001'' (in-front) and ``002'' (behind).

While users found this numbering scheme intuitive, we obtained better training results by first converting these strings to 26 integer triples on the cube from $[-1,1]^3$ ($(0,0,0)$ is not valid as that is the location of the object). 
For example, ``610'' is mapped to $(-1, -1, -1)$, ``230'' is mapped to $(1, 1, 1)$, and ``210'' is mapped to $(1, 1, -1)$.

\begin{figure*}[t]
	\centering
	\includegraphics[width=\textwidth]{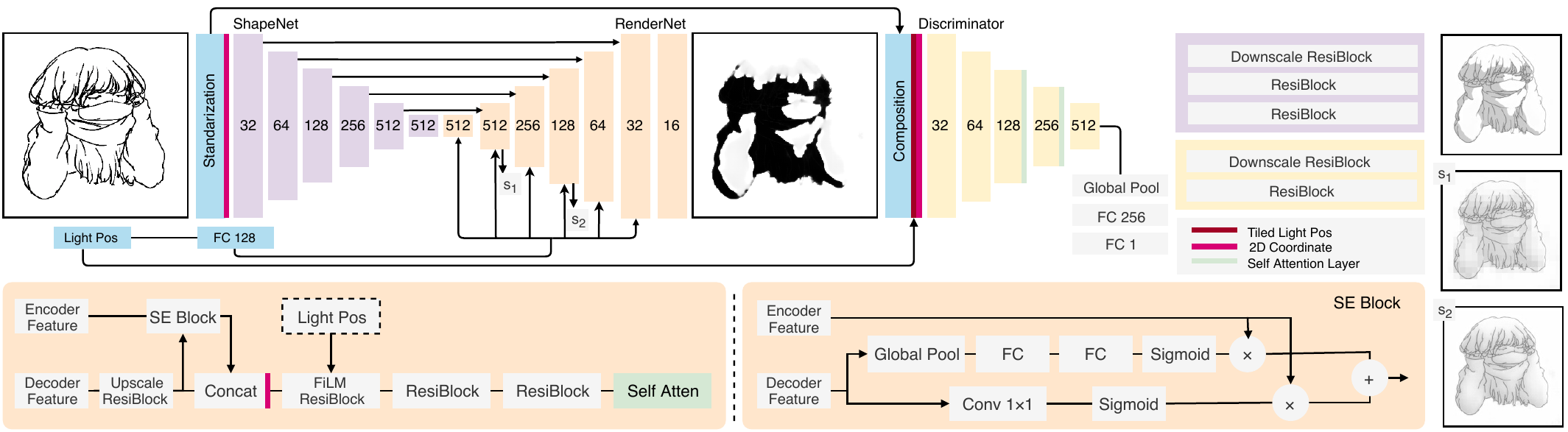}
	\caption{Our GANs architecture. The line drawings are standardized first (same as in Section~\ref{section:data}) before being inputted into the {\em ShapeNet}. Lighting directions are repeatedly added into the FiLM residual block in each stage in {\em RenderNet}. $s_{1}$ and $s_{2}$ are the up-sampled intermediate outputs from the second and the forth stage in {\em RenderNet}. In the training process, the line drawings and pure shadows are inverted from black-on-white to white-on-black. More details are in supplementary material.}
	\label{fig:gans}
\end{figure*}

\subsection{Network Architecture}
\label{section:network}
Our generator incorporates the following modules: residual blocks \cite{He_2016_CVPR} \cite{He2016IdentityMI}, FiLM \cite{perez2018film} residual blocks, and Squeeze-and-Excitation (SE) blocks \cite{hu2018senet}. The general architecture of our generator follows the architecture of U-net with skip connections \cite{ronneberger2015u, isola2017image}. 
Our Discriminator uses residual blocks. Details are shown in Figure~\ref{fig:gans}. 

\subsubsection{Generative Network}

~~~~We propose a novel non-linear model with two parts - {\em ShapeNet}, which encodes the underlying 3D structure from 2D sketches, and {\em RenderNet}, which renders artistic shadows based on the encoded structure.

\textit{ShapeNet} encodes a line drawing of an object into a high dimensional latent space and represents the object's 3D geometric information. We concatenate 2D coordinate channels \cite{liu2018intriguing} to the line drawings to assist \textit{ShapeNet} in encoding 3D spatial information. 
 
 \textit{RenderNet} performs reasoning about 3D shadows. Starting from the bottle neck, we input the embedded lighting direction using the normalization method from FiLM residual blocks \cite{perez2018film}. The model then starts to learn the relationship between the lighting direction and the various high dimensional features. We repeatedly add the lighting direction into each stage of the \textit{RenderNet} to enhance the reasoning of decoding. In the bottom of each stage in \textit{RenderNet}, a Self-attention \cite{zhang2018self} layer complements the connection of holistic features.

The shadowing problem involves holistic visual reasoning because shadows can be cast by distant geometry. For this reason we deploy Self-attention layers \cite{zhang2018self} and FiLM residual blocks \cite{perez2018film} to enhance the visual reasoning; networks that consist of only residual blocks have limited receptive fields and are ill-suited to holistic visual reasoning. The SE \cite{hu2018senet} blocks filter out unnecessary features imported from the skipped encoder output. 

We also extract two supervision intermediate outputs, $s_1$ and $s_2$, to facilitate backpropagation. Early stages of our \textit{RenderNet} generate continuous, soft shadow images.  In the final stage, the network transforms these images to binary shadows. The quality of the soft shadows in the intermediate outputs, $s_{1}$ and $s_{2}$, is shown in Figure~\ref{fig:gans}. We note again that our output does not require any post processing to generate binary shadows; the images in this paper result directly from compositing the output our generator with the input sketch.

\subsubsection{Discriminator Network}
~~~~The basic modules of our discriminator include downscaling residual blocks and residual blocks. Since many local features of different shadows are similar to one another, we deploy Self-attention layers to make our discriminator sensitive to the distant features. In Figure~\ref{fig:gans}, the last of the discriminator consists of global average pooling, dropout with $0.3$ probabilities, and a fully connected layer with $256$ filters. Because generating shadows is more difficult than discriminating between fake and real shadows, a simple discriminator is sufficient and simplifies training. 

\begin{figure*}
	\centering
    \includegraphics[width=\linewidth]{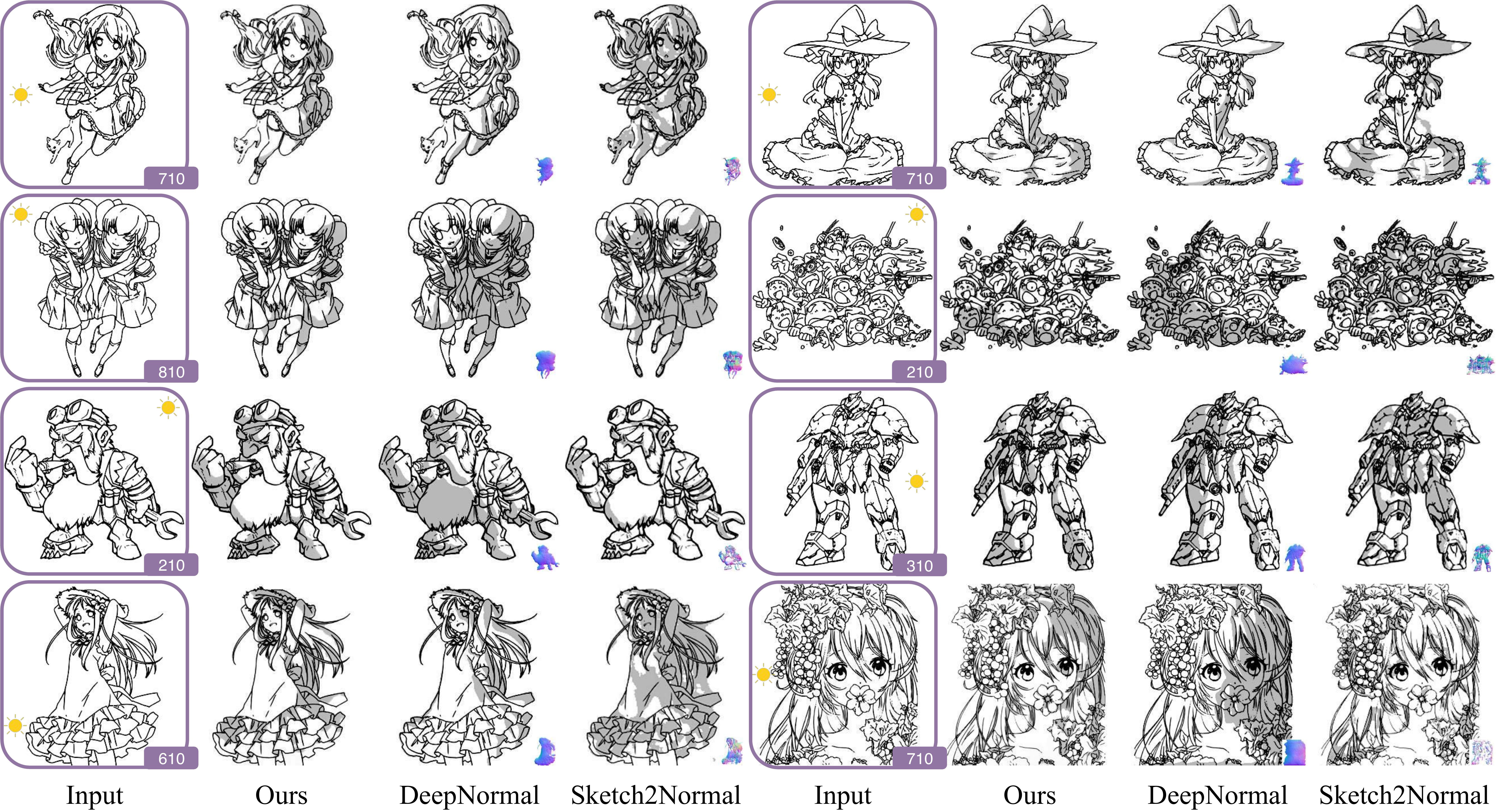}
	\caption{Shadows for lighting depth ``1'' - in front of the plane (front lighting), compared with previous work DeepNormal \cite{hudon2018deep} and Sketch2Normal \cite{su2018interactive}. The little sun denotes the lighting direction. \textcopyright Derori-san, Imomushi-san, Eric ou}
	\label{fig:compare1}
\end{figure*}

\subsection{Loss Function}
The adversarial loss of our Generative Adversarial Network can be expressed as 
\begin{equation}
\label{eq:advloss}
\begin{aligned}
\mathcal{L}_{cGAN}\left(G,D\right) = \mathbb{E}_{x, y, z}\left[logD\left(C\left(x,y\right), z\right)\right]\\
+\mathbb{E}_{x, z}\left[log\left(1-D\left(C\left(x, G\left(x, z\right)\right), z\right)\right)\right],
\end{aligned}
\end{equation}
where $x$ is the sketch, $y$ is the ground truth shadow, and $z$ is the lighting direction. $C(\cdot)$ is a function that composite the ground truth shadow and the input sketch as a ``real'' image, and composite the generated shadow and the input sketch as a ``fake'' image.

The generator $G$ aims to minimize the loss value, and the discriminator $D$ aims to maximize the loss value. For the loss value of our generator network, we add MSE losses of the two deep supervised outputs, which are the intermediate outputs of the first and third stage in the decoder, to the loss of the generator's final output. 

The three losses of the generator network can be expressed as
\begin{equation}
\label{eq:outputloss}
\begin{aligned}
  \mathcal{L}_{output}(G) = \mathbb{E}_{x, y, z}\left[\left\lVert y-G(x,z) \right\lVert_{2}^{2}\right] + \xi \cdot \mathrm{TV}\left(G(x,z)\right),
\end{aligned}
\end{equation}
where $\mathcal{L}_{output}$ is the loss between generated shadow and the ground truth.~$\mathcal{L}_{output}$ consists of a total variation (TV) regularizer and an MSE loss.~The~TV~regularizer, weighted by $\xi$, encourages smooth details around the boundaries of shadows. We set $\xi$ to $2\times 10^{-6}$, a $5\times$ smaller value than the total number of pixels in the input sketch. We will show how the value of $\xi$ affects the final output in the ablation study. The deep supervised outputs are upsampled and their losses are computed as by MSE loss from ground truth,

\begin{equation}
\label{eq:s1loss}
\begin{aligned}
\mathcal{L}_{s_{i}}\left(G\right) = \mathbb{E}_{x, y, z}\left[\left\lVert y-G_{s_{i}}(x,z) \right\lVert_{2}^{2}\right],~ i=1,2.
\end{aligned}
\end{equation}

Final objective is the sum of $\mathcal{L}_{output}$, $\mathcal{L}_{s_{1}}$, $\mathcal{L}_{s_{2}}$, and the $\mathcal{L}_{cGAN}$,
\begin{equation}
\label{eq:totalloss}
\begin{aligned}
G^{*} = arg\min_{G} \max_{D} \lambda_{1} \mathcal{L}_{cGAN}\left(G,D\right)\\
+\lambda_{2} \mathcal{L}_{output}\left(G\right) + \lambda_{3}\mathcal{L}_{s_{1}}\left(G\right) + \lambda_{4}\mathcal{L}_{s_{2}}\left(G\right).
\end{aligned}
\end{equation}
In our experiments, the four losses are weighted by $\lambda_{1}=0.4$, $\lambda_{2}=0.5$, $\lambda_{3}=0.2$, and $\lambda_{4}=0.2$.

\section{Experiments and Evaluation}
\label{section:4}

In this section, we evaluate the performance of our shadowing model.  In particular, we discuss implementation details, provide comparisons with the baseline pix2pix \cite{isola2017image} and U-net \cite{ronneberger2015u} and the previous work DeepNormal  \cite{hudon2018deep} and Sketch2Normal \cite{su2018interactive}, describe a small user study, and detail our ablation study. 

\subsection{Implementation Details}

All the lines of sketch images in our dataset are normalized and thinned to produce a standard data representation. If the user input sketch is not normalized and thinned, we apply a pre-trained line normalization model modified from \cite{simo2018real} to preprocess the user input. 

In the training process, the line drawings are first inverted from black-on-white to white-on-black and input to the network. The final output and the intermediate outputs $s_{1}$ and $s_{2}$ from the generator are similarly white shadows on black backgrounds. Inverting the images causes the network to converge faster. The generated shadows are composited with the line drawings as the ``fake'' image input to the discriminator. Similarly we composite the sketch and pure shadow in our dataset as the ``real'' image input to discriminator.

Because of limited size of our dataset of sketch/shadow pairs with annotated lighting direction we used the entire dataset for training---we did not reserve any of our training dataset for testing.  We trained for $80,000$ iterations with Adam optimizer \cite{kingma2014adam}. The optimizer parameters are set to $\mathrm{learning\ rate}=0.0002$, $\beta_1=0$, and $\beta_2=0.9$. The network is trained using one 12G Titan Xp with a batch size of 8 and $320\times320$ input image size. 

We shift, zoom in/out, and rotate to augment our dataset. When we rotate our line drawing input by each of \{0, 45, 90, 135, 180, 225, 270, 315\} degrees, we also rotate the ground truth shadow images and modify the lighting direction labels, by adding $1$ to the first digit for every 45 degrees of rotation. Shifting and zooming does not affect the lighting direction.

\subsection{Comparison with Prior Work}
\label{sec:comparison}
In this subsection, we qualitatively compare our approach to DeepNormal \cite{hudon2018deep} and Sketch2Normal \cite{su2018interactive}. Also, we compare our network to two baselines, Pix2pix \cite{isola2017image} and U-net \cite{ronneberger2015u}. The evaluation dataset is not included in training. The line drawings (without shadows) used for evaluation are collected from other artists and prior work to which we compare.

We generated the output from DeepNormal and Sketch2Normal using their source codes and trained models, unmodified. We use the scripts provided by DeepNormal to render shadows from normal maps. All normal maps are rendered under the same settings in this paper. To generate binary shadows, we threshold the continuous shadings at $0.5$. We note that DeepNormal additionally requires a mask to eliminate space outside the object; Sketch2Normal and our work do not require this mask.~We provide a hand-drawn mask as input to DeepNormal.~Our method and DeepNormal are predicted from $320\times320$ inputs and Sketch2Normal is predicted from $256\times256$ inputs. Since DeepNormal claims their results are consistent in various size of input, and Sketch2Normal experiment in $256\times256$ inputs. 

\begin{figure}
	\centering
	\includegraphics[width=\linewidth]{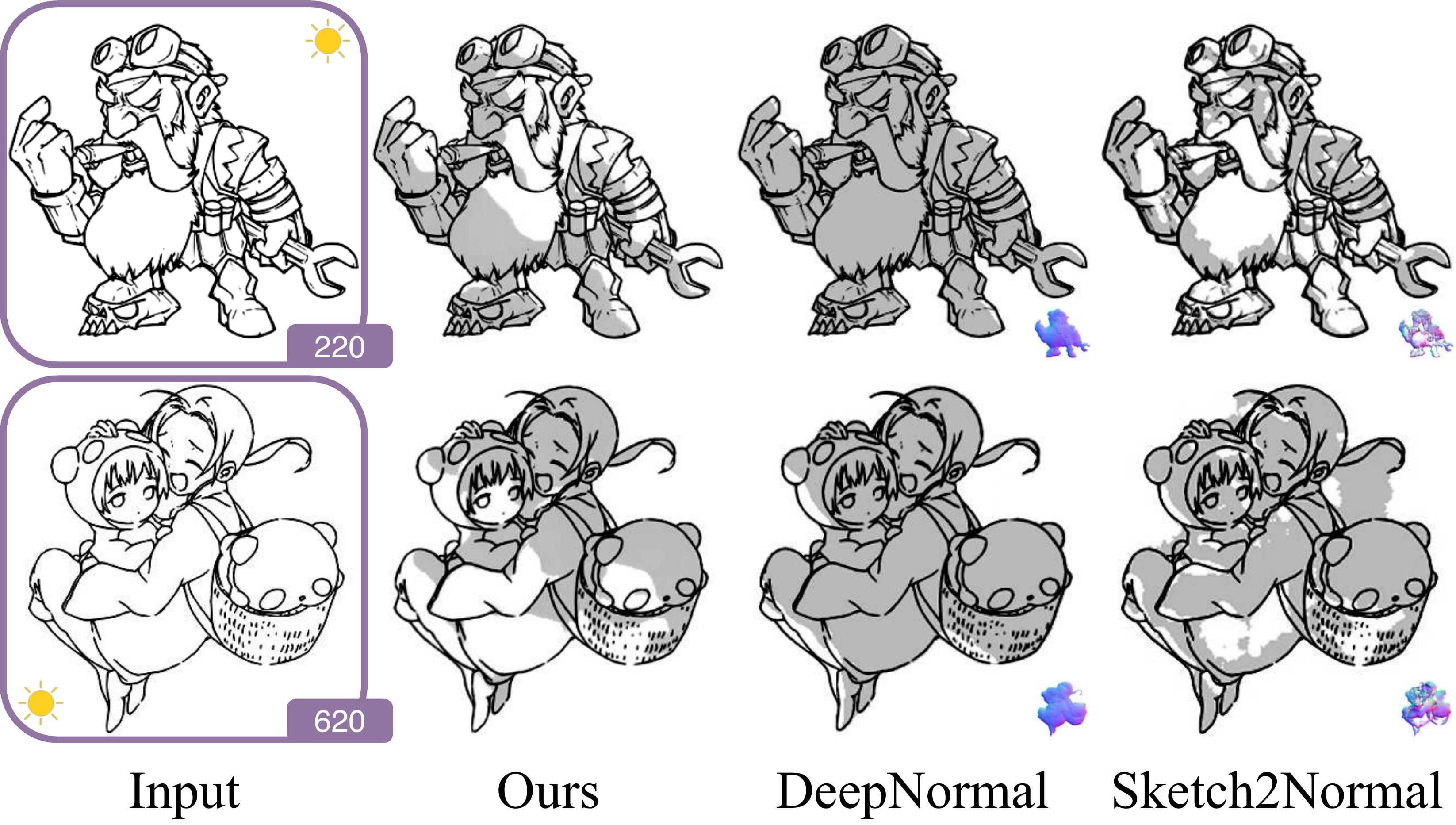}
	\caption{Comparisons with previous works DeepNormal \cite{hudon2018deep} and Sketch2Normal \cite{su2018interactive} with lighting depth ``2'' - in the plane (side lighting). The little sun denotes the lighting direction.}
	\label{fig:compare2}
\end{figure}

\begin{figure}
	\centering
	\includegraphics[width=\linewidth]{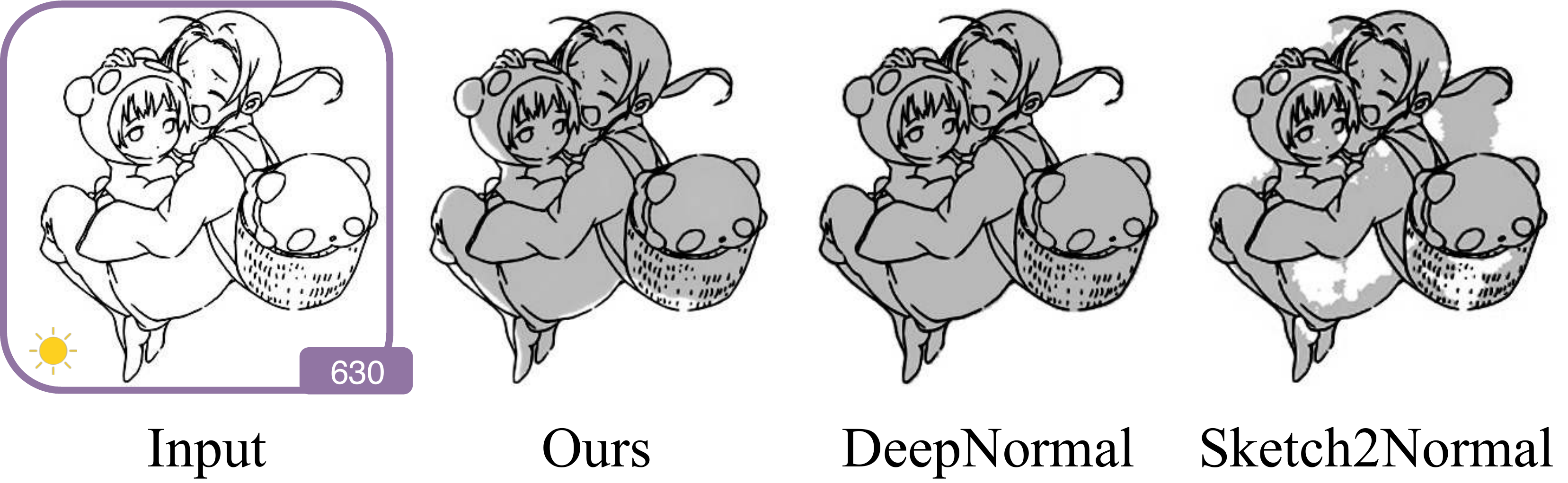}
	
	\caption{Comparison with DeepNormal \cite{hudon2018deep} and Sketch2Normal \cite{su2018interactive} when the light's depth is ``3'' - behind the plane (back lighting). Our approach demonstrates rim lighting.}
	\label{fig:compare3}
\end{figure}

\begin{figure}[t]
	\centering
	\begin{subfigure}{.19\columnwidth}
		\includegraphics[width=\columnwidth]{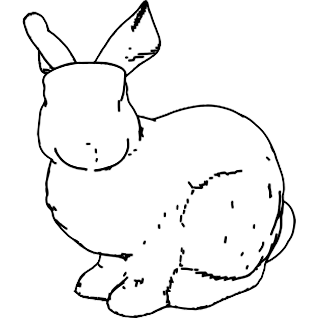}
		\subcaption*{Sketch}
	\end{subfigure}%
	\begin{subfigure}{.19\columnwidth}
		\includegraphics[width=\columnwidth]{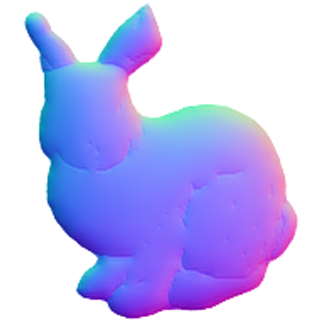}
		\subcaption*{\cite{hudon2018deep}-normal}
	\end{subfigure}%
	\begin{subfigure}{.19\columnwidth}
		\includegraphics[width=\columnwidth]{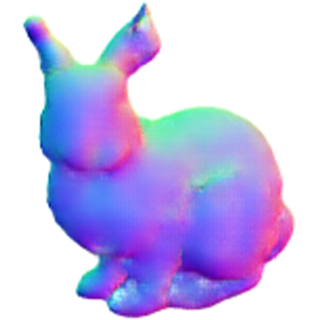}
		\subcaption*{\cite{su2018interactive}-normal}
	\end{subfigure}%
	\begin{subfigure}{.19\columnwidth}
		\includegraphics[width=\columnwidth]{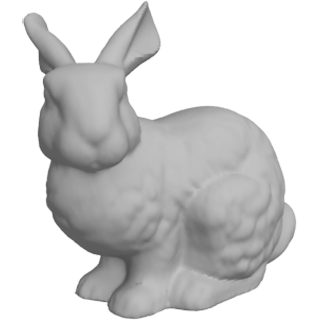}
		\subcaption*{3D model}
	\end{subfigure}%
	\begin{subfigure}{.19\columnwidth}
		\includegraphics[width=\columnwidth]{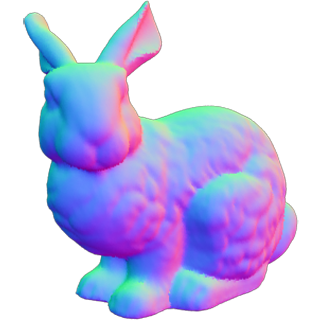}
		\subcaption*{GT-normal}
	\end{subfigure}
	
	\begin{subfigure}{.19\columnwidth}
		\includegraphics[width=\columnwidth]{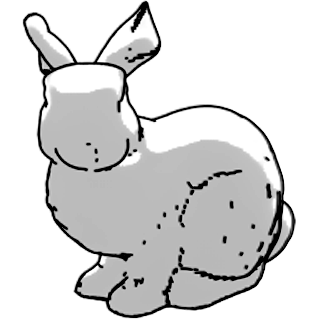}
		\subcaption*{Ours-120}
	\end{subfigure}%
	\begin{subfigure}{.19\columnwidth}
		\includegraphics[width=\columnwidth]{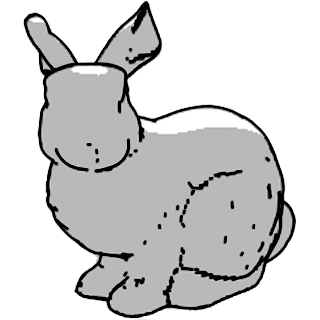}
		\subcaption*{\cite{hudon2018deep}-120}
	\end{subfigure}%
	\begin{subfigure}{.19\columnwidth}
		\includegraphics[width=\columnwidth]{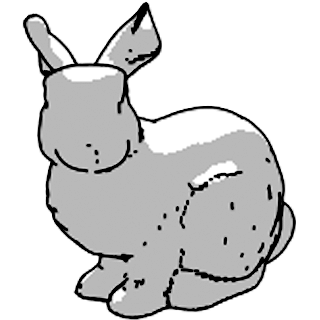}
		\subcaption*{\cite{su2018interactive}-120}
	\end{subfigure}%
	\begin{subfigure}{.19\columnwidth}
		\includegraphics[width=\columnwidth]{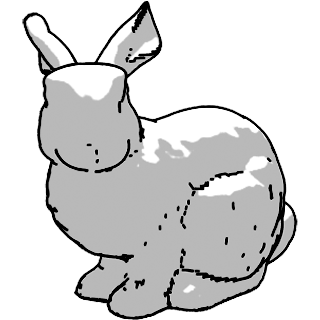}
		\subcaption*{$GT_{3D}$-120}
	\end{subfigure}%
	\begin{subfigure}{.19\columnwidth}
		\includegraphics[width=\columnwidth]{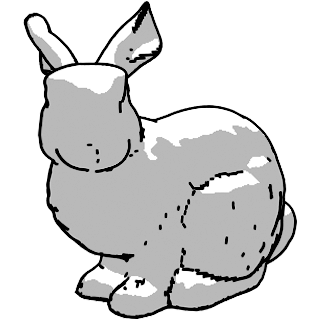}
		\subcaption*{GT-120}
	\end{subfigure}
	
	\begin{subfigure}{.19\columnwidth}
		\includegraphics[width=\columnwidth]{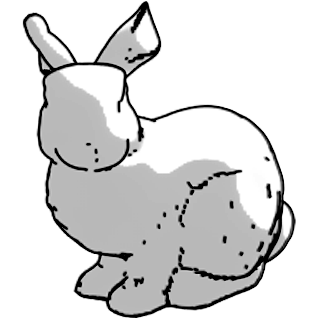}
		\subcaption*{Ours-220}
	\end{subfigure}%
	\begin{subfigure}{.19\columnwidth}
		\includegraphics[width=\columnwidth]{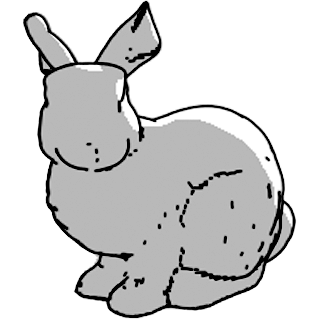}
		\subcaption*{\cite{hudon2018deep}-220}
	\end{subfigure}%
	\begin{subfigure}{.19\columnwidth}
		\includegraphics[width=\columnwidth]{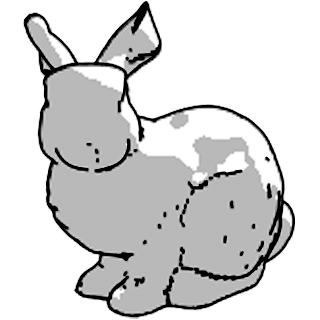}
		\subcaption*{\cite{su2018interactive}-220}
	\end{subfigure}%
	\begin{subfigure}{.19\columnwidth}
		\includegraphics[width=\columnwidth]{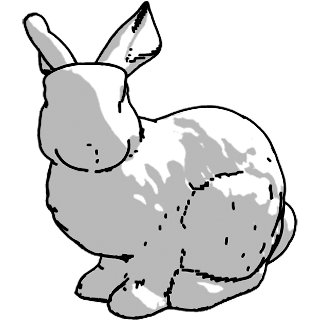}
		\subcaption*{$GT_{3D}$-220}
	\end{subfigure}%
	\begin{subfigure}{.19\columnwidth}
		\includegraphics[width=\columnwidth]{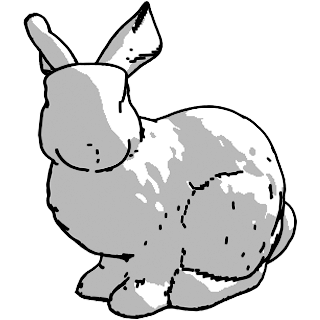}
		\subcaption*{GT-220}
	\end{subfigure}
	\caption{Comparisons between ground truth (GT), our approach, DeepNormal \cite{hudon2018deep}, and Sketch2Normal \cite{su2018interactive} rendered with a 3D bunny, with lighting depth ``2''. ``120": top, side lighting. ``220": upper right, side lighting. ``$GT_{3D}$'': rendered from commercial 3D software. ``GT'': rendered from its normal map.}
	\label{fig:normals}
\end{figure}

\begin{figure}
	\centering
	\includegraphics[width=\linewidth]{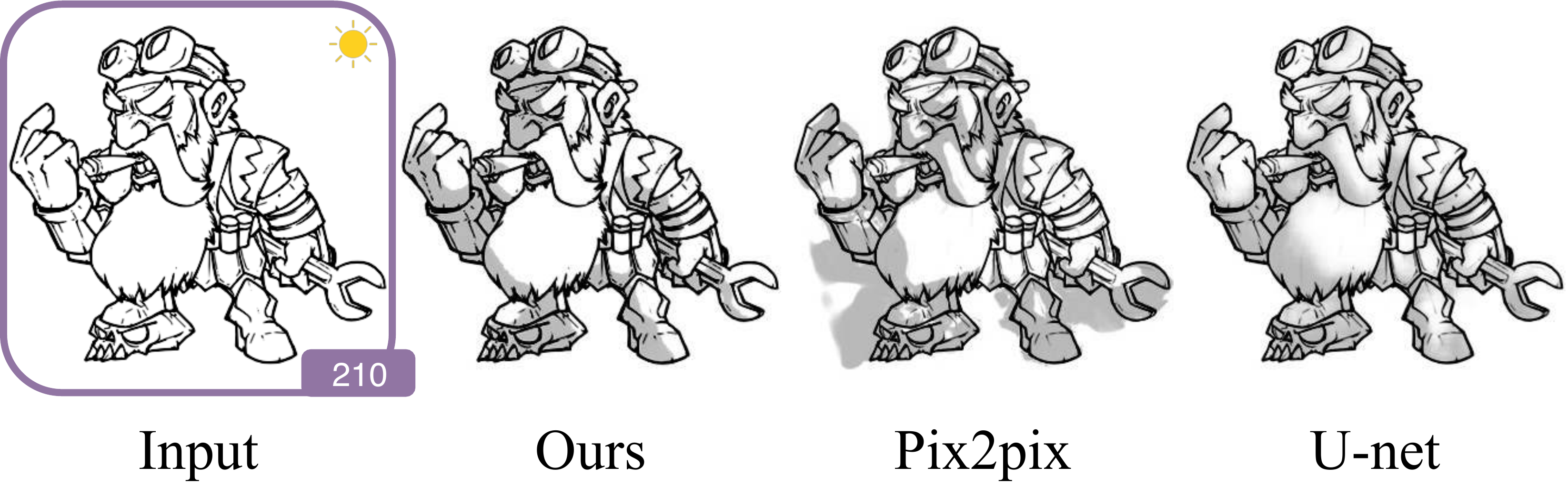}
	
	\caption{Comparison with Pix2pix \cite{isola2017image} and U-net \cite{ronneberger2015u} architectures trained on our dataset. Light depth is ``1''.}
	\label{fig:baseline}
\end{figure}

As shown in Figure~\ref{fig:compare1}, \ref{fig:compare2}, \ref{fig:normals}, \ref{fig:compare3}, \ref{fig:baseline}, our work performs favorably. For example, on the two-people and multiple-people line drawings (Figure~\ref{fig:compare1} second row), our work is able to shadow each character, however, DeepNormal and Sketch2Normal treat multiple people as one object. Notably, our work is superior in generating highly detailed shadows, such as in girl's hair and skirt. In terms of the complexity of sketch, though our training datasets have a moderate level of detail, our network performs well on complex sketches as shown in Figure~\ref{fig:compare1}. We also perform well beyond the object's boundary without requiring a mask.

Moreover, our work produces more precise details when the light source changes depth. As we can see in Figure~\ref{fig:compare2}, the shadows from DeepNormal \cite{hudon2018deep} cover almost the entire image, so that it seems as though the light is behind the object. However, in these images, the light source is in the same plane as the object, resulting in side lighting. In Figure~\ref{fig:normals}, we explain why DeepNormal \cite{hudon2018deep} underperforms when the light is in the object's plane by comparing with a 3D test model.~In particular, using our technique the shadows on the bunny's head and leg are closer to the ground truth and demonstrate self-shadowing. As highlighted in Figure~\ref{fig:compare2}, DeepNormal's normal maps have low variance due to multiple average of $256\times256$ tiles (refer to section 3.4 of DeepNormal). This low variance results in front lighting appearing to be side lighting and side lighting appearing to be back lighting. Some images generated by Sketch2Normal have some artifacts because the predicted normal maps have some blank areas. Because it is trained on simple sketches, Sketch2Normal struggles with complex sketches. Finally, we note that our approach produces artistic rim highlights from back lighting. Please refer to the supplementary material for the normal maps in Figure~\ref{fig:compare1} and more comparison figures.

Our architecture also performs favorable when qualitatively compared to Pix2pix and U-net trained on our dataset (Figure~\ref{fig:baseline}). Generally, U-net generates inaccurate soft shadows that are far from our goal of binary shadows. Pix2pix generates shadows far outside the object's boundary and ignores the geometric information in the sketch.~In our early research, we used a residual block autoencoder with skip connections, which generated soft shadows.~To achieve our goal of binary shadows, we added a discriminator and adopted a deeper \textit{RenderNet}. If the artist desires soft shadows, the intermediate output $s_{2}$ can be used.

\subsection{Artistic Control}
\label{section:artistic}
Though our network is trained with a discrete set of 26 lighting directions, the lighting direction is inputted to the network using floating point values in $[-1,1]^3$, allowing for the generation of shadows from arbitrary light locations.  Intuitively, our network learns a continuous representation of lighting direction from the discrete set of examples.  Furthermore, when a series of light locations are chosen the shadows move smoothly over the scene as in time-lapse video footage.  Please refer to the supplementary material for gifs demonstrating moving shadows. 

Although the final output of our network is binary shadows, if an artist desires soft shadows, the intermediate output, $s_{2}$, can be used, as shown in Figure~\ref{fig:gans}. 

\begin{figure}
	\centering
	\includegraphics[width=\linewidth]{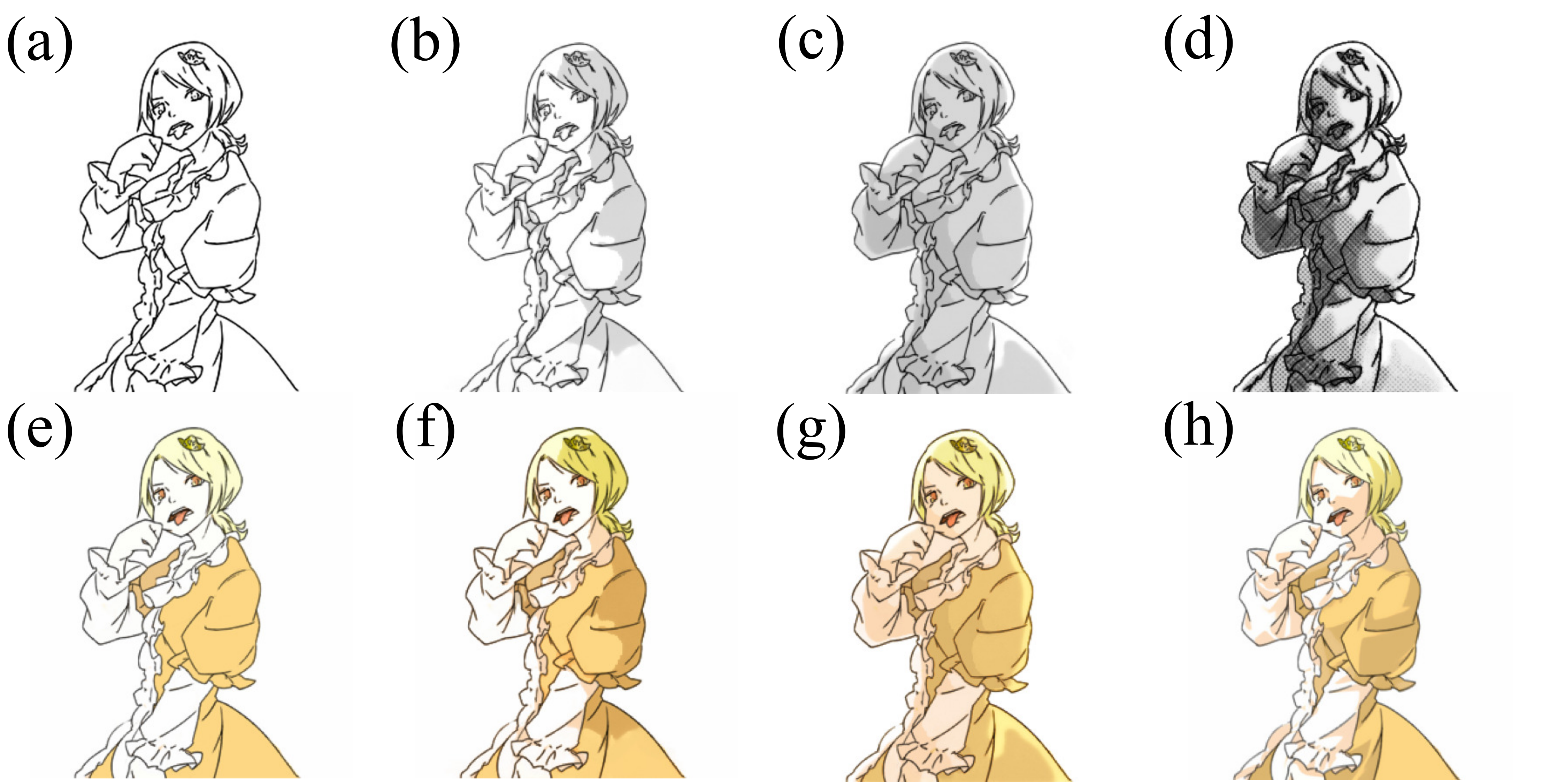}
	\caption{Combining our shadows with color. (a) Input sketch. (b) Our shadow with lighting direction ``710''. (c) Our shadows in complex lighting conditions created by compositing shadows from ``001'', ``730'', and ``210''. (d) Our shadows composited from lighting directions ``001'', ``210'', ``220'' with dots and soft shadow to produce a manga style. (e) Colorized sketch with commercial software. (f) Composite of (e) and (b). (g) Composite of (e) and (c). (h) Original artist's image. \textcopyright nico-opendata
	}
	\label{fig:color}
\end{figure}

Our work is complementary to prior work on automatic colorization of sketches \cite{yonetsuji2017paintschainer, kim2019tag2pix, zhang2018two, furusawa2017comicolorization, frans2017outline}.  Figure~\ref{fig:color} demonstrates that our shadows can be combined with these colorization approaches.  While most prior work on colorization combines shading and shadowing effects, it would be interesting to separate these effects into indpendent image layers for further artistic editing.

\begin{table}
	\begin{center}
		\begin{tabular}{c c c c c c c}
			\toprule
			Methods & GT & Ours & \cite{hudon2018deep} & \cite{su2018interactive} & \cite{isola2017image} & \cite{ronneberger2015u} \\
			\midrule
			Turing & 68\% & \textbf{69\%} & 51\% & 11\% & 23\% & 19\% \\
			Scores & 6.37 & \textbf{6.70} & 5.78 & 3.35 & 3.77 & 3.06 \\
			\midrule
			Methods & GT & Ours & \cite{hudon2018deep} & \cite{su2018interactive} & \cite{isola2017image} & \cite{ronneberger2015u} \\
			\midrule
			Turing & \textbf{70\%} & 65\% & 45\% & 10\% & 25\% & 17\% \\
			Scores & 6.50 & \textbf{6.66} & 5.69 & 3.44 & 3.91 & 3.03 \\
			\bottomrule
		\end{tabular}
	\end{center}
	\caption{Results of user study comparing Ground truth (GT) in our datasets, Ours, DeepNormal \cite{hudon2018deep}, Sketch2Normal \cite{su2018interactive}, Pix2pix \cite{isola2017image} baseline and U-net \cite{ronneberger2015u} baseline. First row: percentage that pass the Turing test. Second row: average scores. 9 is the best score. Top: total results. Bottom: results of people with drawing experience.}
	\label{table:userstudy}
\end{table}

\subsection{User Study}
To evaluate our approach we conducted a small user study. We generated shadows using six different techniques and asked users to evaluate the results. We train the users with some samples from our dataset at the beginning. Our user study had two stages: a ``Turing'' test that asked the simple question ``{\em Do you think this shadow was drawn by a human?  Yes or no?}'' and another stage where the user is shown an image and asked to rate the quality of the shadow with the prompt ``{\em Under this lighting direction, evaluate the appearance of this shadow}'' on a Likert scale from 1 to 9 (9 being best).  In each stage the user was shown 36 images generated from six input sketches and each of six shadow generation methods: ground truth shadows created by artists, Ours, DeepNormal \cite{hudon2018deep}, Sketch2Normal \cite{su2018interactive}, Pix2pix \cite{isola2017image}, and U-net \cite{ronneberger2015u}. For the synthetic shadows, the lighting directions were chosen randomly (Ours exclude the directions in ground truth), with the restriction that we did not use back lighting.  We only used front lighting for DeepNormal for the reasons described in Section \ref{sec:comparison}.  For the quality rating, lighting directions were described with text, e.g. ``upper right, front lighting.''  For the Turing test, no lighting directions were given.  Users were shown one image at a time, but could use the ``back'' and ``forward'' buttons.

Users received a brief training that displayed 15 ground truth shadowed sketches from our dataset and highlighted the differences between front lighting and side lighting.  We also asked the users to rate their drawing experience as ``professional'', ``average'', ``beginner'' or ``0 experience''.  We distributed the survey online and received 60 results. Forty participants had drawing experience: 13 were professional artists, 11 were average level, and 16 were beginners.  The results are shown in Table~\ref{table:userstudy}.  Our approach performs favorably, almost matching the ground truth shadows created by artists. We ran a one-way ANOVA to analyze the Likert scores. The results confirmed that our results were quantitatively similar to ground truth ($p=0.24$) and better than the other methods ($p < 0.05$ for all of the comparisons). Please refer to the supplementary material for more details of the statistical significance report of our user study.

\begin{figure}
	\centering
	\includegraphics[width=.9\columnwidth]{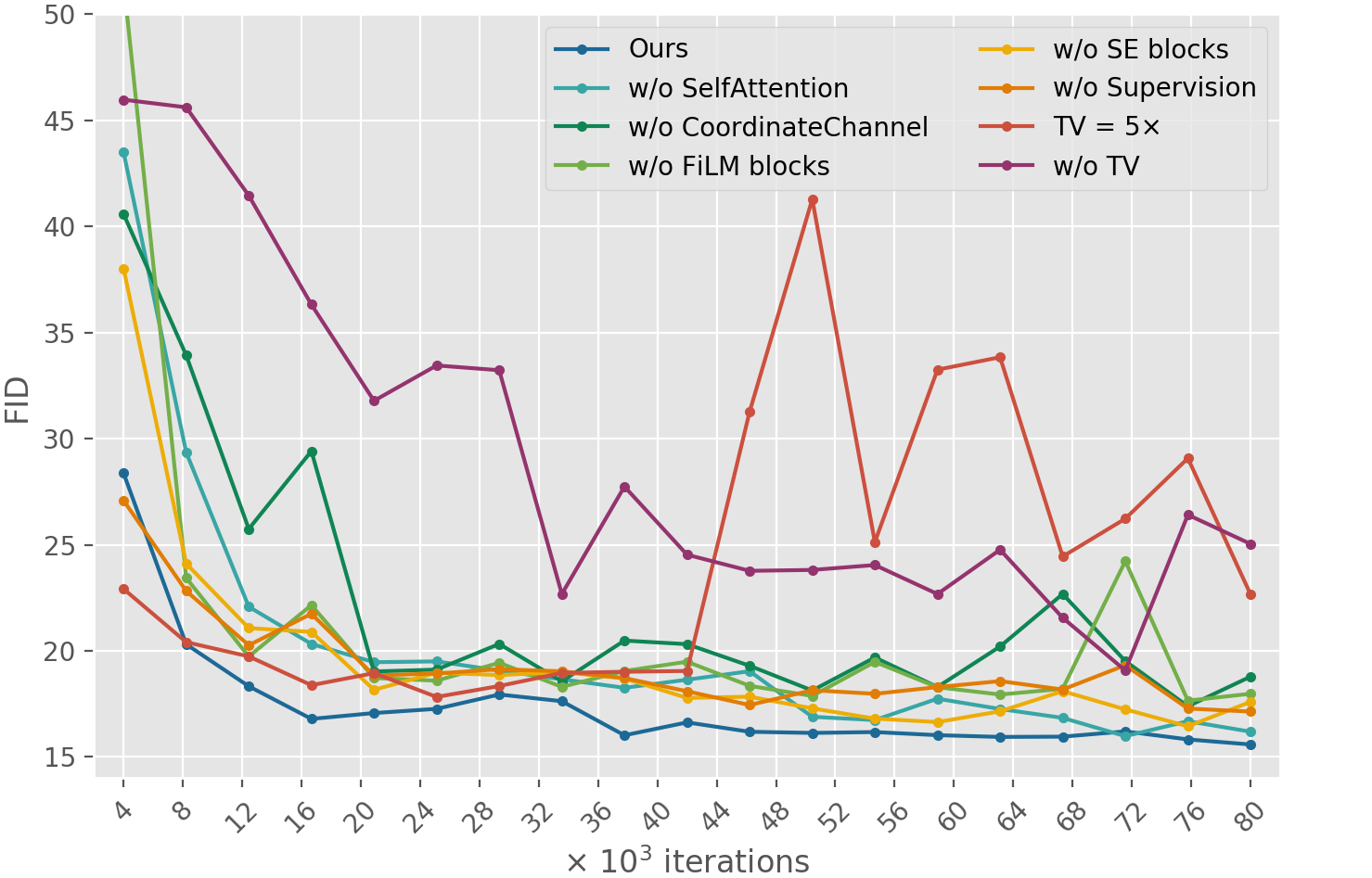}
	\caption{FID scores of ours and ablation studies. Our model's line is on the most bottom.}
	\label{fig:fid}
\end{figure}

\begin{figure}
	\centering
	\begin{subfigure}{.25\columnwidth}
		\includegraphics[width=\columnwidth]{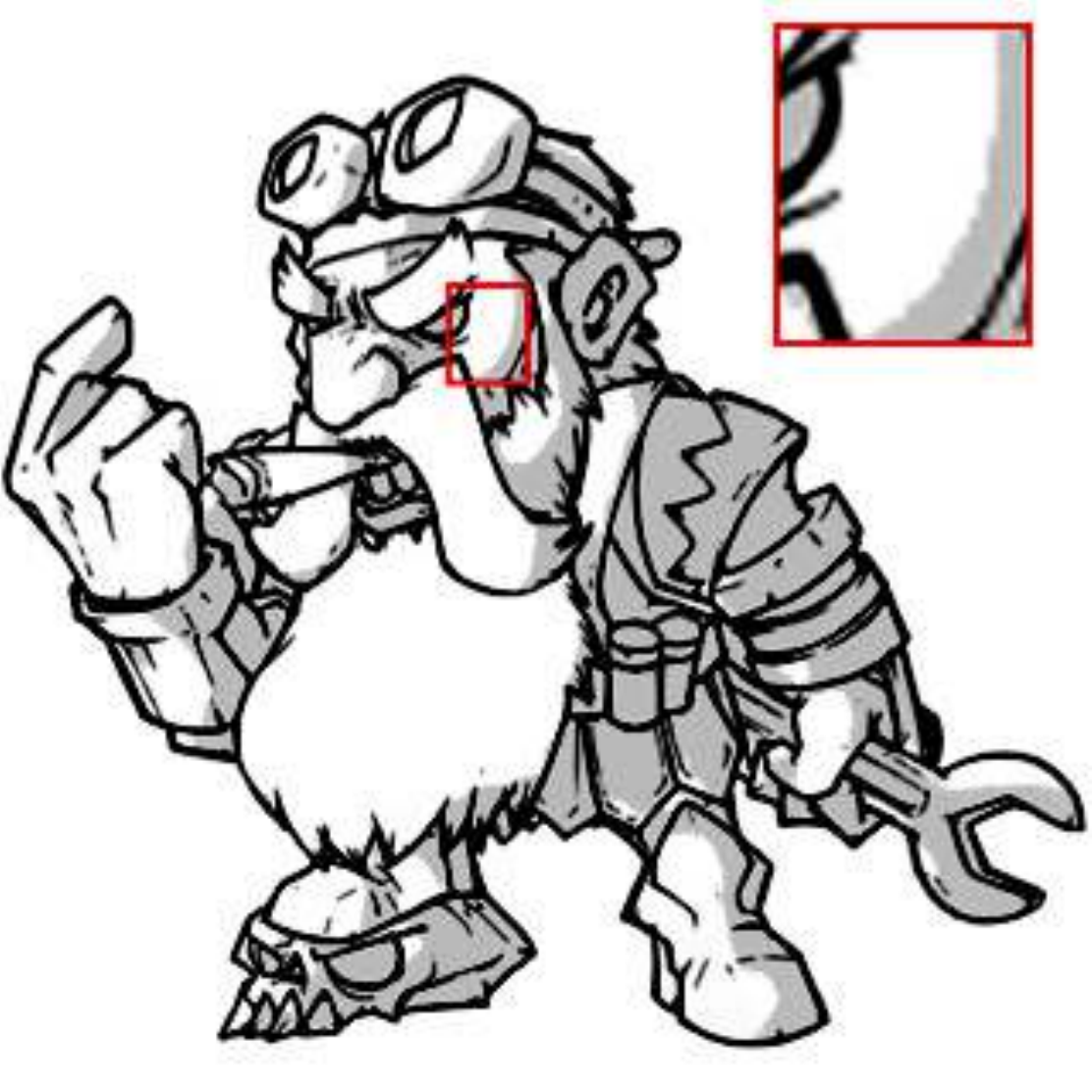}
		\subcaption{ours}
	\end{subfigure}%
	\begin{subfigure}{.25\columnwidth}
		\includegraphics[width=\columnwidth]{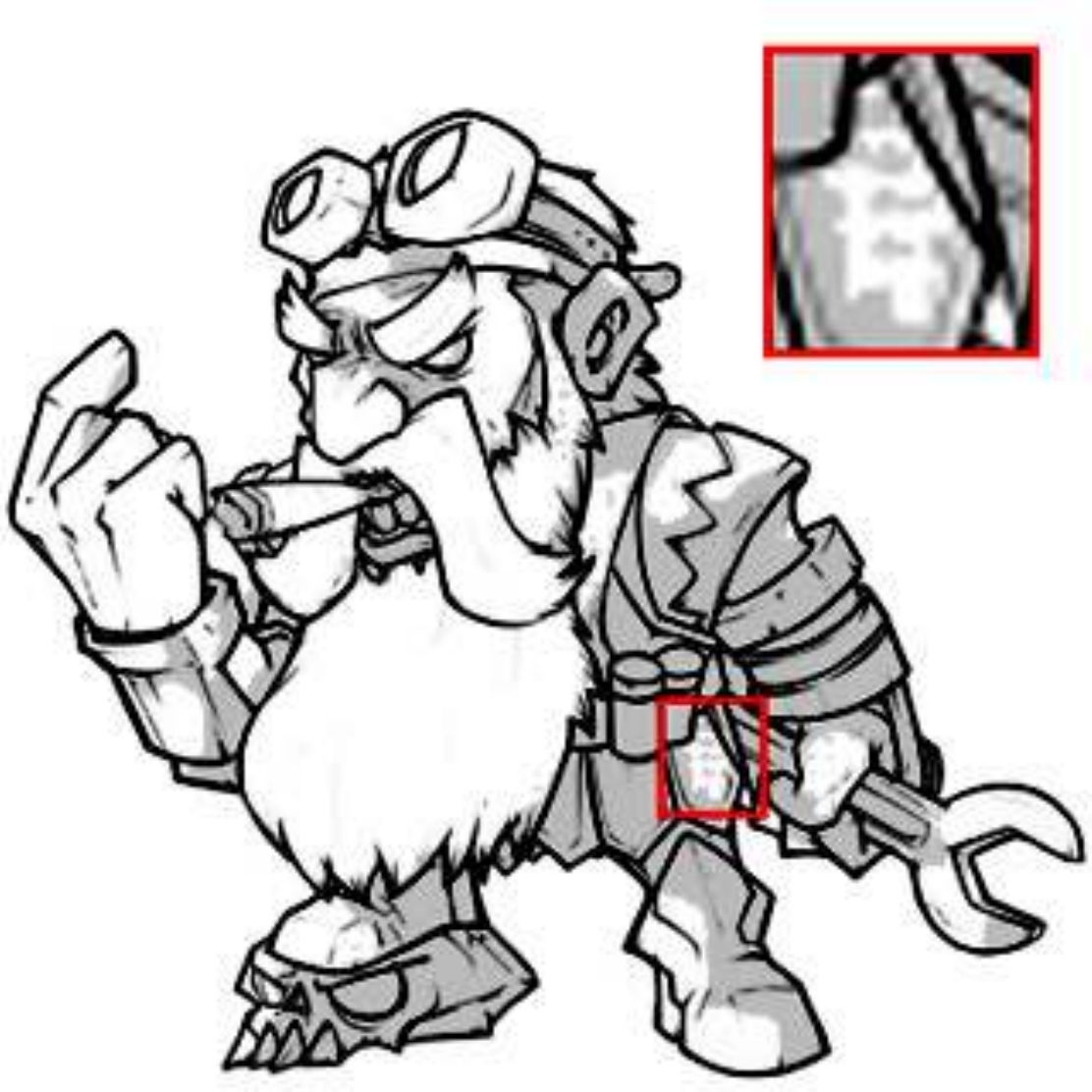}
		\subcaption{w/o SA}
	\end{subfigure}%
	\begin{subfigure}{.25\columnwidth}
		\includegraphics[width=\columnwidth]{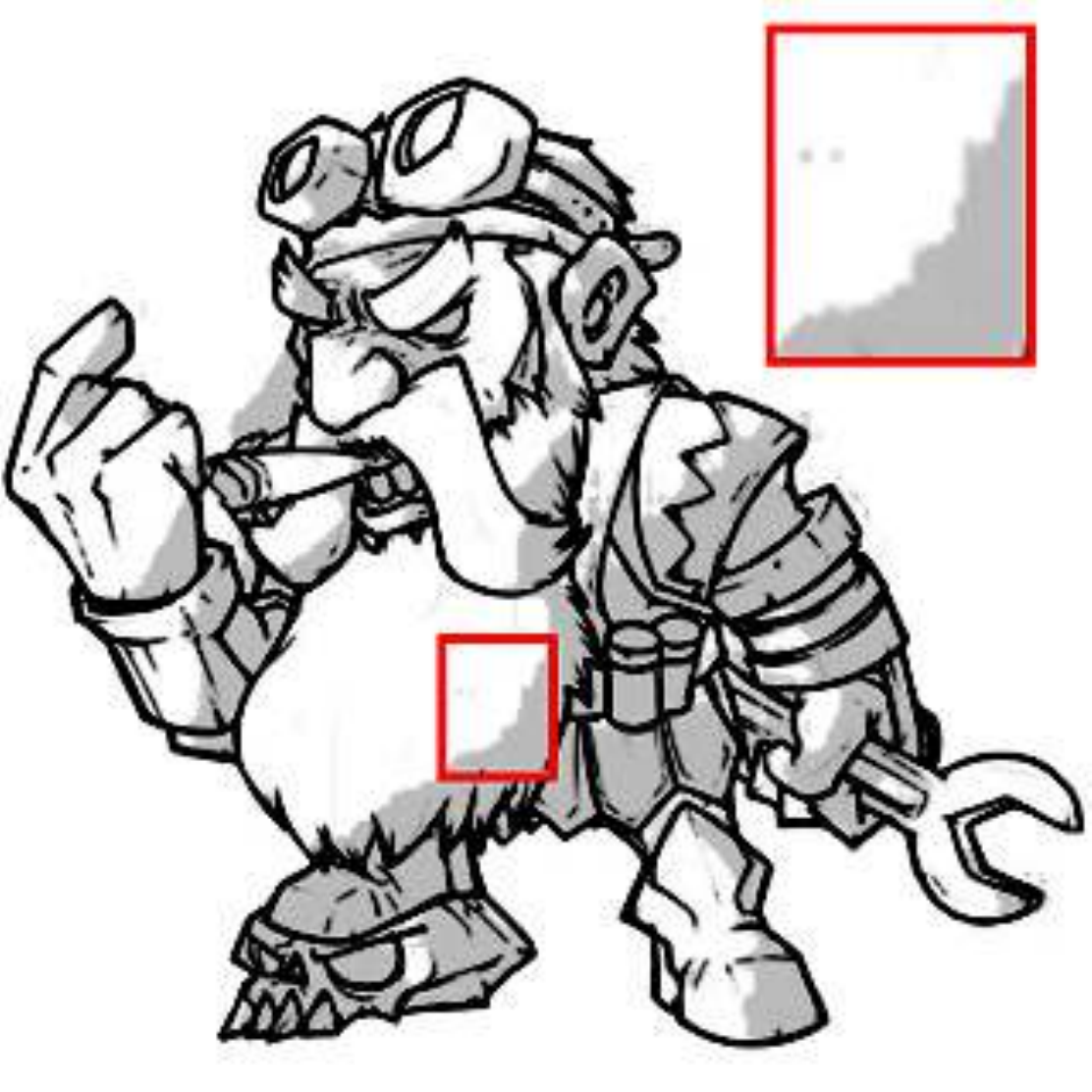}
		\subcaption{w/o CC}
	\end{subfigure}%
	\begin{subfigure}{.25\columnwidth}
		\includegraphics[width=\columnwidth]{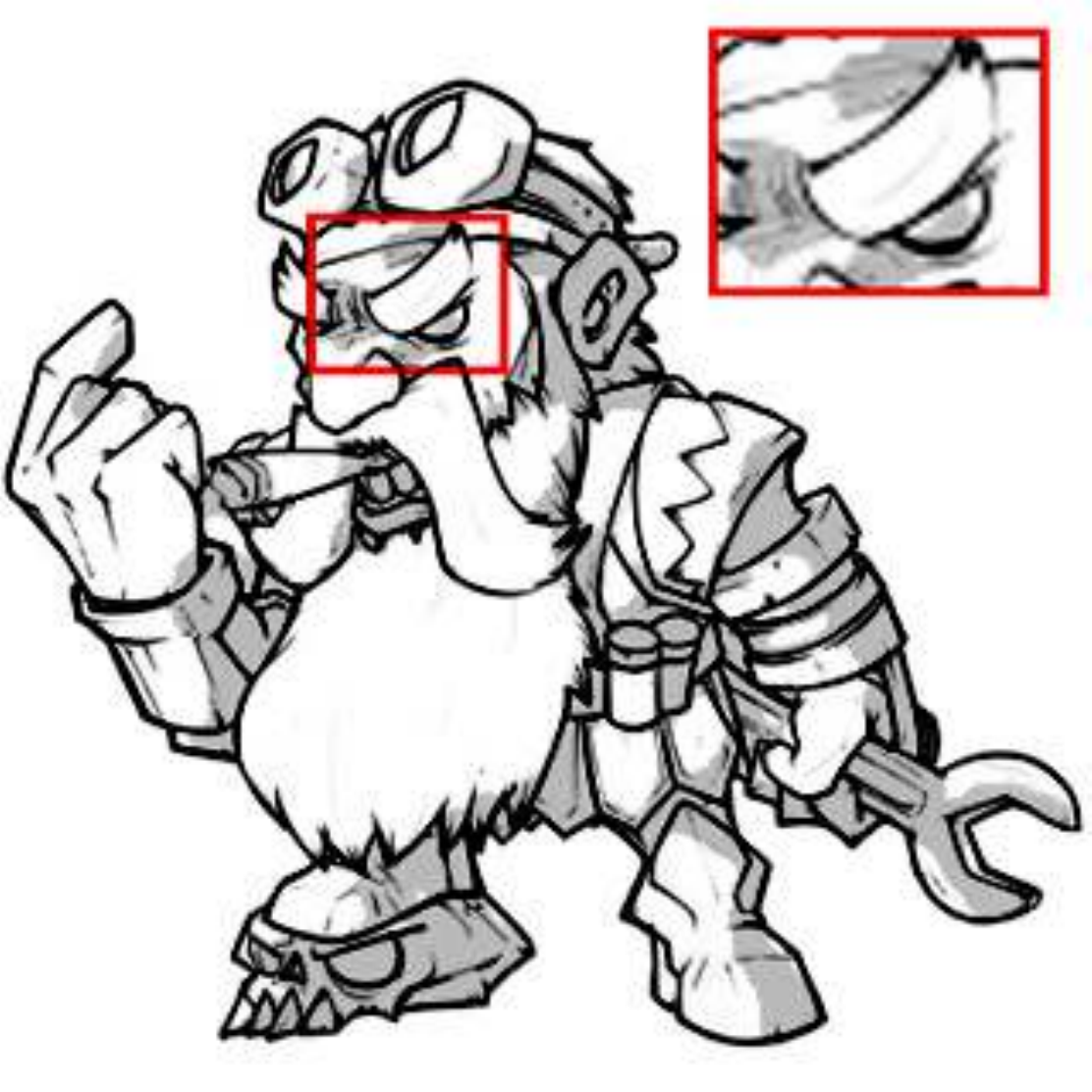}
		\subcaption{w/o FiLM}
	\end{subfigure}
	
	\begin{subfigure}{.25\columnwidth}
		\includegraphics[width=\columnwidth]{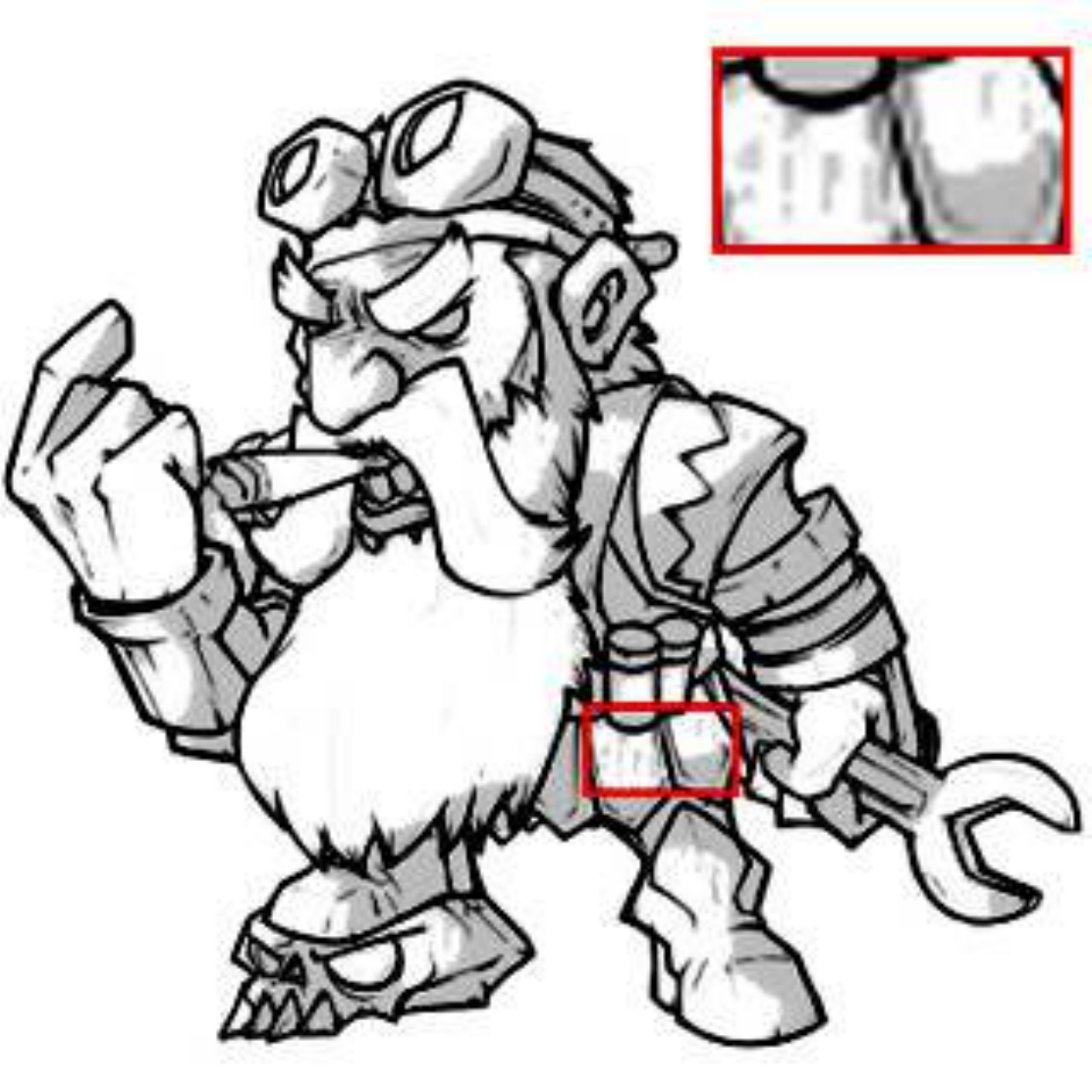}
		\subcaption{w/o SE}
	\end{subfigure}%
	\begin{subfigure}{.25\columnwidth}
		\includegraphics[width=\columnwidth]{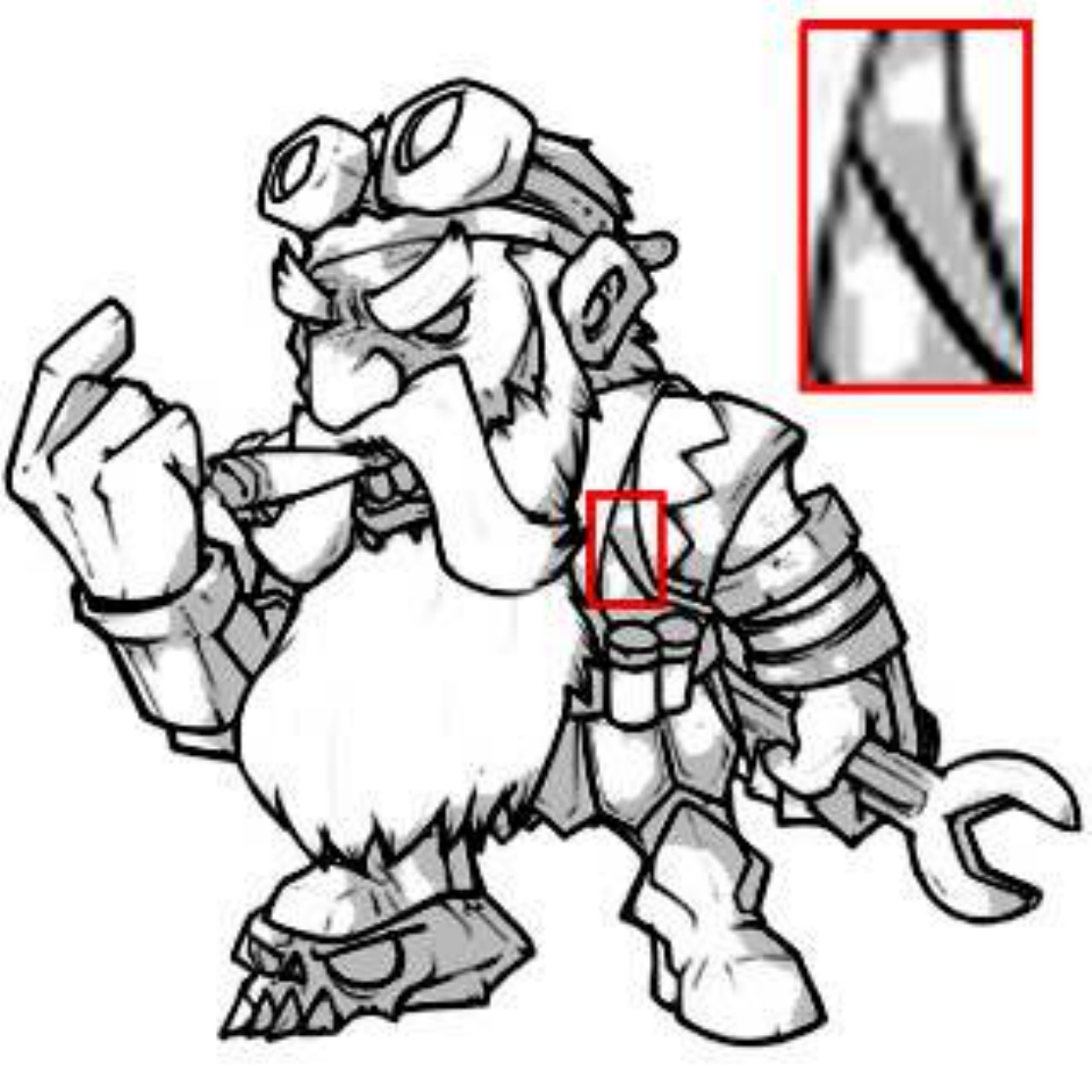}
		\subcaption{w/o $s_{1}$, $s_{2}$}
	\end{subfigure}%
	\begin{subfigure}{.25\columnwidth}
		\includegraphics[width=\columnwidth]{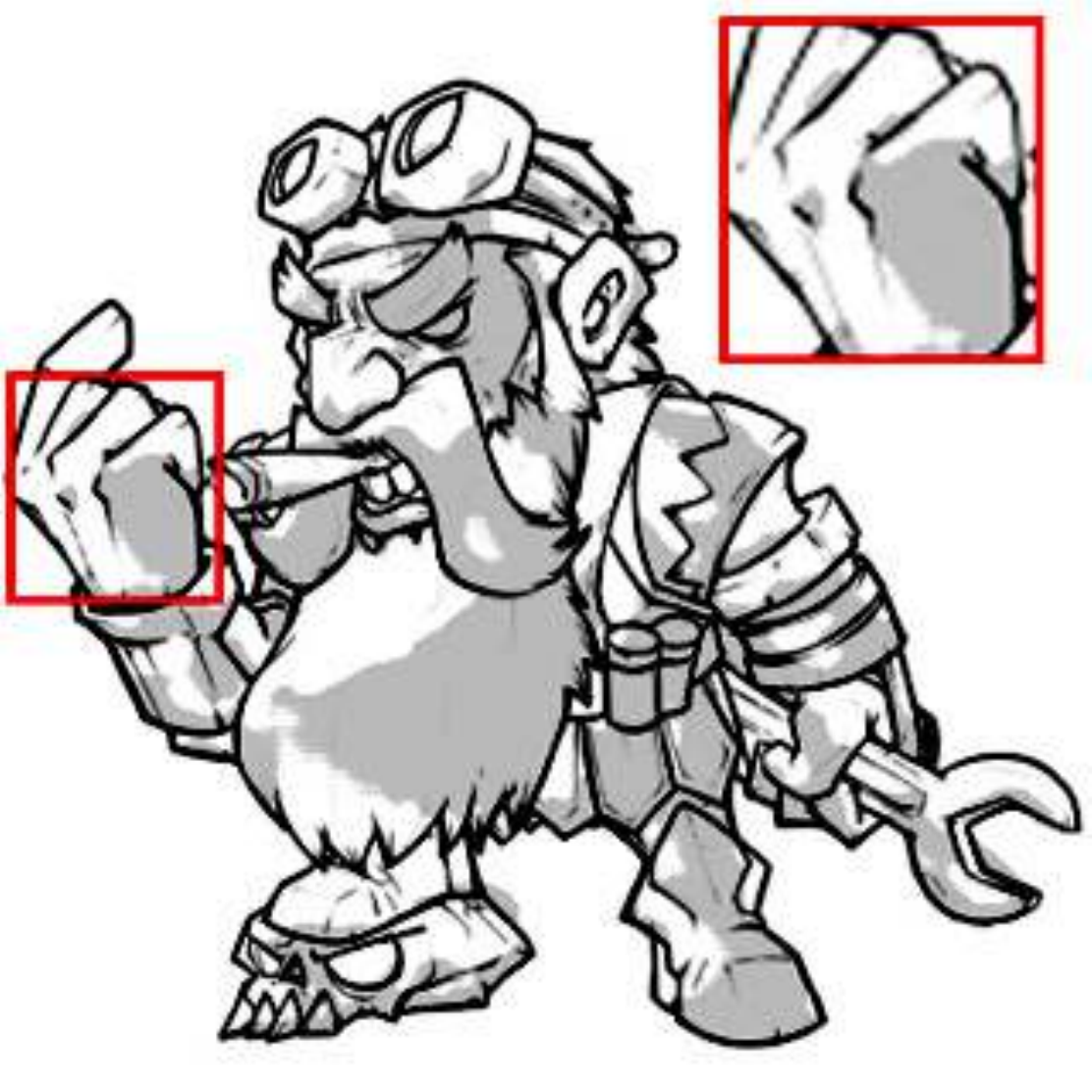}
		\subcaption{$TV=5\times$}
	\end{subfigure}%
	\begin{subfigure}{.25\columnwidth}
		\includegraphics[width=\columnwidth]{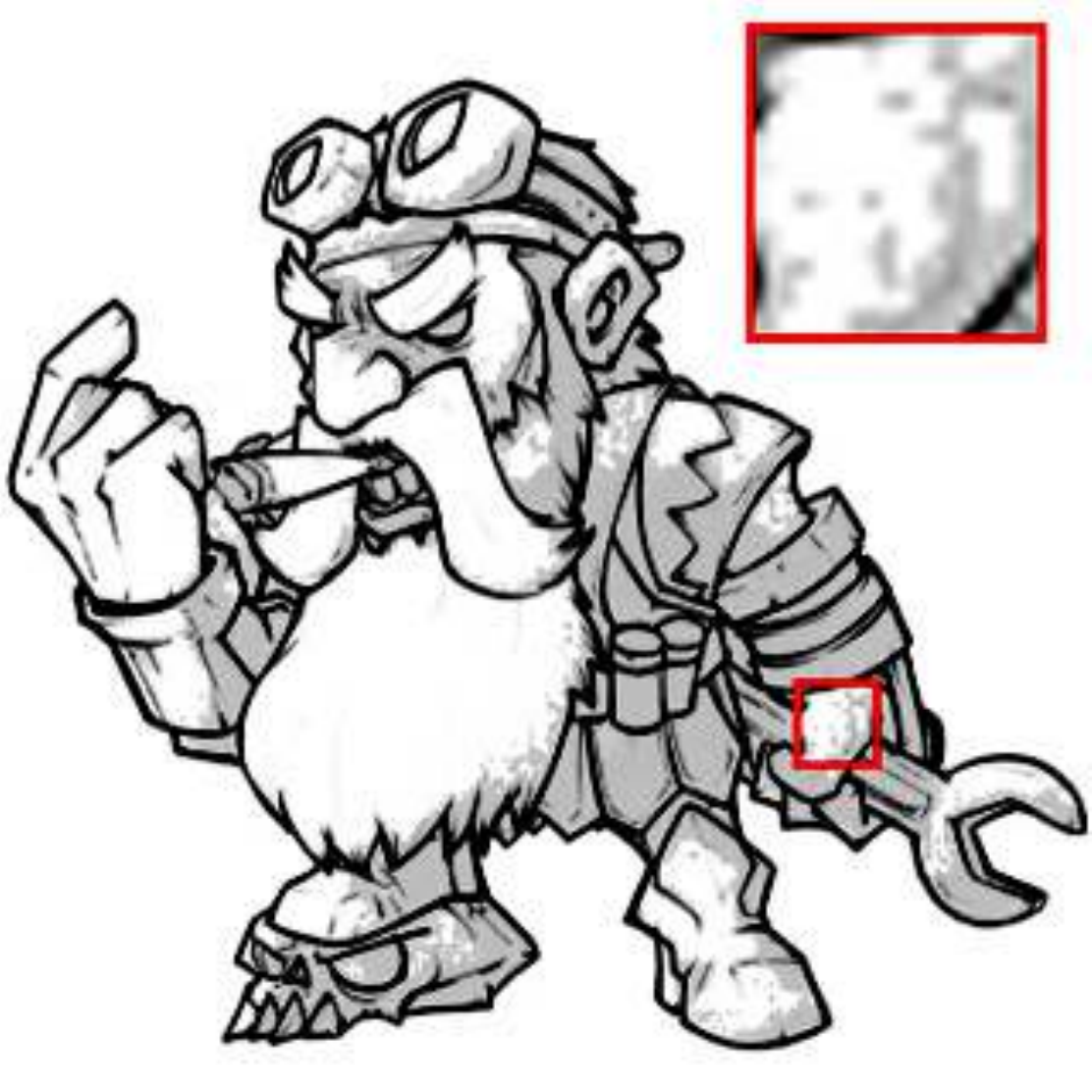}
		\subcaption{w/o TV}
	\end{subfigure}
	
	\caption{Ablation studies. (b) removing the Self-attention (SA) layers, (c) removing the Coordinate Channel (CC), (d) removing the FiLM residual blocks, (e) removing the SE blocks, (f) removing the two deep supervised outputs ($s_{1}$, $s_{2}$), (g) increasing the TV loss weights to $e^{-5}$, (h) removing the total variant (TV) regularizer.}
	\label{fig:abstudy}
\end{figure}

\subsection{Ablation Study}
We performed seven ablation studies as shown in Figures~\ref{fig:abstudy} and \ref{fig:fid}. For quantitative comparison, we calculated the Fr{\'e}chet Inception Distance (FID) \cite{heusel2017gans} per 4000 iterations of our work and the ablation studies using the entire dataset. Figure~\ref{fig:fid} shows that our work has the lowest and most stable FID. This demonstrate that each feature we propose is essential and that the total variation regularizer was critically important.

Figure~\ref{fig:abstudy} qualitatively demonstrates that without the elements we propose, the networks performance is degraded: boundaries become aliased and artifacts appear in shadows. Among all ablation studies, ``w/o Self-attention'' has the least influence, as the shown Figure~\ref{fig:abstudy} (b) and the FID in Figure~\ref{fig:fid}. Setting the coefficient of the total variation regularizer $5\times$ larger or removing the regularizer has the most influence on the overall performance and ruins the smoothness of shadow. The corresponding FID also highlight the importance of the total variant regularizer. 

In Figure~\ref{fig:abstudy}, all of the images use the same lighting direction ``810''. Generally, when the Self-attention layers are removed, the network performs poorly with details and there are tiny artifacts within the shadow block; without the Coordinate Channel or FiLM residual block, the output will have unrealistic shadow boundaries and shadows outside the object's boundary; without SE blocks, there will be shadow ``acne'' and the overall appearance looks messy; without the two deep supervised outputs ($\lambda_{1}=.4$, $\lambda_{2}=.9$, $\lambda_{3}=\lambda_{4}=0$), the output will have dot artifacts in a grid pattern and lower accuracy; if the network has $5\times$ higher weight for the TV regularizer or is missing the TV regularizer, the network will converge too fast and trap in a local minimum. 

\section{Future Work}
The network performance is not invariant on different sizes of input images. Mostly the $320\times320$ inputs have the best performance, because our network is trained on $320\times320$ size inputs. $480\times480$ input images also have good performance. Though we almost match the ground truth in user study, our generated shadows are not so much detailed as ground truth, especially on hard surface object (\eg desk, laptop). Also, if inputting a local part of the line drawing, the network is not able to reason the correct shadows. As future work, we will develop a network that can output various image sizes to meet the high resolution requirements of painting.

\section{Conclusion}
Our conditional Generative Adversarial Network learns a non-photorealistic renderer that can automatically generate shadows from hand-drawn sketches. We are the first to attempt to directly generate shadows from sketches through deep learning.~Our results compare favorably to prior art that renders normal maps from sketches on both simple and sophisticated images. We also demonstrate that our network architecture can ``understand" the 3D spatial relationships implied by 2D line drawings well enough to generate detailed and accurate shadows.

{\bf Acknowledgements:} The authors wish to thank Yuchen Ma and Kejun Liu for annotating the dataset and the reviewers, Tiantian Xie, and Lvmin Zhang for many suggestions.

{\small
\bibliographystyle{ieee_fullname}
\bibliography{paperbib}
}

\cleardoublepage

\appendix

\noindent{\Large\bfseries Appendix\par}

\section{Lighting Directions}
We found `810' numbering scheme to be more intuitive for the users than the other two methods ($[-1, 0, 1]^3$ or an integer between 1 and 26). Therefore, we use `810' scheme as the user inputs, then transfer the `810' scheme (first column of Table~\ref{table1}) to the $[-1, 0, 1]^3$ scheme (third column of Table~\ref{table1}) in programming.

\begin{table}[h]
	\begin{center}
		\begin{tabular}{c c c}
			\toprule
			\textbf{Label} & \textbf{Direction} & \textbf{Position}  \\
			\midrule
			001 & rear center & [0,0,1]  \\
			002 & front center & [0,0,-1]\\
			110 & center top, front lighting & [0,1,-1]\\
			120 & center top, side lighting & [0,1,0]\\
			130 & center top, back lighting & [0,1,1]\\ 
			210 & upper right, front lighting & [1,1,-1]\\
			220 & upper right, side lighting & [1,1,0]\\ 
			230 & upper right, back lighting & [1,1,1]\\ 
			310 & center right, front lighting & [1,0,-1]\\
			320 & center right, side lighting & [1,0,0]\\ 
			330 & center right, back lighting & [1,0,1]\\
			410 & lower right, front lighting & [1,-1,-1]\\
			420 & lower right, side lighting  & [1,-1,0]\\
			430 & lower right, back lighting & [1,-1,1]\\
			510 & bottom, front lighting & [0,-1,-1]\\
			520 & bottom, side lighting & [0,-1,0]\\
			530 & bottom, back lighting & [0,-1,1]\\
			610 & lower left, front lighting & [-1,-1,-1]\\
			620 & lower left, side lighting & [-1,-1,0]\\
			630 & lower left, back lighting & [-1,-1,1]\\
			710 & center left, front lighting & [-1,0,-1]\\
			720 & center left, side lighting & [-1,0,0]\\
			730 & center left, back lighting & [-1,0,1]\\
			810 & upper left, front lighting & [-1,1,-1]\\
			820 & upper left, side lighting & [-1,1,0]\\
			830 & upper left, back lighting & [-1,1,1]\\
			\bottomrule
		\end{tabular}
	\end{center}
	\caption{A lookup table of our 26 lighting direction labels, the actual lighting directions, and $[-1, 0, 1]^3$ style positions in programming.}
	\label{table1}
\end{table}

\section{Pre-processing}
The pre-processing is a light neural network modified from smart inker \cite{simo2018real}. We re-trained the network with synthetic data (0.2--2 px, cairosvg standard, various darkness).  This pre-processing network was sufficiently robust for the Japanese and Disney style images we have tested, which had line widths in the range 1--6px.

Figures~\ref{fig:art1}, ~\ref{fig:art2}, ~\ref{fig:art3} and ~\ref{fig:norm} give some indication of how this pre-processing network performs ``in the wild on a wide range of line styles.''

\section{Network Architecture}
\begin{figure}[!htb]
	\centering
	\includegraphics[scale=0.94]{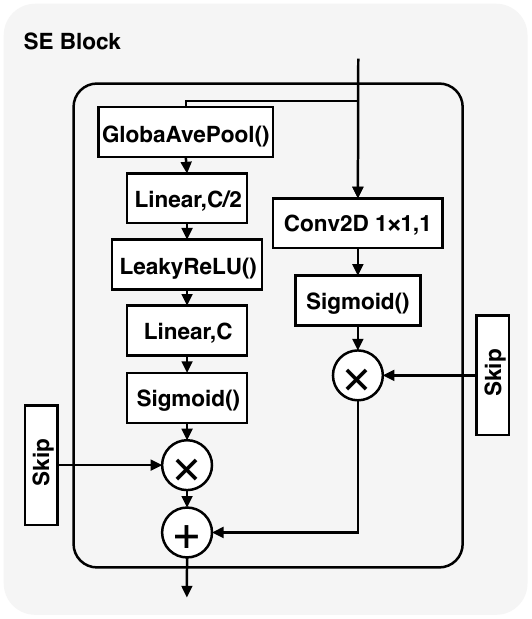}
	\caption{\textbf{SE (SE Block)}. `C' is filter size.}
	\label{fig:seblock}
\end{figure}

\begin{figure}[!htb]
	\centering
	\includegraphics[scale=0.94]{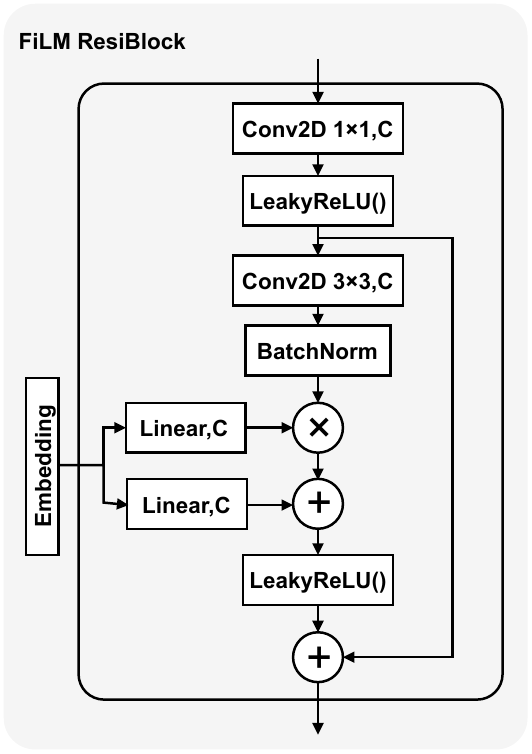}
	\caption{\textbf{FiLMResiBlock}\cite{perez2018film}. `C' is filter size.}
	\label{fig:filmblock}
\end{figure}

More details of the network architecture are in Table~\ref{table:G_encoder_light}, ~\ref{table:resi_block}, ~\ref{table:down_resi_block}, ~\ref{table:up_resi_block}, ~\ref{table:G_encoder}, ~\ref{table:D}, ~\ref{table:G_decoder} and Figure~\ref{fig:seblock}, ~\ref{fig:filmblock}. Please refer to the main body of our paper for the network architecture figure.

`ResiBlock': Residual Blocks. `DownResiBlock': Downscale Residual Blocks. `UpResiBlock': Upscale Residual Blocks. `ShapeNet': the encoder of Generator. `RenderNet': the decoder of Generator.


We use a fully connected layer (Table~\ref{table:G_encoder_light}) to embed the $[-1,1]^3$ lighting positions. We repeatedly input the embedded lighting position into each stage of RenderNet where a FiLMResiBlock exists.  However, we only input the lighting direction once without embedding at the beginning of Discriminator.

The inputs of the Generator are the line drawing, pure shadow (ground truth), and the lighting direction. The outputs of the Generator are the final output (pure binary shadow), $s_{1}$ and $s_{2}$. The inputs of the Discriminator are the composition of line drawing and pure shadow (ground truth and the final output of Generator), and the lighting direction.

\begin{table}[!htb]
	\begin{center}
		\begin{tabular}{c c c}
			\toprule
			\textbf{Layer} & \textbf{Filter} & \textbf{Output Size}\\
			\midrule
			Linear & 128 & 128\\
			Tanh() & - & 128\\
			\bottomrule
		\end{tabular}
	\end{center}
	\caption{\textbf{Light Position embedding}}
	\label{table:G_encoder_light}
\end{table}

\begin{table}[!htb]
	\begin{center}
		\begin{tabular}{c}
			\toprule
			\textbf{F(x)}\\
			\midrule
			BatchNorm\\
			LeakyReLU()\\
			Conv2D($1\times1$)\\
			BatchNorm\\
			LeakyRuLU()\\
			Conv2D($3\times3$)\\
			BatchNorm\\
			LeakyReLU()\\
			Conv2D($1\times1$)\\
			\midrule
			\textbf{Shortcut Branch}\\
			\midrule
			Conv2D($1\times1$)\\
			\bottomrule
		\end{tabular}
	\end{center}
	\caption{\textbf{ResiBlock}}
	\label{table:resi_block}
\end{table}

\begin{table}[!htb]
	\begin{center}
		\begin{tabular}{c}
			\toprule
			\textbf{F(x)}\\
			\midrule
			BatchNorm\\
			LeakyReLU()\\
			Conv2D($1\times1$)\\
			BatchNorm\\
			LeakyRuLU()\\
			Conv2D($3\times3$, strides=2)\\
			BatchNorm\\
			LeakyReLU()\\
			Conv2D($1\times1$)\\
			\midrule
			\textbf{Shortcut Branch}\\
			\midrule
			Conv2D($1\times1$, strides=2)\\
			\bottomrule
		\end{tabular}
	\end{center}
	\caption{\textbf{DownResiBlock}}
	\label{table:down_resi_block}
\end{table}

\begin{table}[!htb]
	\begin{center}
		\begin{tabular}{c}
			\toprule
			\textbf{F(x)}\\
			\midrule
			BatchNorm\\
			LeakyReLU()\\
			Conv2D($1\times1$)\\
			BatchNorm\\
			LeakyRuLU()\\
			SubPixelConv2D($3\times3$, strides=2)\\
			BatchNorm\\
			LeakyReLU()\\
			Conv2D($1\times1$)\\
			\midrule
			\textbf{Shortcut Branch}\\
			\midrule
			SubPixelConv2D($1\times1$, strides=2)\\
			\bottomrule
		\end{tabular}
	\end{center}
	\caption{\textbf{UpResiBlock}. Dropout($0.1$) was added after the residual addition.}
	\label{table:up_resi_block}
\end{table}

\begin{table}[!htb]
	\begin{center}
		\begin{tabular}{c c c}
			\toprule
			\textbf{Layer} & \textbf{Filter} & \textbf{Output Size}\\
			\midrule
			Concat(Coord) & - & $3\times320\times320$\\
			ResiBlock & 8, 8, 32 &  $32\times320\times320$\\
			ResiBlock & 8, 8, 32 & $32\times320\times320$\\
			\midrule
			DownResiBlock & 16, 16, 64 & $64\times160\times160$\\
			ResiBlock & 16, 16, 64 & $64\times160\times160$\\
			ResiBlock & 16, 16, 64 & $64\times160\times160$\\
			\midrule
			DownResiBlock & 32, 32, 128 & $128\times80\times80$\\
			ResiBlock & 32, 32, 128 & $128\times80\times80$\\
			ResiBlock & 32, 32, 128 & $128\times80\times80$\\
			\midrule
			DownResiBlock & 64, 64, 256 & $256\times40\times40$\\
			ResiBlock & 64, 64, 256 & $256\times40\times40$\\
			ResiBlock & 64, 64, 256 & $256\times40\times40$\\
			\midrule
			DownResiBlock & 64, 64, 256 & $256\times20\times20$\\
			ResiBlock & 64, 64, 256 & $256\times20\times20$\\
			ResiBlock & 64, 64, 256 & $256\times20\times20$\\
			\midrule
			DownResiBlock & 128, 128, 512 & $512\times10\times10$\\
			ResiBlock & 128, 128, 512 & $512\times10\times10$\\
			ResiBlock & 128, 128, 512 & $512\times10\times10$\\
			\bottomrule
		\end{tabular}
	\end{center}
	\caption{ \textbf{ShapeNet}}
	\label{table:G_encoder}
\end{table}

\begin{table}[!htb]
	\begin{center}
		\begin{tabular}{c c c}
			\toprule
			\textbf{Layer} & \textbf{Filter} & \textbf{Output Size}\\
			\midrule
			Concat(Pos, Coord) & - & $6\times320\times320$\\
			DownResiBlock & 8, 8, 32 & $32\times160\times160$\\
			ResiBlock & 8, 8, 32 & $32\times160\times160$\\
			\midrule
			DownResiBlock & 16, 16, 64 & $64\times80\times80$\\
			ResiBlock & 16, 16, 64 & $64\times80\times80$\\
			\midrule
			DownResiBlock & 32, 32, 128 & $128\times40\times40$\\
			ResiBlock & 32, 32, 128 & $128\times40\times40$\\
			SelfAttention & - & $128\times40\times40$\\
			\midrule
			DownResiBlock & 64, 64, 256 & $256\times20\times20$\\
			ResiBlock & 64, 64, 256 & $256\times20\times20$\\
			SelfAttention & - & $256\times20\times20$\\
			\midrule
			DownResiBlock & 128, 128, 512 & $512\times10\times10$\\
			ResiBlock & 128, 128, 512 & $512\times10\times10$\\
			\midrule
			GlobalAvgPool() & - & 512\\
			Dropout(0.3) & - & 512\\
			Linear & 256 & 256\\
			Linear & 1 & 1\\
			Sigmoid() & - & 1\\
			\bottomrule
		\end{tabular}
	\end{center}
	\caption{\textbf{Discriminator}}
	\label{table:D}
\end{table}

\textbf{}

\begin{table}[!htb]
	\begin{center}
		\begin{tabular}{c c c}
			\toprule
			\textbf{Layer} & \textbf{Filter} & \textbf{Output Size}\\
			\midrule
			Concat(Coord) & - & $514\times10\times10$  \\
			FiLMResiBlock(x, e) & 512 & $512\times10\times10$ \\
			ResiBlock & 128, 128, 512 & $512\times10\times10$\\
			ResiBlock & 128, 128, 512 & $512\times10\times10$\\
			ResiBlock & 128, 128, 512 & $512\times10\times10$\\
			ResiBlock & 128, 128, 512 & $512\times10\times10$\\
			SelfAttention & - & $512\times10\times10$\\
			\midrule
			UpResiBlock & 64, 64, 256 & $256\times20\times20$\\
			Concat(SE(x, s), Coord) & - & $514\times20\times20$\\
			FiLMResiBlock(x, e) & 256 & $256\times20\times20$\\
			ResiBlock & 64, 64, 256 & $256\times20\times20$\\
			ResiBlock & 64, 64, 256 & $256\times20\times20$\\
			SelfAttention & - & $256\times20\times20$\\
			\midrule
			
			UpResiBlock & 64, 64, 256 & $256\times40\times40$\\
			Concat(SE(x, s), Coord) & - & $514\times40\times40$\\
			FiLMResiBlock(x, e) & 256 & $256\times40\times40$\\
			ResiBlock & 64, 64, 256 & $256\times40\times40$\\
			ResiBlock & 64, 64, 256 & $256\times40\times40$\\
			SelfAttention & - & $256\times40\times40$\\
			\midrule
			
			UpResiBlock & 32, 32, 128 & $128\times80\times80$\\
			Concat(SE(x, s), Coord) & - & $258\times80\times80$\\
			FiLMResiBlock(x, e) & 128 & $128\times80\times80$\\
			ResiBlock & 32, 32, 128 & $128\times80\times80$\\
			ResiBlock & 32, 32, 128 & $128\times80\times80$\\
			SelfAttention & - & $128\times80\times80$\\
			\midrule
			
			UpResiBlock & 16, 16, 64 & $64\times160\times160$\\
			Concat(SE(x, s), Coord) & - & $130\times160\times160$\\
			FiLMResiBlock(x, e) & 64 & $64\times160\times160$\\
			ResiBlock & 16, 16, 64 & $64\times160\times160$\\
			ResiBlock & 16, 16, 64 & $64\times160\times160$\\
			SelfAttention & - & $64\times160\times160$\\
			\midrule
			
			UpResiBlock & 8, 8, 32 & $32\times320\times320$\\
			Concat(SE(x, s), Coord) & - & $66\times320\times320$ \\
			FiLMResiBlock(x, e) & 32 & $32\times320\times320$\\
			ResiBlock & 8, 8, 32 & $32\times320\times320$\\
			ResiBlock & 8, 8, 32 & $32\times320\times320$\\
			\midrule
			
			ResiBlock & 4, 4, 16 & $16\times320\times320$\\
			ResiBlock & 4, 4, 16 & $16\times320\times320$\\
			ResiBlock & 4, 4, 16 & $16\times320\times320$\\
			
			Conv2D($1\times1$) & 1 & $1\times320\times320$\\
			Tanh()& - & $1\times320\times320$\\
			
			\bottomrule
		\end{tabular}
	\end{center}
	\caption{\textbf{RenderNet}. `e' is lighting position embedding. `s' is skip connection from ShapeNet.}
	\label{table:G_decoder}
\end{table}

\section{Compositing sketches and shadows}
Our result images ($I$) are composited by a simple weighted sum of the output shadow ($S$) and original line drawing ($L$)
\begin{equation} 
I=0.2S+0.8L, 
\end{equation}
where all images are grayscale in $[0,1]$.

In our training process, both line drawings and shadows are processed.  The line drawings are inverted, $L^\prime = 1-L$, to achieve white lines on a black background. The shadow images are inverted, then scaled and shifted to the interval $[-1,1]$, $S^\prime= (1-S)\times 2 - 1$.  The inverse transform is applied to output from the Generator before compositing as described above.

Additionally, simply concatenating the line drawing and shadow for input to the Discriminator produced poor results; instead we composite these images with another weighted sum,\begin{equation} 
I^\prime=L^\prime+0.25(S^\prime+1). 
\end{equation}This compositing is applied to both ground truth and Generator shadows.

\section{More Results}

\begin{figure*}[h]
	\centering
	\begin{subfigure}{.3\columnwidth}
		\includegraphics[width=\columnwidth, height=\columnwidth]{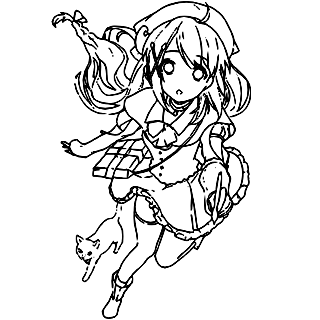}
		\subcaption*{Input}
	\end{subfigure}%
	\begin{subfigure}{.3\columnwidth}
		\includegraphics[width=\columnwidth, height=\columnwidth]{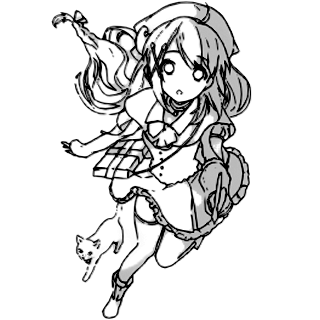}    	
		\subcaption*{Ours - 710}
	\end{subfigure}%
	\begin{subfigure}{.3\columnwidth}
		\includegraphics[width=\columnwidth, height=\columnwidth]{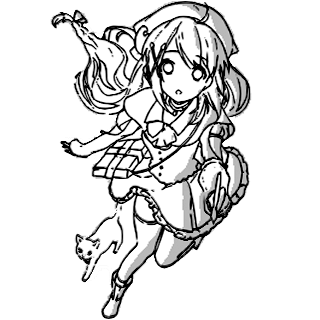}
		\subcaption*{DN - 710}
	\end{subfigure}%
	\begin{subfigure}{.3\columnwidth}
		\includegraphics[width=\columnwidth, height=\columnwidth]{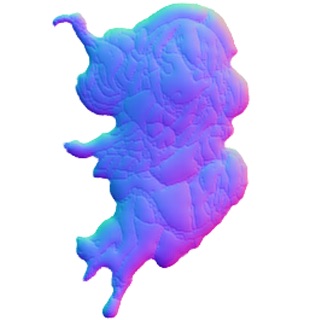}
		\subcaption*{DN - normal}
	\end{subfigure}%
	\begin{subfigure}{.3\columnwidth}
		\includegraphics[width=\columnwidth, height=\columnwidth]{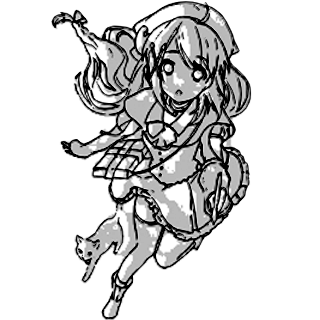}
		\subcaption*{S2N - 710}
	\end{subfigure}%
	\begin{subfigure}{.3\columnwidth}
		\includegraphics[width=\columnwidth, height=\columnwidth]{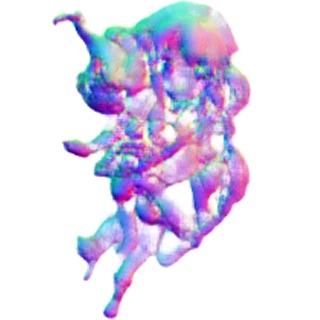}
		\subcaption*{S2N - normal}
	\end{subfigure}
	
	\begin{subfigure}{.3\columnwidth}
		\includegraphics[width=\columnwidth, height=\columnwidth]{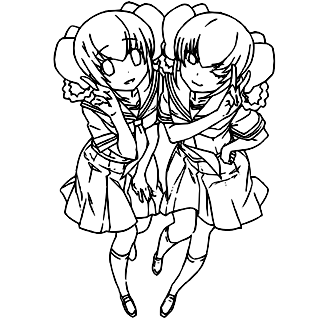}
		\subcaption*{Input}
	\end{subfigure}%
	\begin{subfigure}{.3\columnwidth}
		\includegraphics[width=\columnwidth, height=\columnwidth]{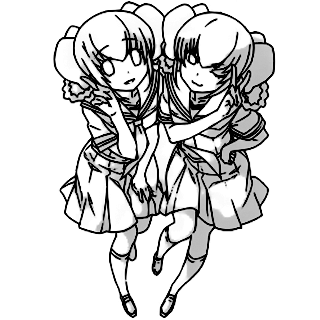}
		\subcaption*{Ours - 810}
	\end{subfigure}%
	\begin{subfigure}{.3\columnwidth}
		\includegraphics[width=\columnwidth, height=\columnwidth]{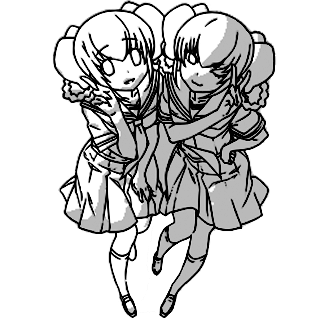}
		\subcaption*{DN - 810}
	\end{subfigure}%
	\begin{subfigure}{.3\columnwidth}
		\includegraphics[width=\columnwidth, height=\columnwidth]{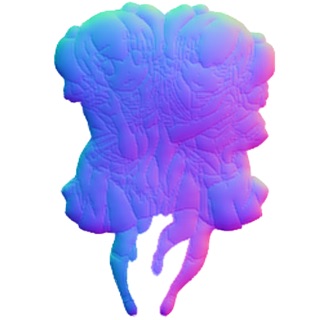}
		\subcaption*{DN - normal}
	\end{subfigure}%
	\begin{subfigure}{.3\columnwidth}
		\includegraphics[width=\columnwidth, height=\columnwidth]{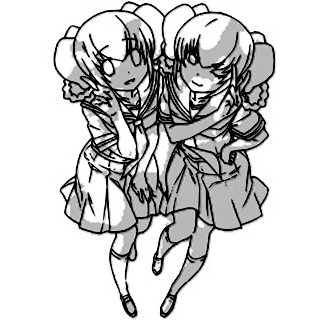}
		\subcaption*{S2N - 810}
	\end{subfigure}%
	\begin{subfigure}{.3\columnwidth}
		\includegraphics[width=\columnwidth, height=\columnwidth]{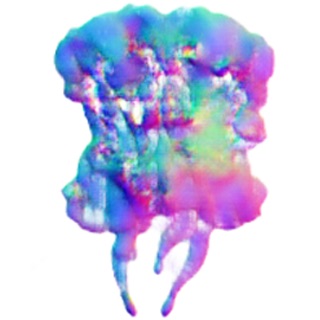}
		\subcaption*{S2N - normal}
	\end{subfigure}
	
	\begin{subfigure}{.3\columnwidth}
		\includegraphics[width=\columnwidth, height=\columnwidth]{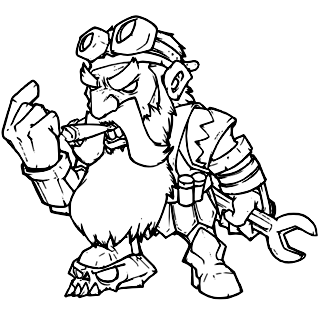}
		\subcaption*{Input}
	\end{subfigure}%
	\begin{subfigure}{.3\columnwidth}
		\includegraphics[width=\columnwidth, height=\columnwidth]{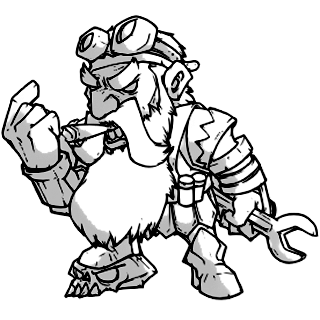}
		\subcaption*{Ours - 210}
	\end{subfigure}%
	\begin{subfigure}{.3\columnwidth}
		\includegraphics[width=\columnwidth, height=\columnwidth]{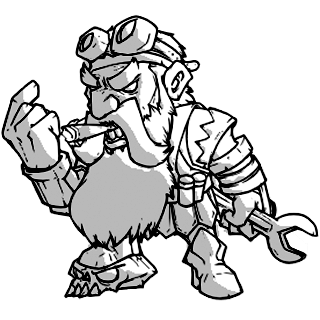}
		\subcaption*{DN - 210}
	\end{subfigure}%
	\begin{subfigure}{.3\columnwidth}
		\includegraphics[width=\columnwidth, height=\columnwidth]{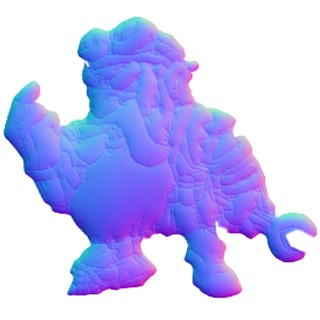}
		\subcaption*{DN - normal}
	\end{subfigure}%
	\begin{subfigure}{.3\columnwidth}
		\includegraphics[width=\columnwidth, height=\columnwidth]{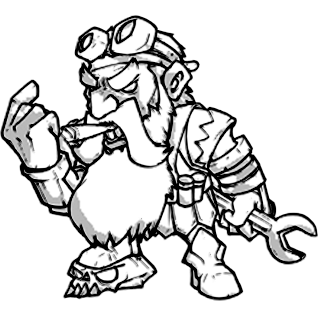}
		\subcaption*{S2N - 210}
	\end{subfigure}%
	\begin{subfigure}{.3\columnwidth}
		\includegraphics[width=\columnwidth, height=\columnwidth]{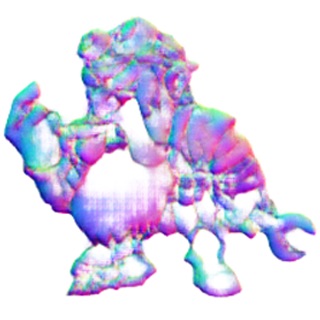}
		\subcaption*{S2N - normal}
	\end{subfigure}
	
	\begin{subfigure}{.3\columnwidth}
		\includegraphics[width=\columnwidth, height=\columnwidth]{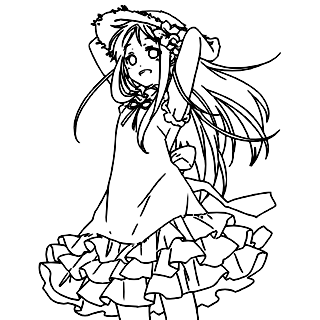}
		\subcaption*{Input}
	\end{subfigure}%
	\begin{subfigure}{.3\columnwidth}
		\includegraphics[width=\columnwidth, height=\columnwidth]{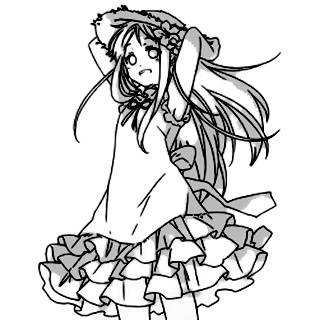}
		\subcaption*{Ours - 610}
	\end{subfigure}%
	\begin{subfigure}{.3\columnwidth}
		\includegraphics[width=\columnwidth, height=\columnwidth]{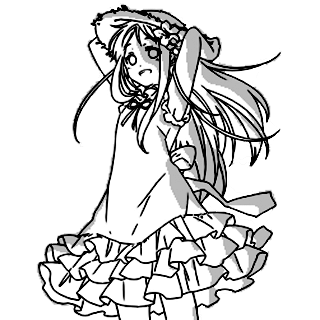}
		\subcaption*{DN - 610}
	\end{subfigure}%
	\begin{subfigure}{.3\columnwidth}
		\includegraphics[width=\columnwidth, height=\columnwidth]{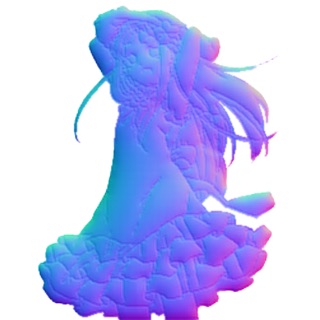}
		\subcaption*{DN - normal}
	\end{subfigure}%
	\begin{subfigure}{.3\columnwidth}
		\includegraphics[width=\columnwidth, height=\columnwidth]{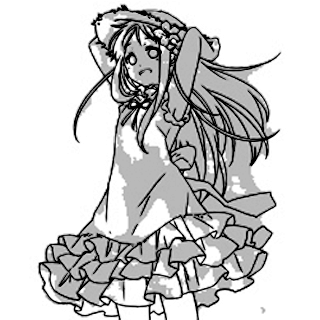}
		\subcaption*{S2N - 610}
	\end{subfigure}%
	\begin{subfigure}{.3\columnwidth}
		\includegraphics[width=\columnwidth, height=\columnwidth]{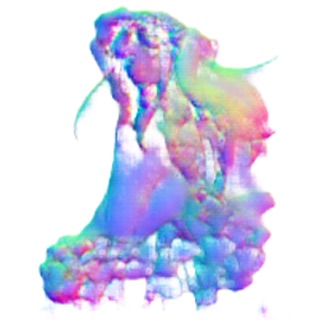}
		\subcaption*{S2N - normal}
	\end{subfigure}
	\caption{Comparisons with previous work DeepNormal (DN) \cite{hudon2018deep} and Sketch2Normal (S2N) \cite{su2018interactive} in front lighting. Zoom in the pictures in Figure~\ref{fig:compare1}. (Part 1)}
	\label{fig:compare1_normal}
\end{figure*}

\begin{figure*}[h]
	\centering
	\begin{subfigure}{.3\columnwidth}
		\includegraphics[width=\columnwidth, height=\columnwidth]{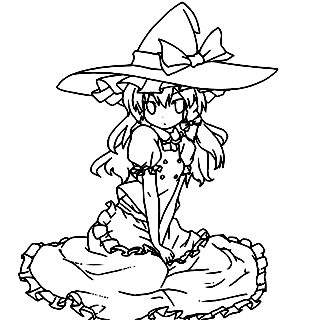}
		\subcaption*{Input}
	\end{subfigure}%
	\begin{subfigure}{.3\columnwidth}
		\includegraphics[width=\columnwidth, height=\columnwidth]{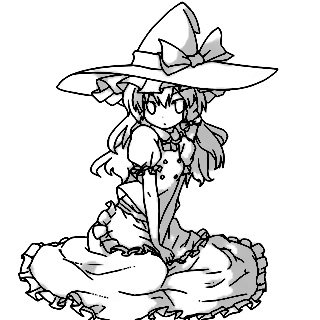}
		\subcaption*{Ours - 710}
	\end{subfigure}%
	\begin{subfigure}{.3\columnwidth}
		\includegraphics[width=\columnwidth, height=\columnwidth]{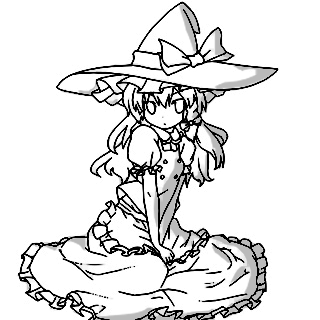}
		\subcaption*{DN - 710}
	\end{subfigure}%
	\begin{subfigure}{.3\columnwidth}
		\includegraphics[width=\columnwidth, height=\columnwidth]{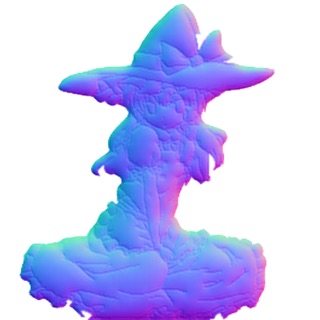}
		\subcaption*{DN - normal}
	\end{subfigure}%
	\begin{subfigure}{.3\columnwidth}
		\includegraphics[width=\columnwidth, height=\columnwidth]{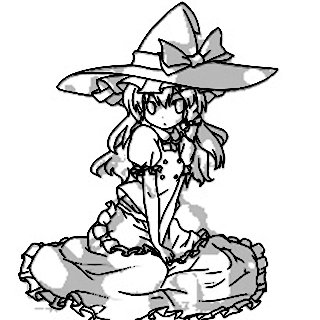}
		\subcaption*{S2N - 710}
	\end{subfigure}%
	\begin{subfigure}{.3\columnwidth}
		\includegraphics[width=\columnwidth, height=\columnwidth]{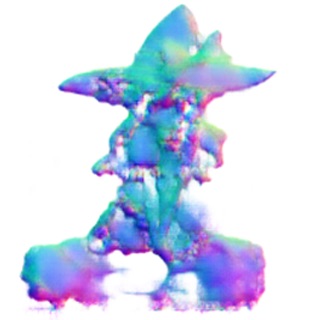}
		\subcaption*{S2N - normal}
	\end{subfigure}
	
	\begin{subfigure}{.3\columnwidth}
		\includegraphics[width=\columnwidth, height=\columnwidth]{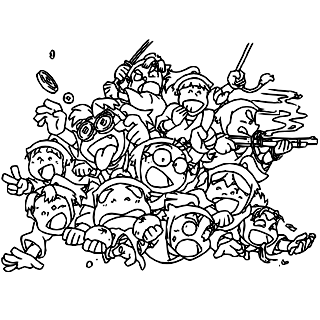}
		\subcaption*{Input}
	\end{subfigure}%
	\begin{subfigure}{.3\columnwidth}
		\includegraphics[width=\columnwidth, height=\columnwidth]{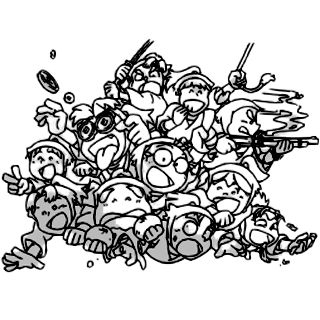}
		\subcaption*{Ours - 210}
	\end{subfigure}%
	\begin{subfigure}{.3\columnwidth}
		\includegraphics[width=\columnwidth, height=\columnwidth]{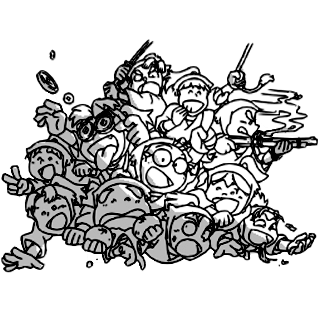}
		\subcaption*{DN - 210}
	\end{subfigure}%
	\begin{subfigure}{.3\columnwidth}
		\includegraphics[width=\columnwidth, height=\columnwidth]{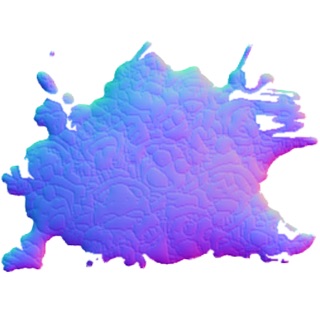}    	
		\subcaption*{DN - normal}
	\end{subfigure}%
	\begin{subfigure}{.3\columnwidth}
		\includegraphics[width=\columnwidth, height=\columnwidth]{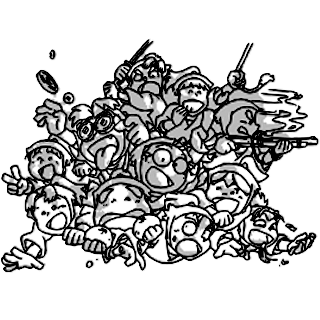}
		\subcaption*{S2N - 210}
	\end{subfigure}%
	\begin{subfigure}{.3\columnwidth}
		\includegraphics[width=\columnwidth, height=\columnwidth]{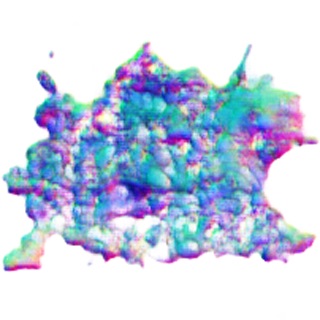}
		\subcaption*{S2N - normal}
	\end{subfigure}%
	
	\begin{subfigure}{.3\columnwidth}
		\includegraphics[width=\columnwidth, height=\columnwidth]{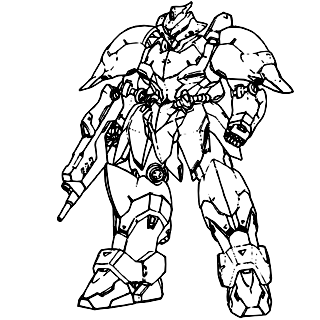}
		\subcaption*{Input}
	\end{subfigure}%
	\begin{subfigure}{.3\columnwidth}
		\includegraphics[width=\columnwidth, height=\columnwidth]{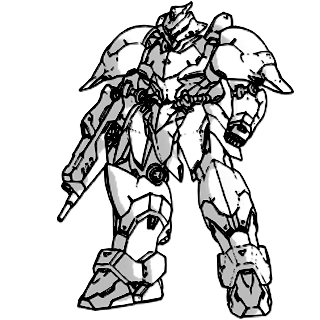}
		\subcaption*{Ours - 310}
	\end{subfigure}%
	\begin{subfigure}{.3\columnwidth}
		\includegraphics[width=\columnwidth, height=\columnwidth]{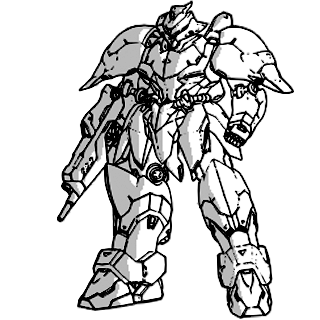}
		\subcaption*{DN - 310}
	\end{subfigure}%
	\begin{subfigure}{.3\columnwidth}
		\includegraphics[width=\columnwidth, height=\columnwidth]{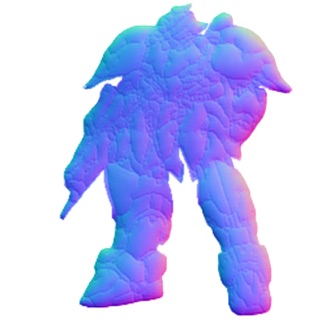}
		\subcaption*{DN - normal}
	\end{subfigure}%
	\begin{subfigure}{.3\columnwidth}
		\includegraphics[width=\columnwidth, height=\columnwidth]{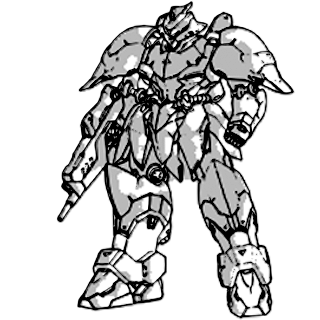}
		\subcaption*{S2N - 310}
	\end{subfigure}%
	\begin{subfigure}{.3\columnwidth}
		\includegraphics[width=\columnwidth, height=\columnwidth]{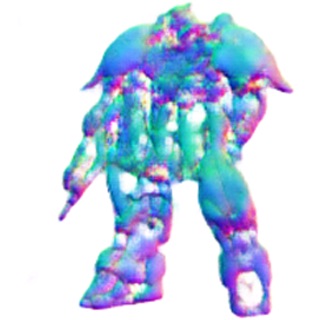}
		\subcaption*{S2N - normal}
	\end{subfigure}
	
	\begin{subfigure}{.3\columnwidth}
		\includegraphics[width=\columnwidth, height=\columnwidth]{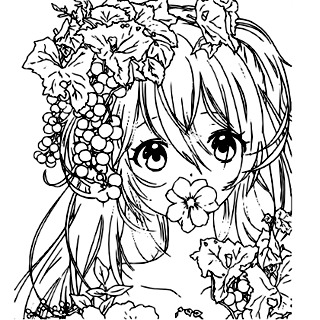}
		\subcaption*{Input}
	\end{subfigure}%
	\begin{subfigure}{.3\columnwidth}
		\includegraphics[width=\columnwidth, height=\columnwidth]{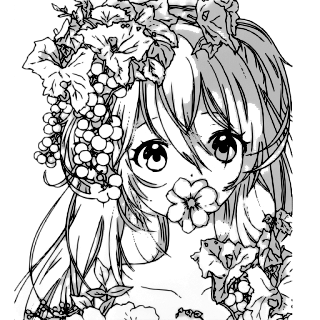}
		\subcaption*{Ours - 710}
	\end{subfigure}%
	\begin{subfigure}{.3\columnwidth}
		\includegraphics[width=\columnwidth, height=\columnwidth]{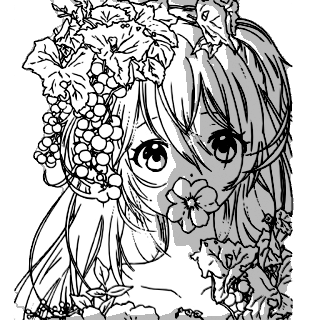}
		\subcaption*{DN - 710}
	\end{subfigure}%
	\begin{subfigure}{.3\columnwidth}
		\includegraphics[width=\columnwidth, height=\columnwidth]{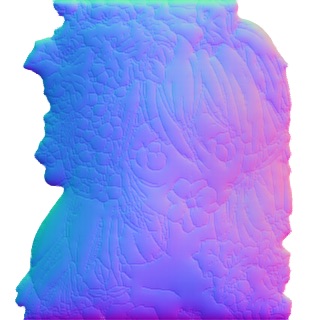}
		\subcaption*{DN - normal}
	\end{subfigure}%
	\begin{subfigure}{.3\columnwidth}
		\includegraphics[width=\columnwidth, height=\columnwidth]{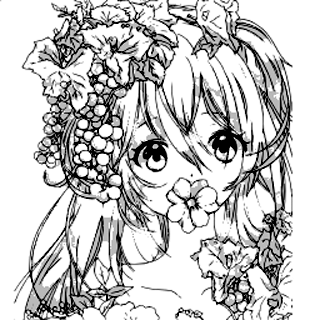}
		\subcaption*{S2N - 710}
	\end{subfigure}%
	\begin{subfigure}{.3\columnwidth}
		\includegraphics[width=\columnwidth, height=\columnwidth]{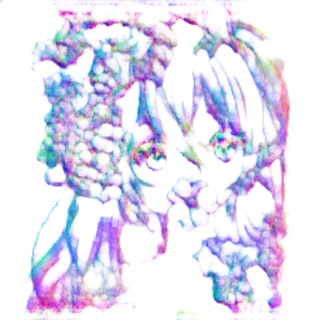}
		\subcaption*{S2N - normal}
	\end{subfigure}
	\caption{Comparisons with previous work DeepNormal (DN) \cite{hudon2018deep} and Sketch2Normal (S2N) \cite{su2018interactive} in front lighting. Zoom in the pictures in Figure~\ref{fig:compare1}. (Part 2)}
	\label{fig:compare2_normal}
\end{figure*}

\begin{figure*}[h]
	\centering
	\begin{subfigure}{.25\columnwidth}
		\includegraphics[width=\columnwidth, height=\columnwidth]{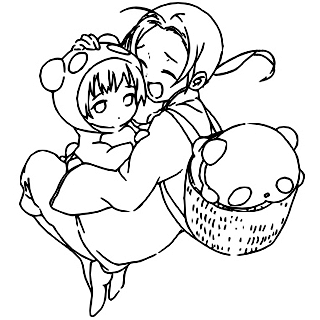}
		\subcaption*{Input}
	\end{subfigure}%
	\begin{subfigure}{.25\columnwidth}
		\includegraphics[width=\columnwidth, height=\columnwidth]{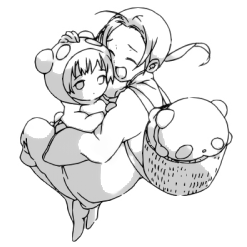}
		\subcaption*{Ours - 210}
	\end{subfigure}%
	\begin{subfigure}{.25\columnwidth}
		\includegraphics[width=\columnwidth, height=\columnwidth]{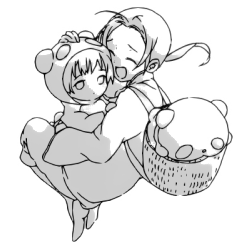}
		\subcaption*{Ours - 220}
	\end{subfigure}%
	\begin{subfigure}{.25\columnwidth}
		\includegraphics[width=\columnwidth, height=\columnwidth]{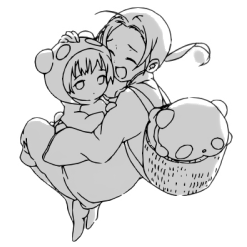}
		\subcaption*{Ours - 230}
	\end{subfigure}%
	\begin{subfigure}{.25\columnwidth}
		\includegraphics[width=\columnwidth, height=\columnwidth]{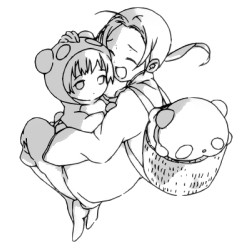}
		\subcaption*{Ours - 410}
	\end{subfigure}%
	\begin{subfigure}{.25\columnwidth}
		\includegraphics[width=\columnwidth, height=\columnwidth]{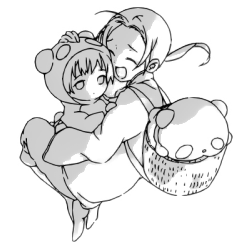}
		\subcaption*{Ours - 420}
	\end{subfigure}%
	\begin{subfigure}{.25\columnwidth}
		\includegraphics[width=\columnwidth, height=\columnwidth]{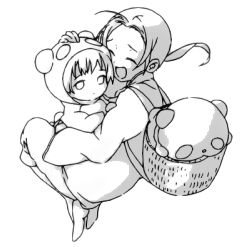}
		\subcaption*{Ours - 810}
	\end{subfigure}%
	\begin{subfigure}{.25\columnwidth}
		\includegraphics[width=\columnwidth, height=\columnwidth]{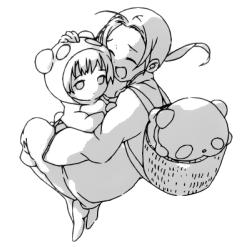}
		\subcaption*{Ours - 820}
	\end{subfigure}
	
	\begin{subfigure}{.25\columnwidth}
		\includegraphics[width=\columnwidth, height=\columnwidth]{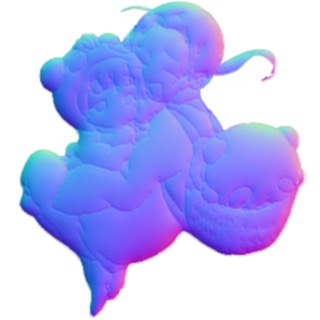}
		\subcaption*{Normal - \cite{hudon2018deep}}
	\end{subfigure}%
	\begin{subfigure}{.25\columnwidth}
		\includegraphics[width=\columnwidth, height=\columnwidth]{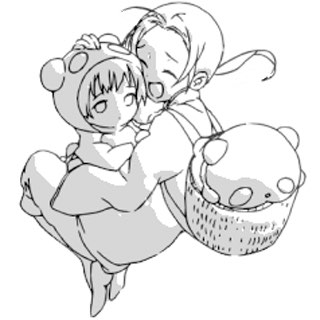}
		\subcaption*{DN - 210}
	\end{subfigure}%
	\begin{subfigure}{.25\columnwidth}
		\includegraphics[width=\columnwidth, height=\columnwidth]{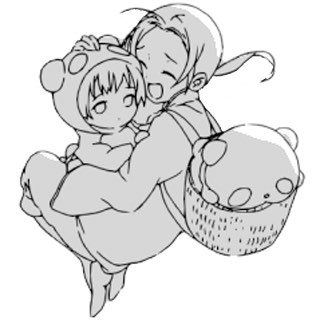}
		\subcaption*{DN - 220}
	\end{subfigure}%
	\begin{subfigure}{.25\columnwidth}
		\includegraphics[width=\columnwidth, height=\columnwidth]{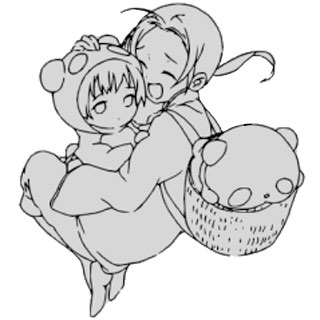}
		\subcaption*{DN - 230}
	\end{subfigure}%
	\begin{subfigure}{.25\columnwidth}
		\includegraphics[width=\columnwidth, height=\columnwidth]{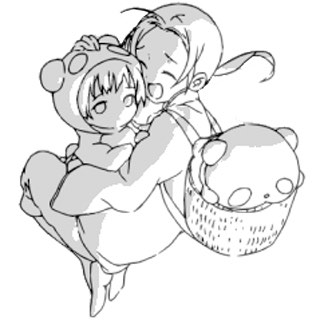}
		\subcaption*{DN - 410}
	\end{subfigure}%
	\begin{subfigure}{.25\columnwidth}
		\includegraphics[width=\columnwidth, height=\columnwidth]{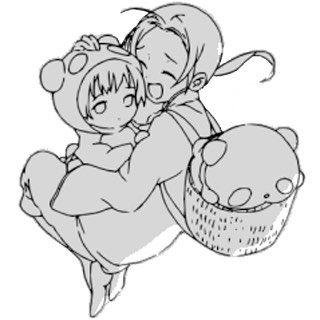}
		\subcaption*{DN - 420}
	\end{subfigure}%
	\begin{subfigure}{.25\columnwidth}
		\includegraphics[width=\columnwidth, height=\columnwidth]{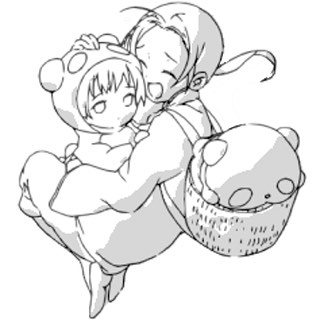}
		\subcaption*{DN - 810}
	\end{subfigure}%
	\begin{subfigure}{.25\columnwidth}
		\includegraphics[width=\columnwidth, height=\columnwidth]{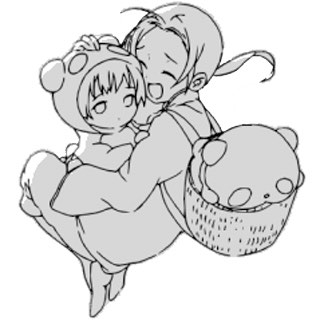}
		\subcaption*{DN - 820}
	\end{subfigure}
	
	\begin{subfigure}{.25\columnwidth}
		\includegraphics[width=\columnwidth, height=\columnwidth]{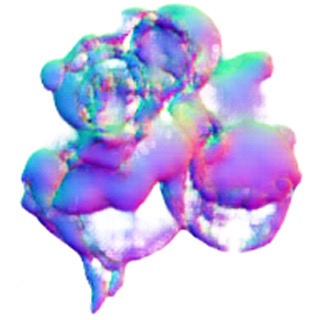}
		\subcaption*{Normal-\cite{su2018interactive}}
	\end{subfigure}%
	\begin{subfigure}{.25\columnwidth}
		\includegraphics[width=\columnwidth, height=\columnwidth]{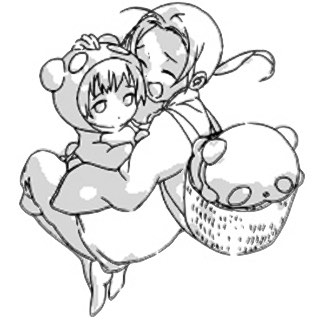}
		\subcaption*{S2N - 210}
	\end{subfigure}%
	\begin{subfigure}{.25\columnwidth}
		\includegraphics[width=\columnwidth, height=\columnwidth]{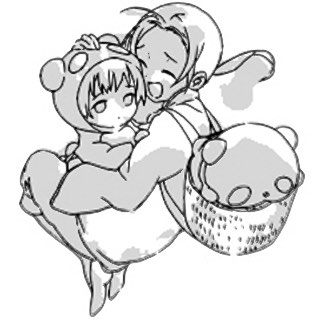}
		\subcaption*{S2N - 220}
	\end{subfigure}%
	\begin{subfigure}{.25\columnwidth}
		\includegraphics[width=\columnwidth, height=\columnwidth]{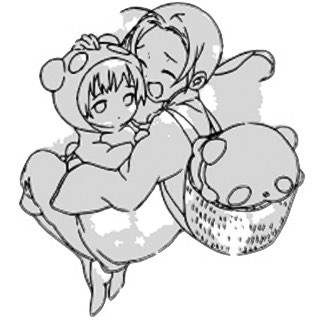}
		\subcaption*{S2N - 230}
	\end{subfigure}%
	\begin{subfigure}{.25\columnwidth}
		\includegraphics[width=\columnwidth, height=\columnwidth]{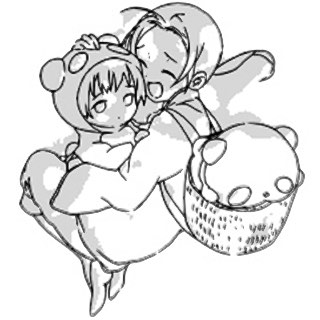}
		\subcaption*{S2N - 410}
	\end{subfigure}%
	\begin{subfigure}{.25\columnwidth}
		\includegraphics[width=\columnwidth, height=\columnwidth]{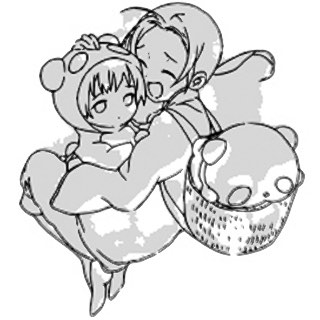}
		\subcaption*{S2N - 420}
	\end{subfigure}%
	\begin{subfigure}{.25\columnwidth}
		\includegraphics[width=\columnwidth, height=\columnwidth]{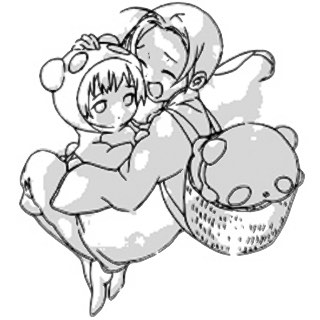}
		\subcaption*{S2N - 810}
	\end{subfigure}%
	\begin{subfigure}{.25\columnwidth}
		\includegraphics[width=\columnwidth, height=\columnwidth]{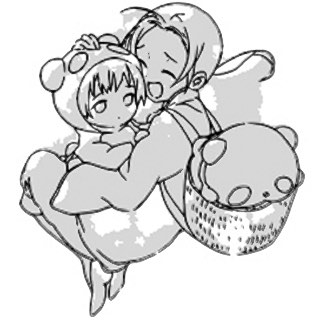}
		\subcaption*{S2N - 820}
	\end{subfigure}
	
	\caption{Comparisons with previous work DeepNormal (DN) \cite{hudon2018deep} and Sketch2Normal (S2N) \cite{su2018interactive} when the light source changes depth. First row is ours. The second row is DeepNormal's. The third row is Sketch2Normal's.}
	\label{fig:depthcompare2}
\end{figure*}

\begin{figure*}[h]
	\centering
	\begin{subfigure}{.3\columnwidth}
		\includegraphics[width=\columnwidth, height=\columnwidth]{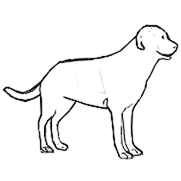}
		\subcaption*{Input}
	\end{subfigure}%
	\begin{subfigure}{.3\columnwidth}
		\includegraphics[width=\columnwidth, height=\columnwidth]{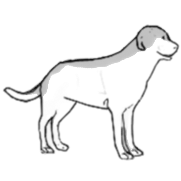}    	
		\subcaption*{Ours-510}
	\end{subfigure}%
	\begin{subfigure}{.3\columnwidth}
		\includegraphics[width=\columnwidth, height=\columnwidth]{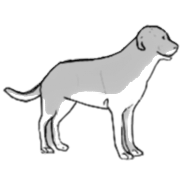}
		\subcaption*{Ours-520}
	\end{subfigure}%
	\begin{subfigure}{.3\columnwidth}
		\includegraphics[width=\columnwidth, height=\columnwidth]{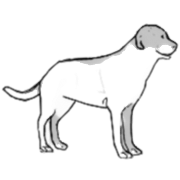}
		\subcaption*{Ours-610}
	\end{subfigure}%
	\begin{subfigure}{.3\columnwidth}
		\includegraphics[width=\columnwidth, height=\columnwidth]{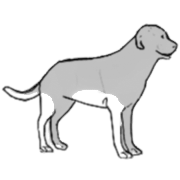}
		\subcaption*{Ours-620}
	\end{subfigure}
	
	\begin{subfigure}{.3\columnwidth}
		\includegraphics[width=\columnwidth, height=\columnwidth]{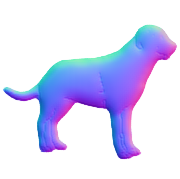}
		\subcaption*{DN-normal}
	\end{subfigure}%
	\begin{subfigure}{.3\columnwidth}
		\includegraphics[width=\columnwidth, height=\columnwidth]{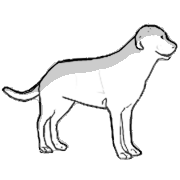}    	
		\subcaption*{DN-510}
	\end{subfigure}%
	\begin{subfigure}{.3\columnwidth}
		\includegraphics[width=\columnwidth, height=\columnwidth]{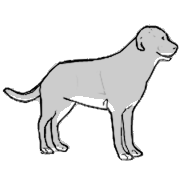}
		\subcaption*{DN-520}
	\end{subfigure}%
	\begin{subfigure}{.3\columnwidth}
		\includegraphics[width=\columnwidth, height=\columnwidth]{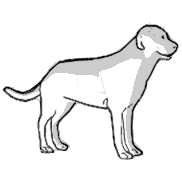}
		\subcaption*{DN-610}
	\end{subfigure}%
	\begin{subfigure}{.3\columnwidth}
		\includegraphics[width=\columnwidth, height=\columnwidth]{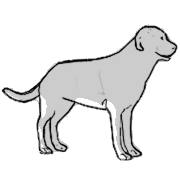}
		\subcaption*{DN-620}
	\end{subfigure}
	
	\begin{subfigure}{.3\columnwidth}
		\includegraphics[width=\columnwidth, height=\columnwidth]{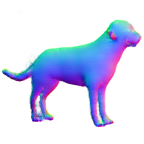}
		\subcaption*{S2N-normal}
	\end{subfigure}%
	\begin{subfigure}{.3\columnwidth}
		\includegraphics[width=\columnwidth, height=\columnwidth]{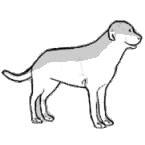}    	
		\subcaption*{S2N-510}
	\end{subfigure}%
	\begin{subfigure}{.3\columnwidth}
		\includegraphics[width=\columnwidth, height=\columnwidth]{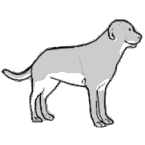}
		\subcaption*{S2N-520}
	\end{subfigure}%
	\begin{subfigure}{.3\columnwidth}
		\includegraphics[width=\columnwidth, height=\columnwidth]{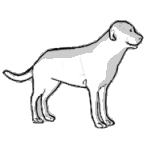}
		\subcaption*{S2N-610}
	\end{subfigure}%
	\begin{subfigure}{.3\columnwidth}
		\includegraphics[width=\columnwidth, height=\columnwidth]{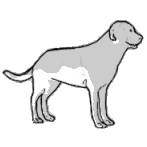}
		\subcaption*{S2N-620}
	\end{subfigure}
	
	\begin{subfigure}{.3\columnwidth}
		\includegraphics[width=\columnwidth, height=\columnwidth]{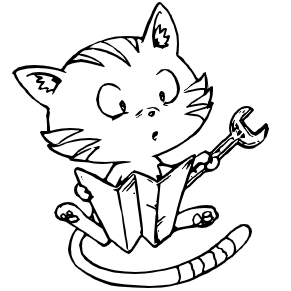}
		\subcaption*{Input}
	\end{subfigure}%
	\begin{subfigure}{.3\columnwidth}
		\includegraphics[width=\columnwidth, height=\columnwidth]{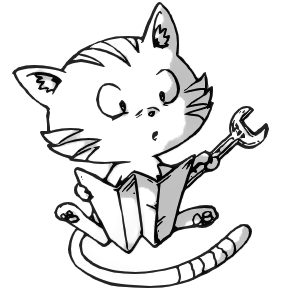}    	
		\subcaption*{Ours-710}
	\end{subfigure}%
	\begin{subfigure}{.3\columnwidth}
		\includegraphics[width=\columnwidth, height=\columnwidth]{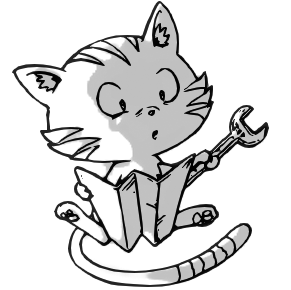}
		\subcaption*{Ours-720}
	\end{subfigure}%
	\begin{subfigure}{.3\columnwidth}
		\includegraphics[width=\columnwidth, height=\columnwidth]{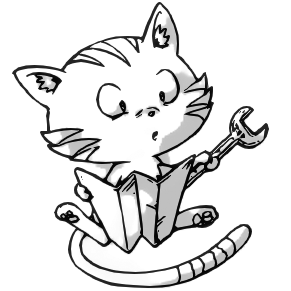}
		\subcaption*{Ours-810}
	\end{subfigure}%
	\begin{subfigure}{.3\columnwidth}
		\includegraphics[width=\columnwidth, height=\columnwidth]{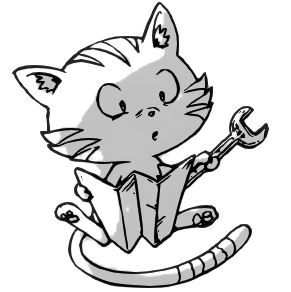}
		\subcaption*{Ours-820}
	\end{subfigure}
	
	\begin{subfigure}{.3\columnwidth}
		\includegraphics[width=\columnwidth, height=\columnwidth]{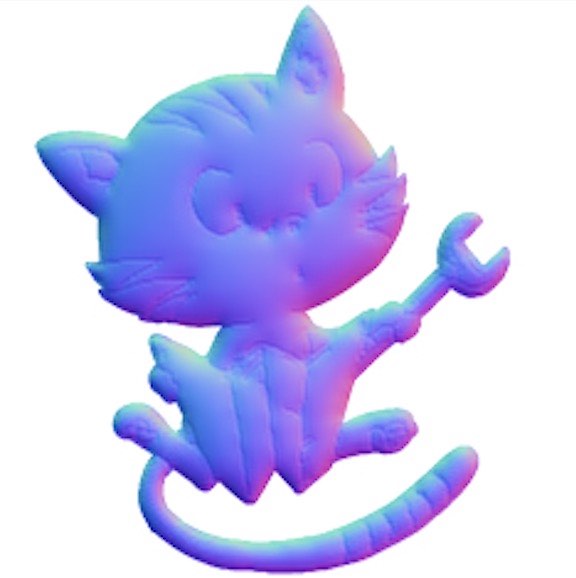}
		\subcaption*{DN-normal}
	\end{subfigure}%
	\begin{subfigure}{.3\columnwidth}
		\includegraphics[width=\columnwidth, height=\columnwidth]{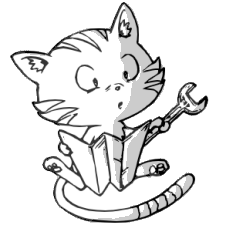}
		\subcaption*{DN-710}    	
	\end{subfigure}%
	\begin{subfigure}{.3\columnwidth}
		\includegraphics[width=\columnwidth, height=\columnwidth]{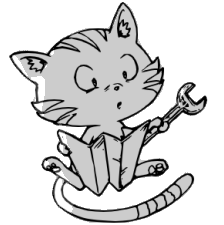}
		\subcaption*{DN-720}
	\end{subfigure}%
	\begin{subfigure}{.3\columnwidth}
		\includegraphics[width=\columnwidth, height=\columnwidth]{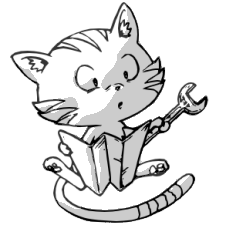}
		\subcaption*{DN-810}
	\end{subfigure}%
	\begin{subfigure}{.3\columnwidth}
		\includegraphics[width=\columnwidth, height=\columnwidth]{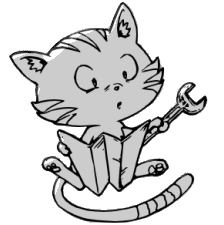}
		\subcaption*{DN-820}
	\end{subfigure}
	
	\begin{subfigure}{.3\columnwidth}
		\includegraphics[width=\columnwidth, height=\columnwidth]{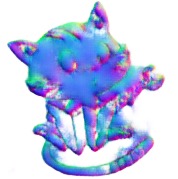}
		\subcaption*{S2N-normal}
	\end{subfigure}%
	\begin{subfigure}{.3\columnwidth}
		\includegraphics[width=\columnwidth, height=\columnwidth]{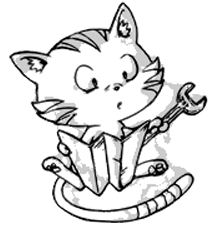}    	
		\subcaption*{S2N-710}
	\end{subfigure}%
	\begin{subfigure}{.3\columnwidth}
		\includegraphics[width=\columnwidth, height=\columnwidth]{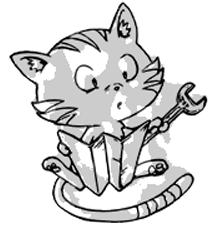}
		\subcaption*{S2N-720}
	\end{subfigure}%
	\begin{subfigure}{.3\columnwidth}
		\includegraphics[width=\columnwidth, height=\columnwidth]{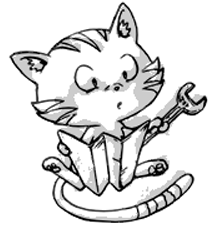}
		\subcaption*{S2N-810}
	\end{subfigure}%
	\begin{subfigure}{.3\columnwidth}
		\includegraphics[width=\columnwidth, height=\columnwidth]{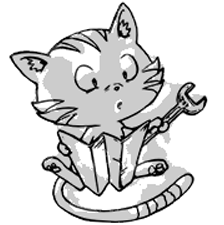}
		\subcaption*{S2N-820}
	\end{subfigure}
	
	\caption{Comparisons with previous work DeepNormal (DN) \cite{hudon2018deep} and Sketch2Normal (S2N) \cite{su2018interactive} using the line drawings from their papers. Dog image is from Sketch2Normal \cite{su2018interactive}. Cat image is from DeepNormal \cite{hudon2018deep}.}
	\label{fig:cat}
\end{figure*}

\begin{figure*}[h]
	\centering
	\begin{subfigure}{.35\columnwidth}
		\includegraphics[width=\columnwidth, height=\columnwidth]{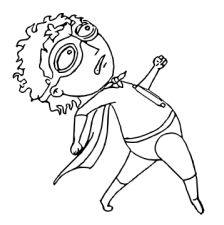}
		\subcaption*{Input}
	\end{subfigure}%
	\begin{subfigure}{.35\columnwidth}
		\includegraphics[width=\columnwidth, height=\columnwidth]{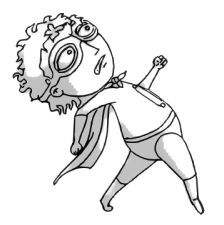}    	
		\subcaption*{Ours-210}
	\end{subfigure}%
	\begin{subfigure}{.35\columnwidth}
		\includegraphics[width=\columnwidth, height=\columnwidth]{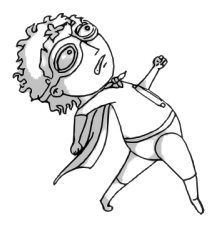}
		\subcaption*{Ours-310}
	\end{subfigure}%
	\begin{subfigure}{.35\columnwidth}
		\includegraphics[width=\columnwidth, height=\columnwidth]{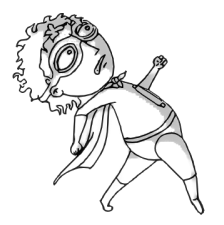}
		\subcaption*{Ours-610}
	\end{subfigure}%
	\begin{subfigure}{.35\columnwidth}
		\includegraphics[width=\columnwidth, height=\columnwidth]{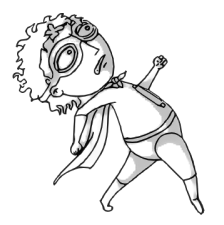}
		\subcaption*{Ours-710}
	\end{subfigure}
	
	\begin{subfigure}{.35\columnwidth}
		\includegraphics[width=\columnwidth, height=\columnwidth]{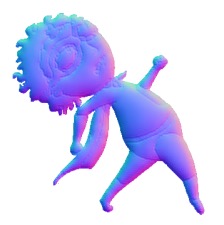}
		\subcaption*{DN-normal}
	\end{subfigure}%
	\begin{subfigure}{.35\columnwidth}
		\includegraphics[width=\columnwidth, height=\columnwidth]{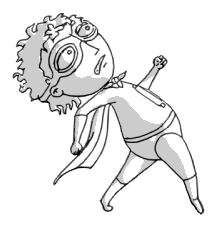}    	
		\subcaption*{DN-210}
	\end{subfigure}%
	\begin{subfigure}{.35\columnwidth}
		\includegraphics[width=\columnwidth, height=\columnwidth]{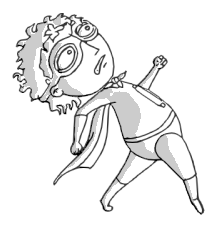}
		\subcaption*{DN-310}
	\end{subfigure}%
	\begin{subfigure}{.35\columnwidth}
		\includegraphics[width=\columnwidth, height=\columnwidth]{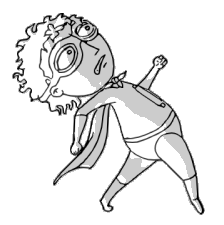}
		\subcaption*{DN-610}
	\end{subfigure}%
	\begin{subfigure}{.35\columnwidth}
		\includegraphics[width=\columnwidth, height=\columnwidth]{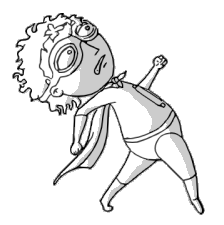}
		\subcaption*{DN-710}
	\end{subfigure}
	
	\begin{subfigure}{.35\columnwidth}
		\includegraphics[width=\columnwidth, height=\columnwidth]{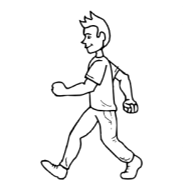}
		\subcaption*{Input}
	\end{subfigure}%
	\begin{subfigure}{.35\columnwidth}
		\includegraphics[width=\columnwidth, height=\columnwidth]{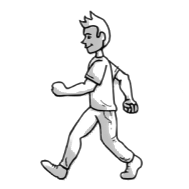}    	
		\subcaption*{Ours-210}
	\end{subfigure}%
	\begin{subfigure}{.35\columnwidth}
		\includegraphics[width=\columnwidth, height=\columnwidth]{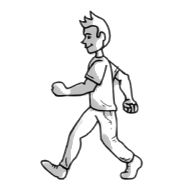}
		\subcaption*{Ours-310}
	\end{subfigure}%
	\begin{subfigure}{.35\columnwidth}
		\includegraphics[width=\columnwidth, height=\columnwidth]{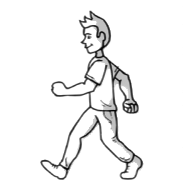}
		\subcaption*{Ours-610}
	\end{subfigure}%
	\begin{subfigure}{.35\columnwidth}
		\includegraphics[width=\columnwidth, height=\columnwidth]{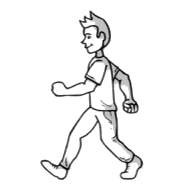}
		\subcaption*{Ours-710}
	\end{subfigure}
	
	\begin{subfigure}{.35\columnwidth}
		\includegraphics[width=\columnwidth, height=\columnwidth]{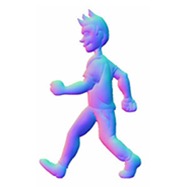}
		\subcaption*{DN-normal}
	\end{subfigure}%
	\begin{subfigure}{.35\columnwidth}
		\includegraphics[width=\columnwidth, height=\columnwidth]{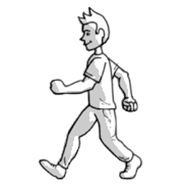}    	
		\subcaption*{DN-210}
	\end{subfigure}%
	\begin{subfigure}{.35\columnwidth}
		\includegraphics[width=\columnwidth, height=\columnwidth]{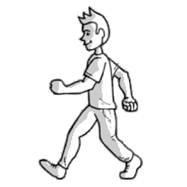}
		\subcaption*{DN-310}
	\end{subfigure}%
	\begin{subfigure}{.35\columnwidth}
		\includegraphics[width=\columnwidth, height=\columnwidth]{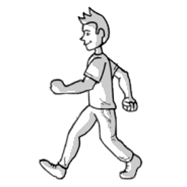}
		\subcaption*{DN-610}
	\end{subfigure}%
	\begin{subfigure}{.35\columnwidth}
		\includegraphics[width=\columnwidth, height=\columnwidth]{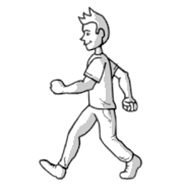}
		\subcaption*{DN-710}
	\end{subfigure}
	
	\caption{Comparisons with DeepNormal (DN) \cite{hudon2018deep} using the line drawings and normal maps in \cite{hudon2018deep}'s paper.}
	\label{human}
\end{figure*}

\begin{figure*}
	\centering
	\begin{subfigure}{.4\columnwidth}
		\includegraphics[width=\columnwidth, height=\columnwidth]{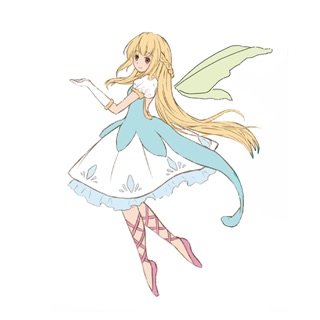}
		\subcaption{colorized}
	\end{subfigure}%
	\begin{subfigure}{.4\columnwidth}
		\includegraphics[width=\columnwidth, height=\columnwidth]{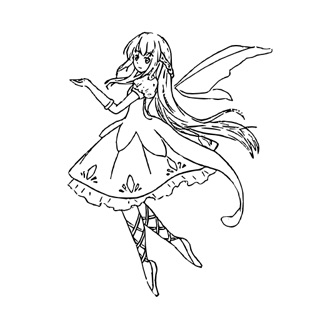}
		\subcaption{lines}
	\end{subfigure}%
	\begin{subfigure}{.4\columnwidth}
		\includegraphics[width=\columnwidth, height=\columnwidth]{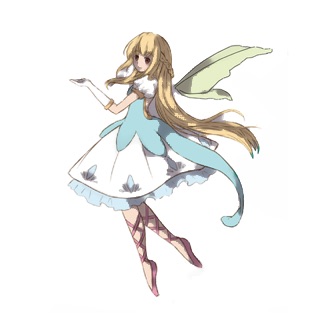}
		\subcaption{shadow - 610}
	\end{subfigure}%
	\begin{subfigure}{.4\columnwidth}
		\includegraphics[width=\columnwidth, height=\columnwidth]{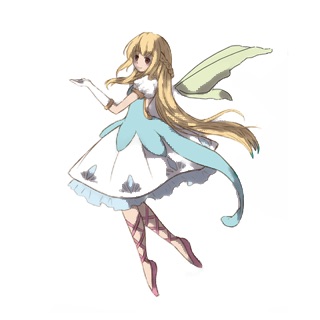}
		\subcaption{shadow - 710}
	\end{subfigure}%
	\begin{subfigure}{.4\columnwidth}
		\includegraphics[width=\columnwidth, height=\columnwidth]{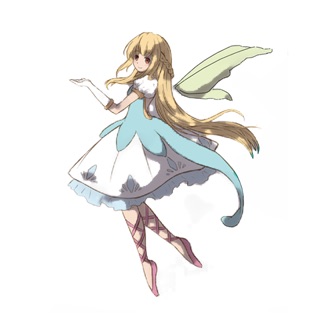}
		\subcaption{shadow - 810}
	\end{subfigure}
	
	\begin{subfigure}{.4\columnwidth}
		\includegraphics[width=\columnwidth, height=\columnwidth]{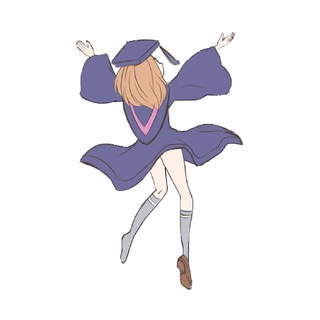}
		\subcaption{colorized}
	\end{subfigure}%
	\begin{subfigure}{.4\columnwidth}
		\includegraphics[width=\columnwidth, height=\columnwidth]{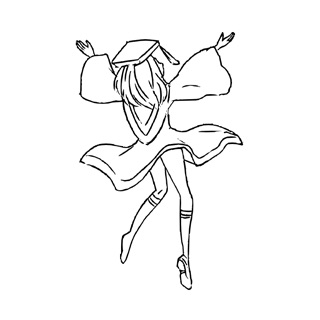}
		\subcaption{lines}
	\end{subfigure}%
	\begin{subfigure}{.4\columnwidth}
		\includegraphics[width=\columnwidth, height=\columnwidth]{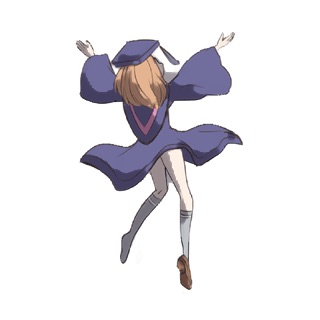}
		\subcaption{shadow - 210}
	\end{subfigure}%
	\begin{subfigure}{.4\columnwidth}
		\includegraphics[width=\columnwidth, height=\columnwidth]{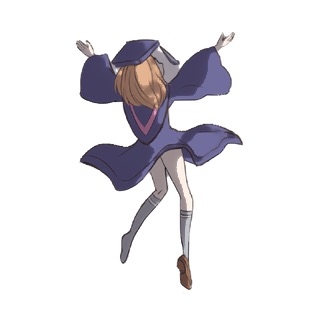}
		\subcaption{shadow - 320}
	\end{subfigure}%
	\begin{subfigure}{.4\columnwidth}
		\includegraphics[width=\columnwidth, height=\columnwidth]{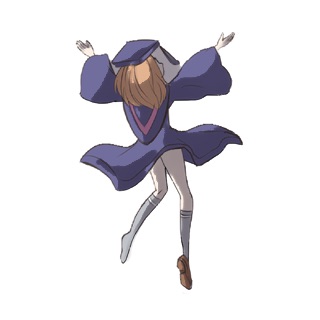}
		\subcaption{shadow - 820}
	\end{subfigure}
	
	\begin{subfigure}{.4\columnwidth}
		\includegraphics[width=\columnwidth, height=\columnwidth]{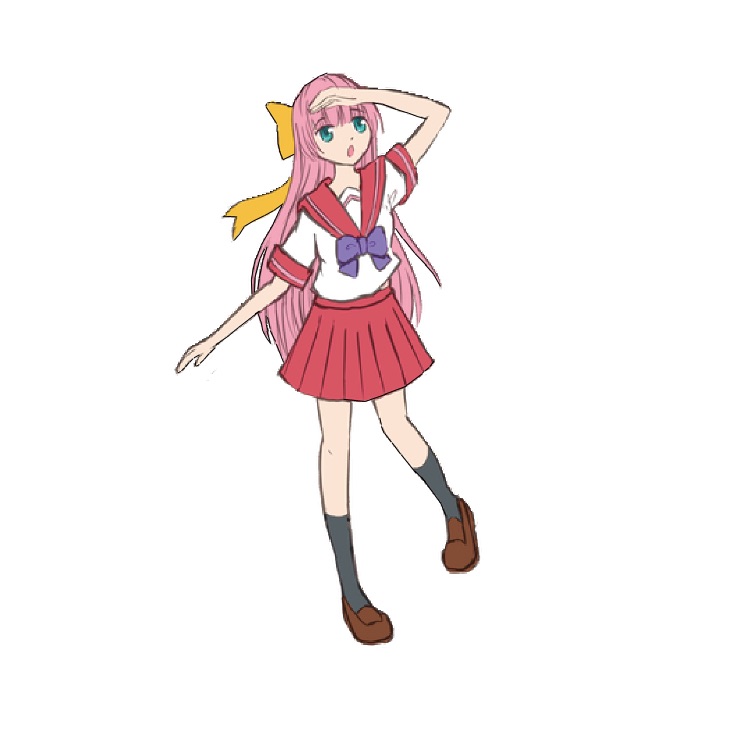}
		\subcaption{colorized}
	\end{subfigure}%
	\begin{subfigure}{.4\columnwidth}
		\includegraphics[width=\columnwidth, height=\columnwidth]{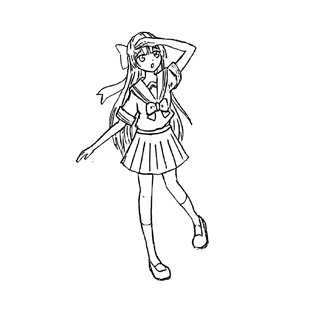}
		\subcaption{lines}
	\end{subfigure}%
	\begin{subfigure}{.4\columnwidth}
		\includegraphics[width=\columnwidth, height=\columnwidth]{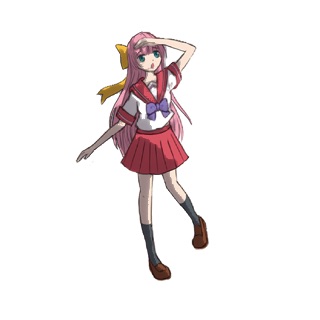}
		\subcaption{shadow - 220}
	\end{subfigure}%
	\begin{subfigure}{.4\columnwidth}
		\includegraphics[width=\columnwidth, height=\columnwidth]{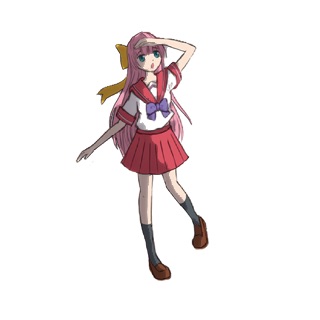}
		\subcaption{shadow - 320}
	\end{subfigure}%
	\begin{subfigure}{.4\columnwidth}
		\includegraphics[width=\columnwidth, height=\columnwidth]{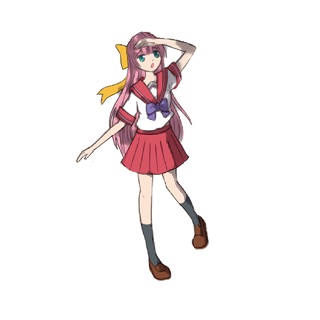}
		\subcaption{shadow - 710}
	\end{subfigure}
	
	\begin{subfigure}{.4\columnwidth}
		\includegraphics[width=\columnwidth, height=\columnwidth]{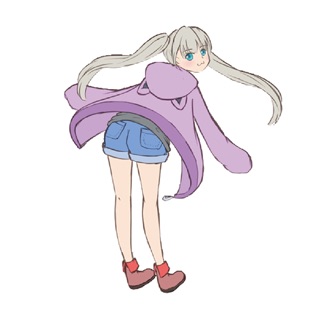}
		\subcaption{colorized}
	\end{subfigure}%
	\begin{subfigure}{.4\columnwidth}
		\includegraphics[width=\columnwidth, height=\columnwidth]{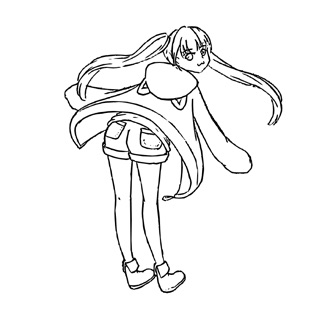}
		\subcaption{lines}
	\end{subfigure}%
	\begin{subfigure}{.4\columnwidth}
		\includegraphics[width=\columnwidth, height=\columnwidth]{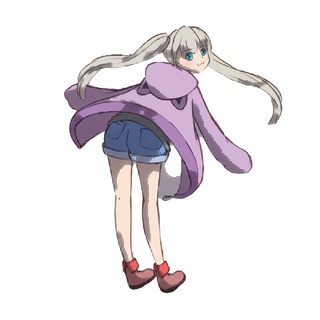}
		\subcaption{shadow - 210}
	\end{subfigure}%
	\begin{subfigure}{.4\columnwidth}
		\includegraphics[width=\columnwidth, height=\columnwidth]{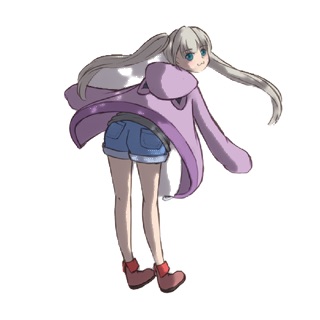}
		\subcaption{shadow - 320}
	\end{subfigure}%
	\begin{subfigure}{.4\columnwidth}
		\includegraphics[width=\columnwidth, height=\columnwidth]{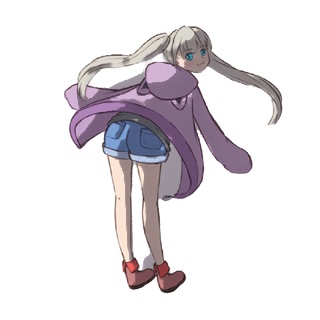}
		\subcaption{shadow - 820}
	\end{subfigure}
	
	\caption{Our results using \cite{gao2018illumination}'s images. \cite{gao2018illumination} solves similar problems, inputting colorized images to predict the normal maps midway then generate the binary shadows. The colorized images (a), (f), (k), (p) are from \cite{gao2018illumination}. (b), (g), (l), (q) are the lines that we subtract from the colorized images. Our system uses (b), (g), (l), (q) as the inputs to predict the pure shadows, then composite the shadows with the colorized images. For each line drawing, we show our results in three lighting directions. Please refer to \cite{gao2018illumination} for the their results in similar lighting directions.}
	\label{fig:nips}
\end{figure*}

\begin{figure*}[h]
	\centering
	\begin{subfigure}{.25\columnwidth}
		\includegraphics[width=\columnwidth, height=\columnwidth]{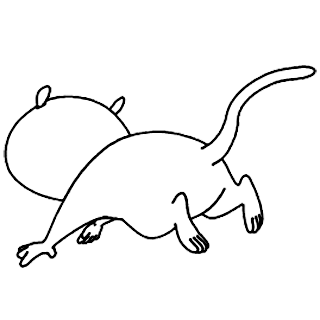}
		\subcaption*{Input}
	\end{subfigure}%
	\begin{subfigure}{.25\columnwidth}
		\includegraphics[width=\columnwidth, height=\columnwidth]{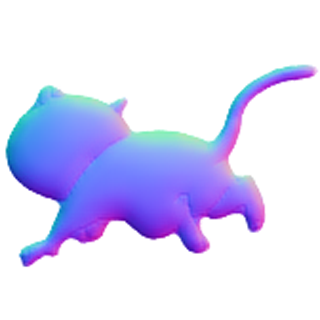}
		\subcaption*{DN-normal}
	\end{subfigure}%
	\begin{subfigure}{.25\columnwidth}
		\includegraphics[width=\columnwidth, height=\columnwidth]{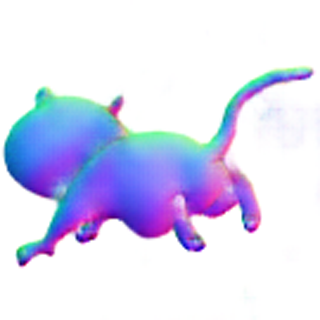}
		\subcaption*{S2N-normal}
	\end{subfigure}%
	\begin{subfigure}{.25\columnwidth}
		\includegraphics[width=\columnwidth, height=\columnwidth]{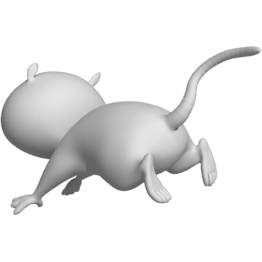}
		\subcaption*{3D model}
	\end{subfigure}%
	\begin{subfigure}{.25\columnwidth}
		\includegraphics[width=\columnwidth, height=\columnwidth]{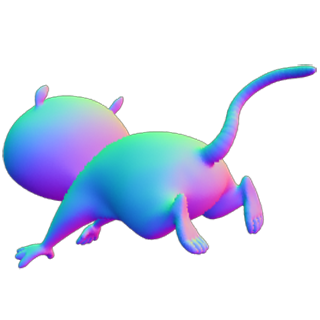}
		\subcaption*{GT-normal}
	\end{subfigure}
	
	\begin{subfigure}{.25\columnwidth}
		\includegraphics[width=\columnwidth, height=\columnwidth]{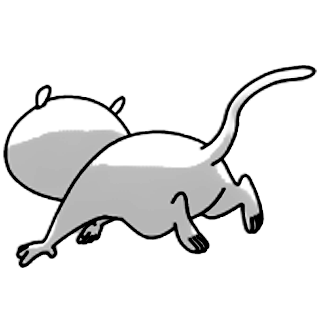}    	
		\subcaption*{Ours-120}
	\end{subfigure}%
	\begin{subfigure}{.25\columnwidth}
		\includegraphics[width=\columnwidth, height=\columnwidth]{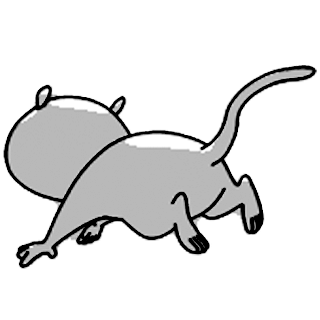}    	
		\subcaption*{DN-120}
	\end{subfigure}%
	\begin{subfigure}{.25\columnwidth}
		\includegraphics[width=\columnwidth, height=\columnwidth]{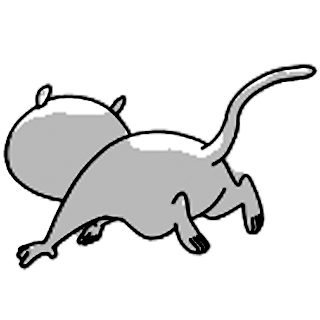}    	
		\subcaption*{S2N-120}
	\end{subfigure}%
	\begin{subfigure}{.25\columnwidth}
		\includegraphics[width=\columnwidth, height=\columnwidth]{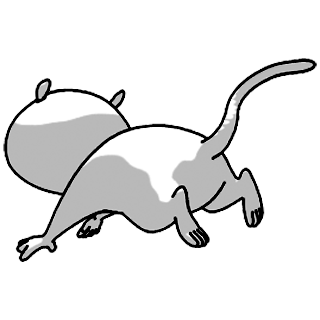}    	
		\subcaption*{$GT_{3D}$-120}
	\end{subfigure}%
	\begin{subfigure}{.25\columnwidth}
		\includegraphics[width=\columnwidth, height=\columnwidth]{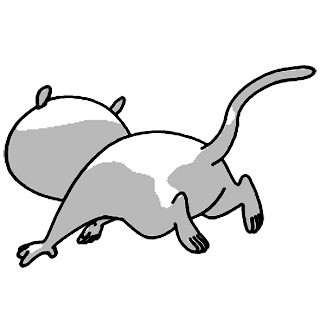}    	
		\subcaption*{GT-120}
	\end{subfigure}
	
	\begin{subfigure}{.25\columnwidth}
		\includegraphics[width=\columnwidth, height=\columnwidth]{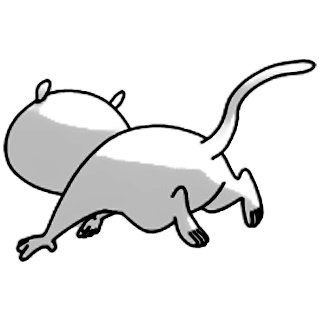}
		\subcaption*{Ours-220}
	\end{subfigure}%
	\begin{subfigure}{.25\columnwidth}
		\includegraphics[width=\columnwidth, height=\columnwidth]{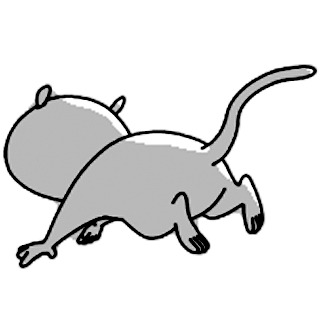}
		\subcaption*{DN-220}
	\end{subfigure}%
	\begin{subfigure}{.25\columnwidth}
		\includegraphics[width=\columnwidth, height=\columnwidth]{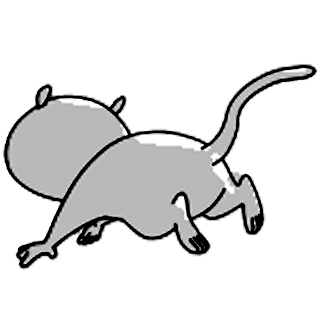}
		\subcaption*{S2N-220}
	\end{subfigure}%
	\begin{subfigure}{.25\columnwidth}
		\includegraphics[width=\columnwidth, height=\columnwidth]{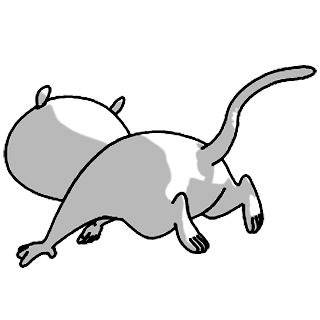}
		\subcaption*{$GT_{3D}$-220}
	\end{subfigure}%
	\begin{subfigure}{.25\columnwidth}
		\includegraphics[width=\columnwidth, height=\columnwidth]{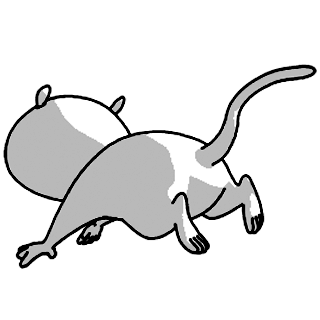}
		\subcaption*{GT-220}
	\end{subfigure}
	
	\caption{Comparisons on 3D test model \cite{goodwin2007isophote} with DeepNormal (DN) \cite{hudon2018deep}, Sketch2Normal (S2N) \cite{su2018interactive} and Ground Truth (GT). $GT_{3D}$: rendered directly from 3D model with commercial software. GT: rendered from its normal map. All of the normal maps, including ground truth, are rendered with the same settings as the paper (use the renderer scripts provided by DeepNormal and threshold the continuous shadings at $0.5$). Along with the bunny 3D test model in paper, our shadows are most close to the ground truth.}
	\label{fig:normal}
\end{figure*}

\begin{figure*}[h]
	\centering
	\includegraphics[width=2\columnwidth]{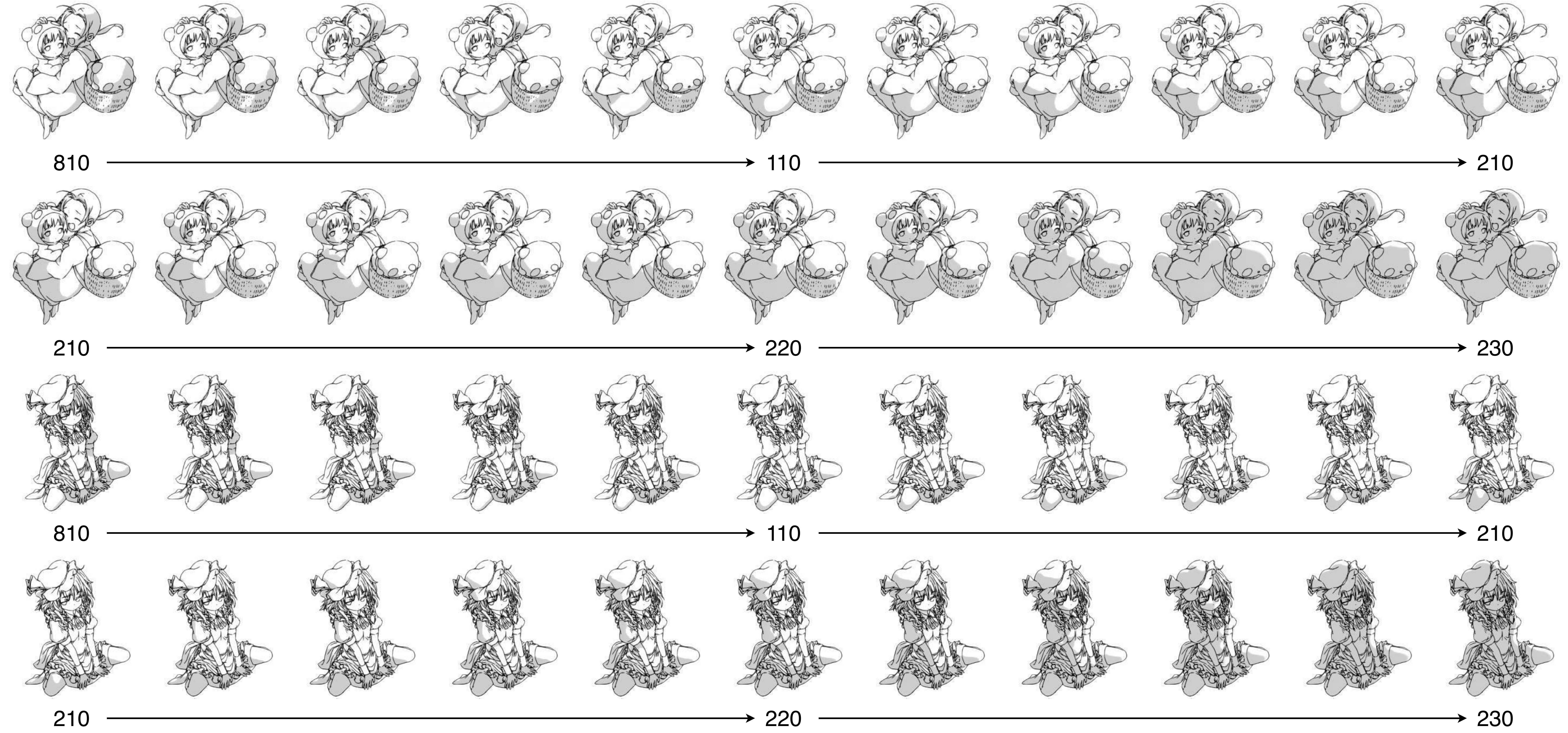}
	\caption{Examples of our continuous shadows between the discrete lighting source.}
	\label{contin}
\end{figure*}

\begin{figure*}[h]
	\centering
	\begin{subfigure}{.32\columnwidth}
		\includegraphics[width=\columnwidth, height=\columnwidth]{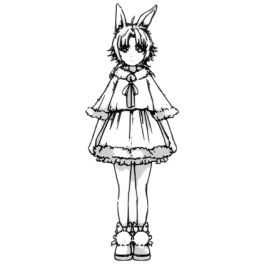}
		\subcaption*{110}
	\end{subfigure}%
	\begin{subfigure}{.32\columnwidth}
		\includegraphics[width=\columnwidth, height=\columnwidth]{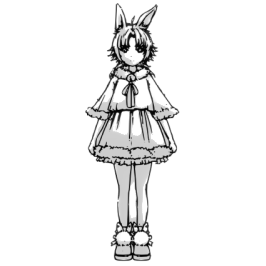}
		\subcaption*{120}
	\end{subfigure}%
	\begin{subfigure}{.32\columnwidth}
		\includegraphics[width=\columnwidth, height=\columnwidth]{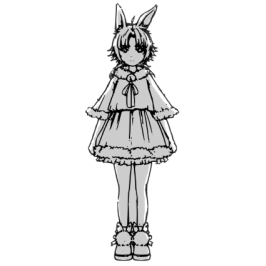}
		\subcaption*{130}
	\end{subfigure}%
	\begin{subfigure}{.32\columnwidth}
		\includegraphics[width=\columnwidth, height=\columnwidth]{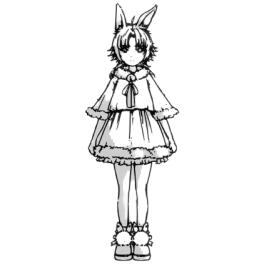}
		\subcaption*{210}
	\end{subfigure}%
	\begin{subfigure}{.32\columnwidth}
		\includegraphics[width=\columnwidth, height=\columnwidth]{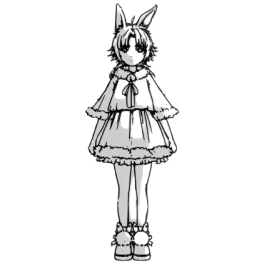}
		\subcaption*{220}
	\end{subfigure}%
	\begin{subfigure}{.32\columnwidth}
		\includegraphics[width=\columnwidth, height=\columnwidth]{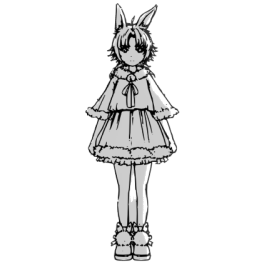}
		\subcaption*{230}
	\end{subfigure}
	
	\begin{subfigure}{.32\columnwidth}
		\includegraphics[width=\columnwidth, height=\columnwidth]{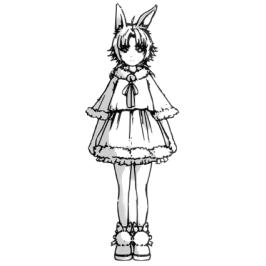}
		\subcaption*{310}
	\end{subfigure}%
	\begin{subfigure}{.32\columnwidth}
		\includegraphics[width=\columnwidth, height=\columnwidth]{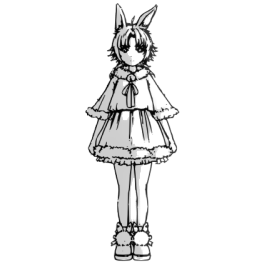}
		\subcaption*{320}
	\end{subfigure}%
	\begin{subfigure}{.32\columnwidth}
		\includegraphics[width=\columnwidth, height=\columnwidth]{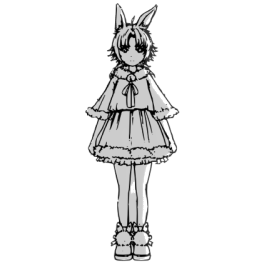}
		\subcaption*{330}
	\end{subfigure}%
	\begin{subfigure}{.32\columnwidth}
		\includegraphics[width=\columnwidth, height=\columnwidth]{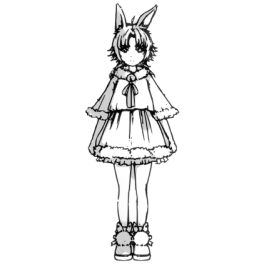}
		\subcaption*{410}
	\end{subfigure}%
	\begin{subfigure}{.32\columnwidth}
		\includegraphics[width=\columnwidth, height=\columnwidth]{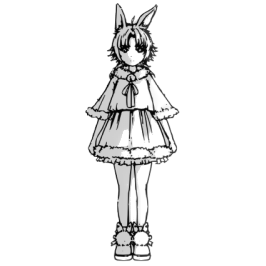}
		\subcaption*{420}
	\end{subfigure}%
	\begin{subfigure}{.32\columnwidth}
		\includegraphics[width=\columnwidth, height=\columnwidth]{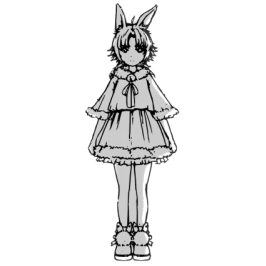}
		\subcaption*{430}
	\end{subfigure}
	
	\begin{subfigure}{.32\columnwidth}
		\includegraphics[width=\columnwidth, height=\columnwidth]{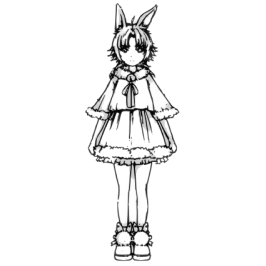}
		\subcaption*{510}
	\end{subfigure}%
	\begin{subfigure}{.32\columnwidth}
		\includegraphics[width=\columnwidth, height=\columnwidth]{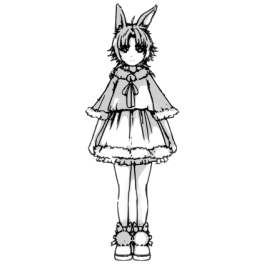}
		\subcaption*{520}
	\end{subfigure}%
	\begin{subfigure}{.32\columnwidth}
		\includegraphics[width=\columnwidth, height=\columnwidth]{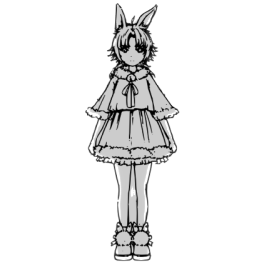}
		\subcaption*{530}
	\end{subfigure}%
	\begin{subfigure}{.32\columnwidth}
		\includegraphics[width=\columnwidth, height=\columnwidth]{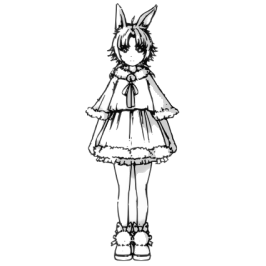}
		\subcaption*{610}
	\end{subfigure}%
	\begin{subfigure}{.32\columnwidth}
		\includegraphics[width=\columnwidth, height=\columnwidth]{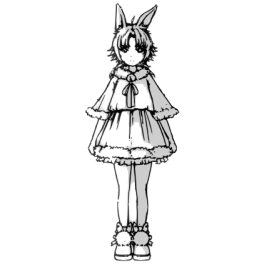}
		\subcaption*{620}
	\end{subfigure}%
	\begin{subfigure}{.32\columnwidth}
		\includegraphics[width=\columnwidth, height=\columnwidth]{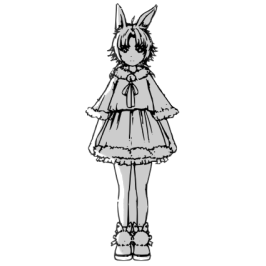}
		\subcaption*{630}
	\end{subfigure}
	
	\begin{subfigure}{.32\columnwidth}
		\includegraphics[width=\columnwidth, height=\columnwidth]{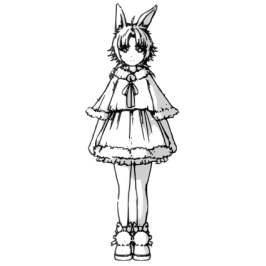}
		\subcaption*{710}
	\end{subfigure}%
	\begin{subfigure}{.32\columnwidth}
		\includegraphics[width=\columnwidth, height=\columnwidth]{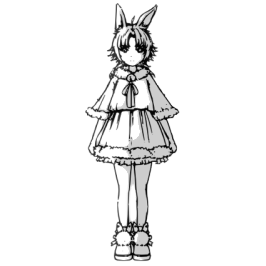}
		\subcaption*{720}
	\end{subfigure}%
	\begin{subfigure}{.32\columnwidth}
		\includegraphics[width=\columnwidth, height=\columnwidth]{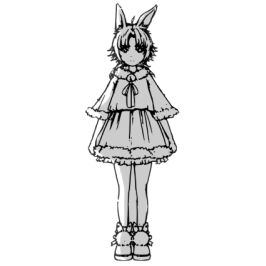}
		\subcaption*{730}
	\end{subfigure}%
	\begin{subfigure}{.32\columnwidth}
		\includegraphics[width=\columnwidth, height=\columnwidth]{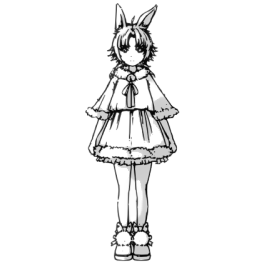}
		\subcaption*{810}
	\end{subfigure}%
	\begin{subfigure}{.32\columnwidth}
		\includegraphics[width=\columnwidth, height=\columnwidth]{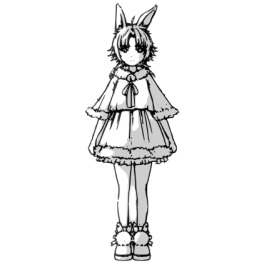}
		\subcaption*{820}
	\end{subfigure}%
	\begin{subfigure}{.32\columnwidth}
		\includegraphics[width=\columnwidth, height=\columnwidth]{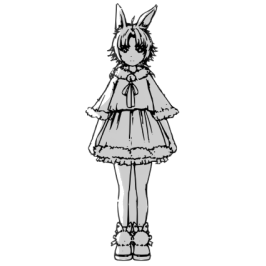}
		\subcaption*{830}
	\end{subfigure}
	
	\begin{subfigure}{.32\columnwidth}
		\includegraphics[width=\columnwidth, height=\columnwidth]{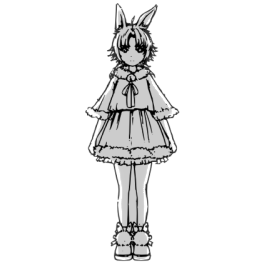}
		\subcaption*{001}
	\end{subfigure}%
	\begin{subfigure}{.32\columnwidth}
		\includegraphics[width=\columnwidth, height=\columnwidth]{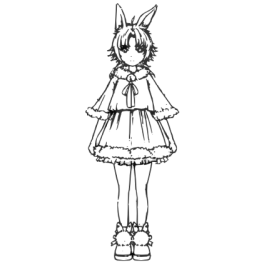}
		\subcaption*{002}
	\end{subfigure}
	
	\caption{Evaluations of our work in all 26 directions.}
	\label{fig:all1}
\end{figure*}

\begin{figure*}
	\centering
	\begin{subfigure}{.32\columnwidth}
		\includegraphics[width=\columnwidth, height=\columnwidth]{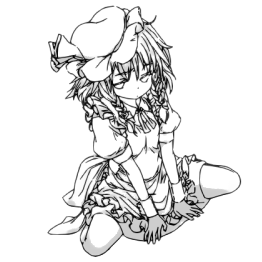}
		\subcaption*{110}
	\end{subfigure}%
	\begin{subfigure}{.32\columnwidth}
		\includegraphics[width=\columnwidth, height=\columnwidth]{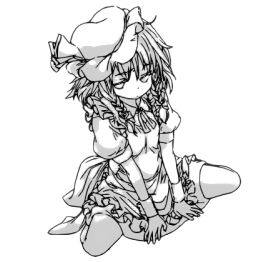}
		\subcaption*{120}
	\end{subfigure}%
	\begin{subfigure}{.32\columnwidth}
		\includegraphics[width=\columnwidth, height=\columnwidth]{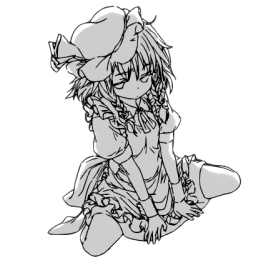}
		\subcaption*{130}
	\end{subfigure}%
	\begin{subfigure}{.32\columnwidth}
		\includegraphics[width=\columnwidth, height=\columnwidth]{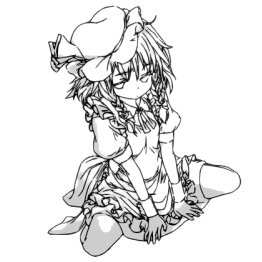}
		\subcaption*{210}
	\end{subfigure}%
	\begin{subfigure}{.32\columnwidth}
		\includegraphics[width=\columnwidth, height=\columnwidth]{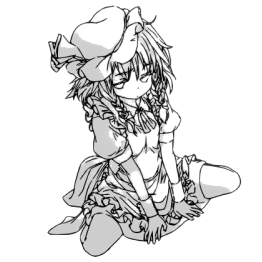}
		\subcaption*{220}
	\end{subfigure}%
	\begin{subfigure}{.32\columnwidth}
		\includegraphics[width=\columnwidth, height=\columnwidth]{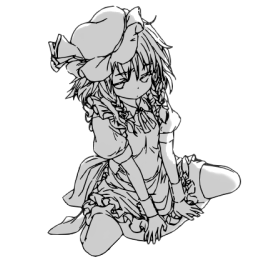}
		\subcaption*{230}
	\end{subfigure}
	
	\begin{subfigure}{.32\columnwidth}
		\includegraphics[width=\columnwidth, height=\columnwidth]{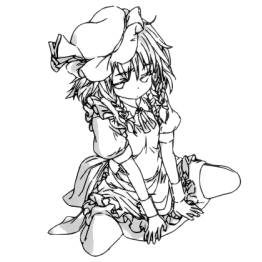}
		\subcaption*{310}
	\end{subfigure}%
	\begin{subfigure}{.32\columnwidth}
		\includegraphics[width=\columnwidth, height=\columnwidth]{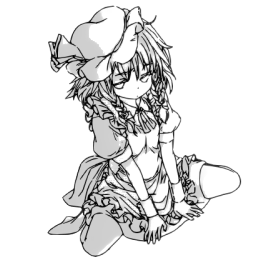}
		\subcaption*{320}
	\end{subfigure}%
	\begin{subfigure}{.32\columnwidth}
		\includegraphics[width=\columnwidth, height=\columnwidth]{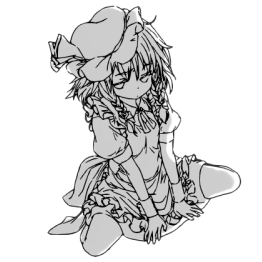}
		\subcaption*{330}
	\end{subfigure}%
	\begin{subfigure}{.32\columnwidth}
		\includegraphics[width=\columnwidth, height=\columnwidth]{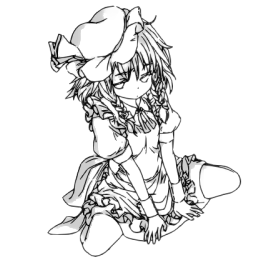}
		\subcaption*{410}
	\end{subfigure}%
	\begin{subfigure}{.32\columnwidth}
		\includegraphics[width=\columnwidth, height=\columnwidth]{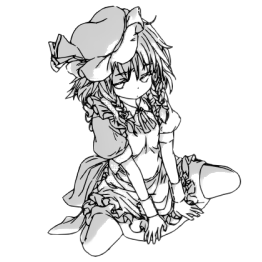}
		\subcaption*{420}
	\end{subfigure}%
	\begin{subfigure}{.32\columnwidth}
		\includegraphics[width=\columnwidth, height=\columnwidth]{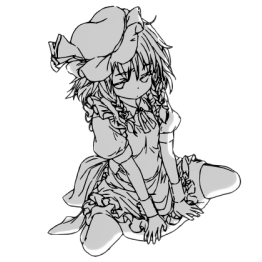}
		\subcaption*{430}
	\end{subfigure}
	
	\begin{subfigure}{.32\columnwidth}
		\includegraphics[width=\columnwidth, height=\columnwidth]{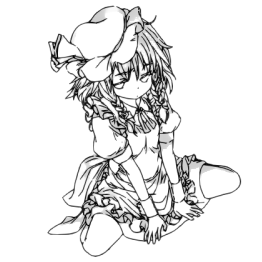}
		\subcaption*{510}
	\end{subfigure}%
	\begin{subfigure}{.32\columnwidth}
		\includegraphics[width=\columnwidth, height=\columnwidth]{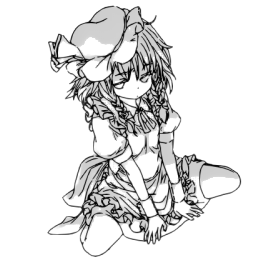}
		\subcaption*{520}
	\end{subfigure}%
	\begin{subfigure}{.32\columnwidth}
		\includegraphics[width=\columnwidth, height=\columnwidth]{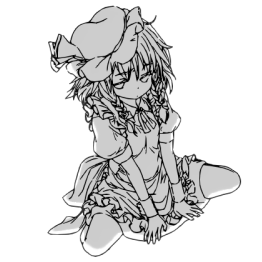}
		\subcaption*{530}
	\end{subfigure}%
	\begin{subfigure}{.32\columnwidth}
		\includegraphics[width=\columnwidth, height=\columnwidth]{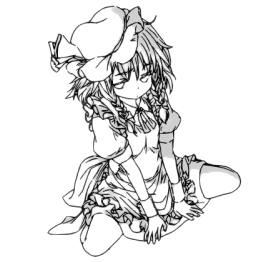}
		\subcaption*{610}
	\end{subfigure}%
	\begin{subfigure}{.32\columnwidth}
		\includegraphics[width=\columnwidth, height=\columnwidth]{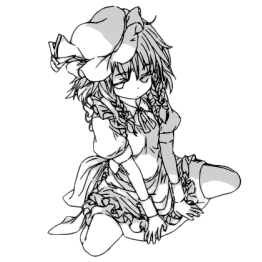}
		\subcaption*{620}
	\end{subfigure}%
	\begin{subfigure}{.32\columnwidth}
		\includegraphics[width=\columnwidth, height=\columnwidth]{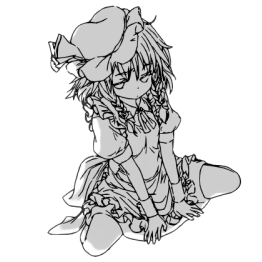}
		\subcaption*{630}
	\end{subfigure}
	
	\begin{subfigure}{.32\columnwidth}
		\includegraphics[width=\columnwidth, height=\columnwidth]{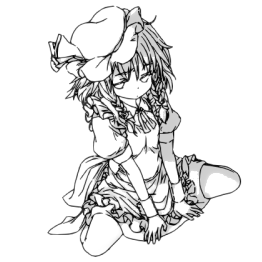}
		\subcaption*{710}
	\end{subfigure}%
	\begin{subfigure}{.32\columnwidth}
		\includegraphics[width=\columnwidth, height=\columnwidth]{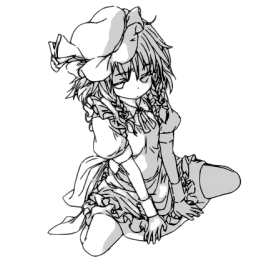}
		\subcaption*{720}
	\end{subfigure}%
	\begin{subfigure}{.32\columnwidth}
		\includegraphics[width=\columnwidth, height=\columnwidth]{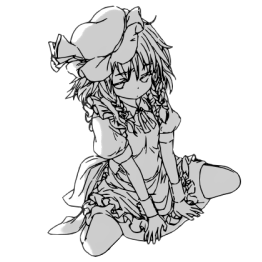}
		\subcaption*{730}
	\end{subfigure}%
	\begin{subfigure}{.32\columnwidth}
		\includegraphics[width=\columnwidth, height=\columnwidth]{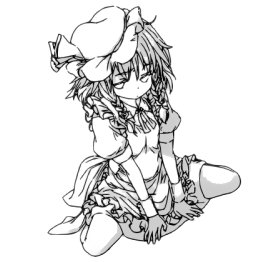}
		\subcaption*{810}
	\end{subfigure}%
	\begin{subfigure}{.32\columnwidth}
		\includegraphics[width=\columnwidth, height=\columnwidth]{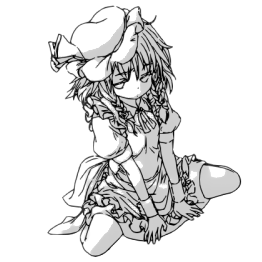}
		\subcaption*{820}
	\end{subfigure}%
	\begin{subfigure}{.32\columnwidth}
		\includegraphics[width=\columnwidth, height=\columnwidth]{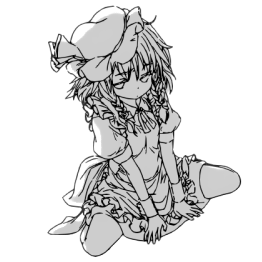}
		\subcaption*{830}
	\end{subfigure}
	
	\begin{subfigure}{.32\columnwidth}
		\includegraphics[width=\columnwidth, height=\columnwidth]{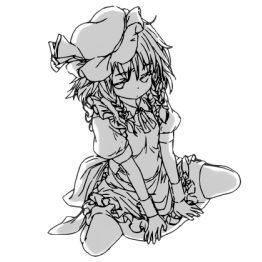}
		\subcaption*{001}
	\end{subfigure}%
	\begin{subfigure}{.32\columnwidth}
		\includegraphics[width=\columnwidth, height=\columnwidth]{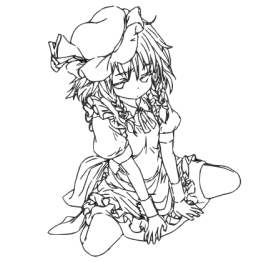}
		\subcaption*{002}
	\end{subfigure}
	
	\caption{Evaluations of our work in all 26 directions.}
	\label{fig:all2}
\end{figure*}

\begin{figure*}
	\centering
	\includegraphics[width=1.9\columnwidth]{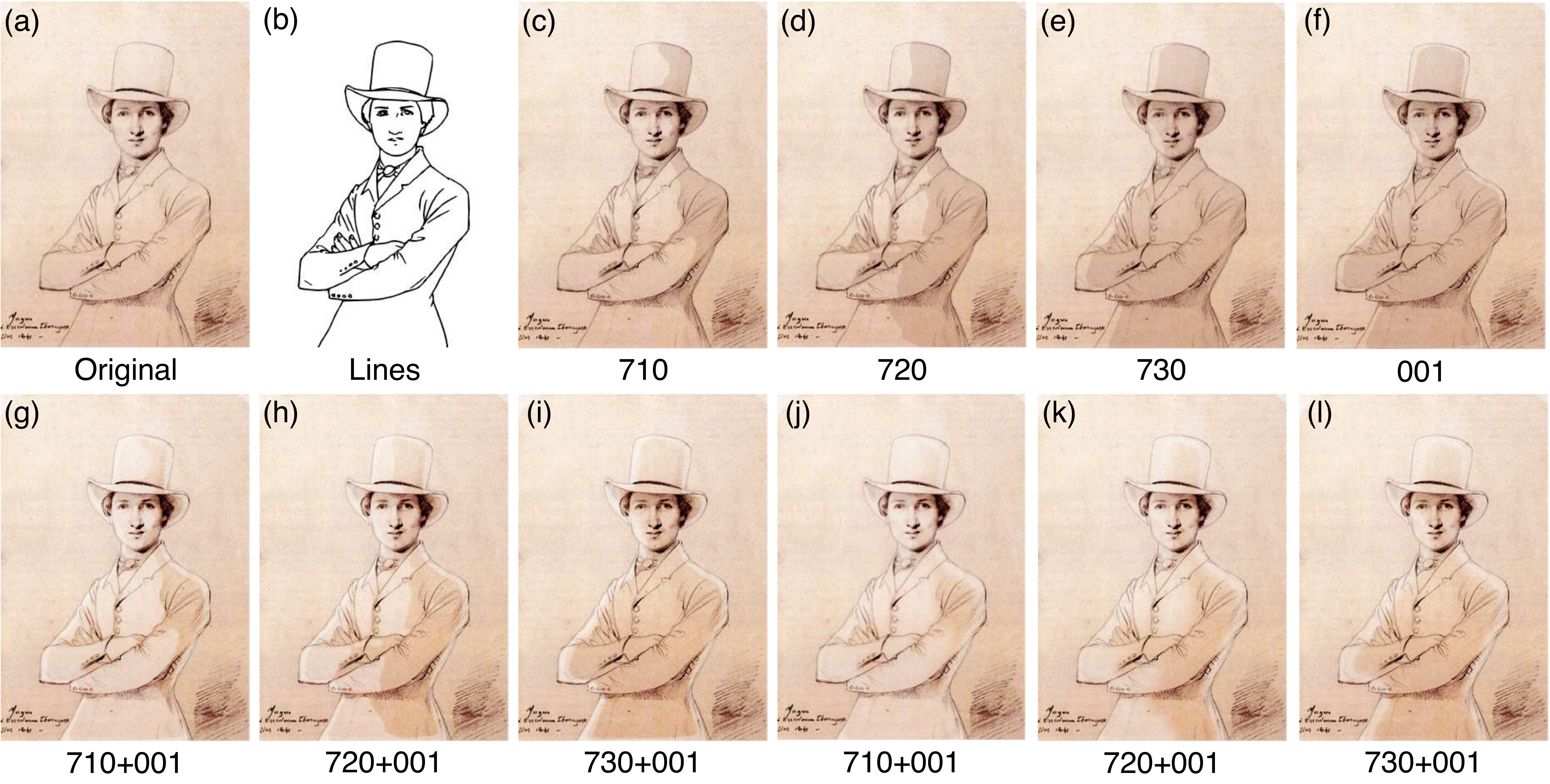}
	\caption{Examples of our shadowing system applying to artistic line drawing (Antoine Thomeguex, drawn by Jean Auguste Dominique Ingres. Public domain.). (a): original sketch. (b): extract lines from (a). (c)-(f): binary shadows in 710, 720, 730 and 001 lighting directions. (g)-(i): composites of binary shadows in dual lighting directions. (j)-(l): soft shadows in dual lighting directions. The results show that our shadowing system can give artists hints or a starting point to study shadows in different lighting sources. }
	\label{fig:art1}
\end{figure*}

\begin{figure*}
	\centering
	\begin{subfigure}{.27\columnwidth}
		\includegraphics[width=\columnwidth]{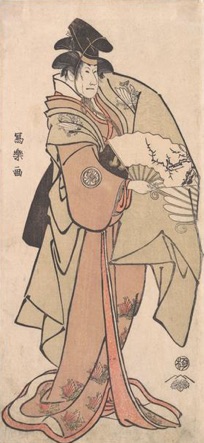}
		\subcaption{Original}
	\end{subfigure}%
	\begin{subfigure}{.27\columnwidth}
		\includegraphics[width=\columnwidth]{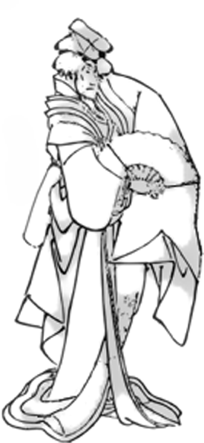}
		\subcaption{710}
	\end{subfigure}%
	\begin{subfigure}{.27\columnwidth}
		\includegraphics[width=\columnwidth]{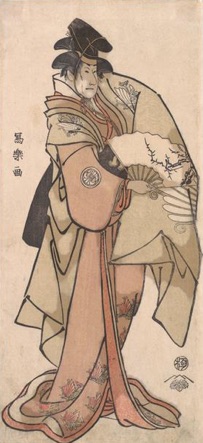}
		\subcaption{(a)+(b)}
	\end{subfigure}%
	\begin{subfigure}{.27\columnwidth}
		\includegraphics[width=\columnwidth]{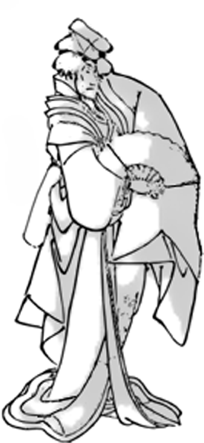}
		\subcaption{720}
	\end{subfigure}%
	\begin{subfigure}{.27\columnwidth}
		\includegraphics[width=\columnwidth]{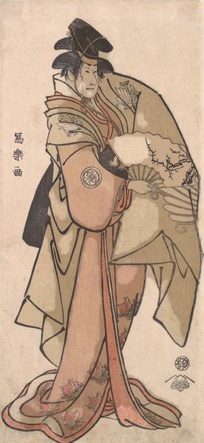}
		\subcaption{(a)+(d)}
	\end{subfigure}%
	\begin{subfigure}{.27\columnwidth}
		\includegraphics[width=\columnwidth]{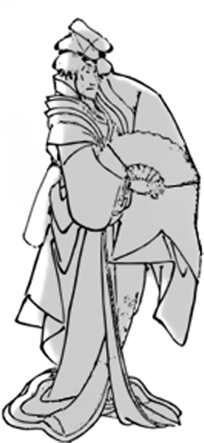}
		\subcaption{730}
	\end{subfigure}%
	\begin{subfigure}{.27\columnwidth}
		\includegraphics[width=\columnwidth]{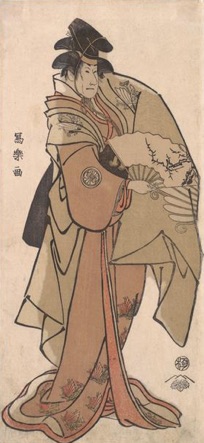}
		\subcaption{(a)+(f)}
	\end{subfigure}
	
	\begin{subfigure}{.27\columnwidth}
		\includegraphics[width=\columnwidth]{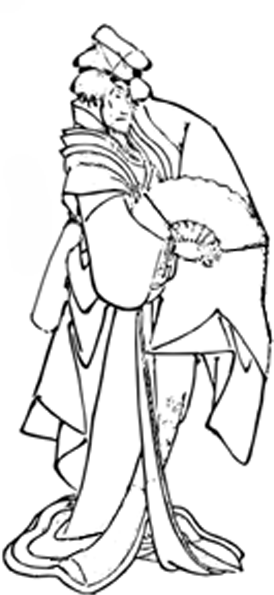}
		\subcaption{Lines}
	\end{subfigure}%
	\begin{subfigure}{.27\columnwidth}
		\includegraphics[width=\columnwidth]{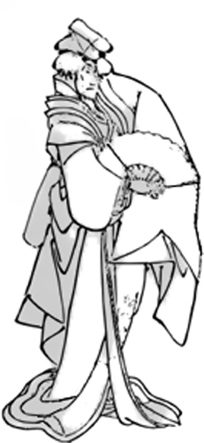}
		\subcaption{310}
	\end{subfigure}%
	\begin{subfigure}{.27\columnwidth}
		\includegraphics[width=\columnwidth]{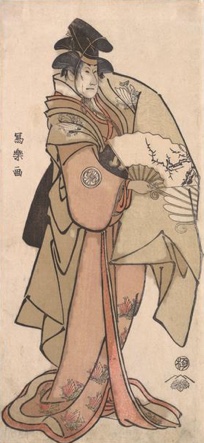}
		\subcaption{(a)+(h)}
	\end{subfigure}%
	\begin{subfigure}{.27\columnwidth}
		\includegraphics[width=\columnwidth]{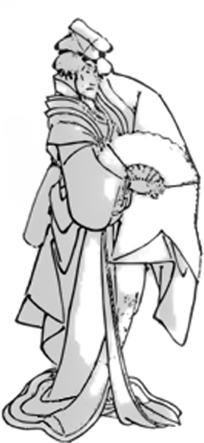}
		\subcaption{320}
	\end{subfigure}%
	\begin{subfigure}{.27\columnwidth}
		\includegraphics[width=\columnwidth]{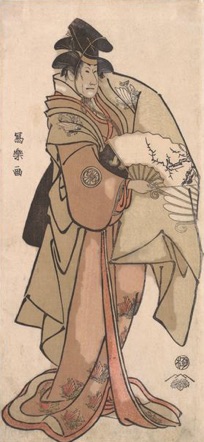}
		\subcaption{(a)+(j)}
	\end{subfigure}%
	\begin{subfigure}{.27\columnwidth}
		\includegraphics[width=\columnwidth]{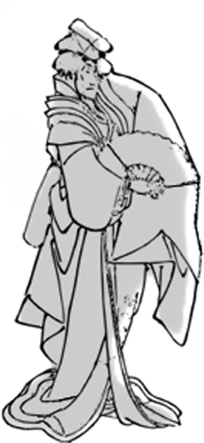}
		\subcaption{330}
	\end{subfigure}%
	\begin{subfigure}{.27\columnwidth}
		\includegraphics[width=\columnwidth]{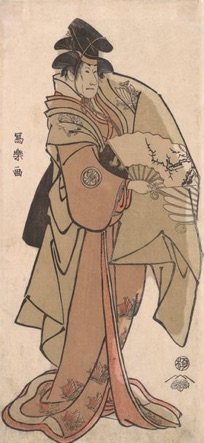}
		\subcaption{(a)+(l)}
	\end{subfigure}
	
	\caption{Examples of our shadowing system applying to Ukiyo-e (Kabuki Actor Segawa Kikunojō III as the Shirabyōshi Hisakata Disguised as Yamato Manzai, by Toshusai Sharaku. Public domain.). Composite our shadows with pure colorized artwork.}
	\label{fig:art2}
\end{figure*}

\begin{figure*}
	\centering
	\begin{subfigure}{.3\columnwidth}
		\includegraphics[width=\columnwidth]{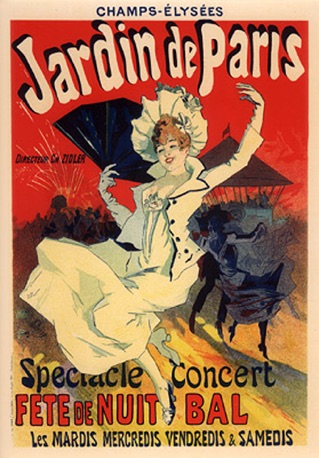}
		\subcaption{Original}
	\end{subfigure}%
	\begin{subfigure}{.3\columnwidth}
		\includegraphics[width=\columnwidth]{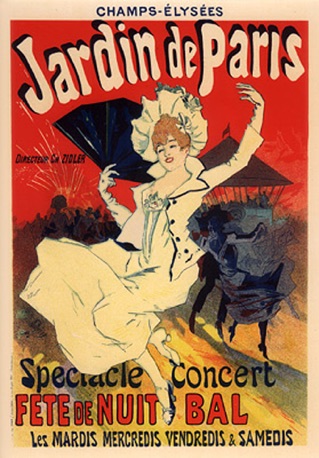}
		\subcaption{w/o shadows}
	\end{subfigure}%
	\begin{subfigure}{.3\columnwidth}
		\includegraphics[width=\columnwidth]{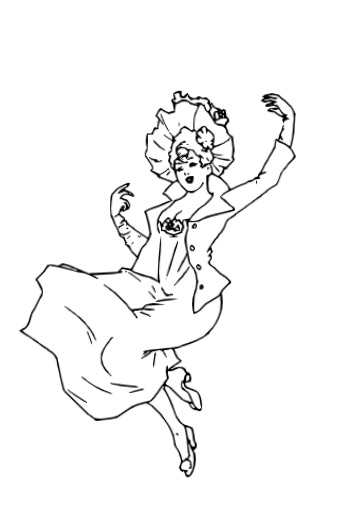}
		\subcaption{Lines}
	\end{subfigure}%
	\begin{subfigure}{.3\columnwidth}
		\includegraphics[width=\columnwidth]{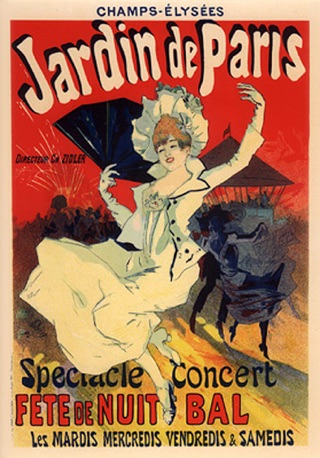}
		\subcaption{510}
	\end{subfigure}%
	\begin{subfigure}{.3\columnwidth}
		\includegraphics[width=\columnwidth]{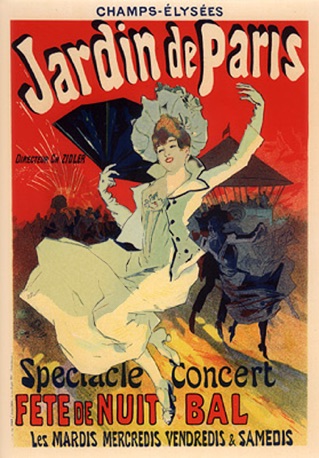}
		\subcaption{520}
	\end{subfigure}%
	\begin{subfigure}{.3\columnwidth}
		\includegraphics[width=\columnwidth]{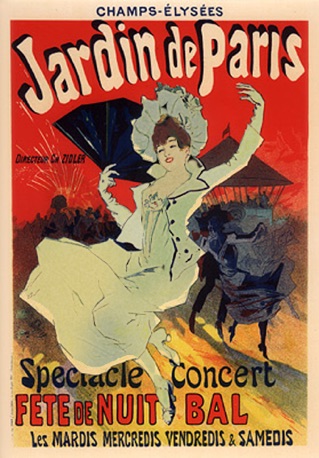}
		\subcaption{530}
	\end{subfigure}
	
	\begin{subfigure}{.3\columnwidth}
		\includegraphics[width=\columnwidth]{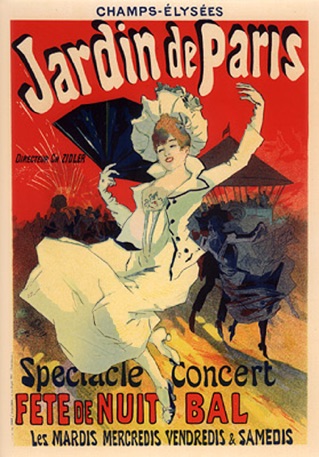}
		\subcaption{210}
	\end{subfigure}%
	\begin{subfigure}{.3\columnwidth}
		\includegraphics[width=\columnwidth]{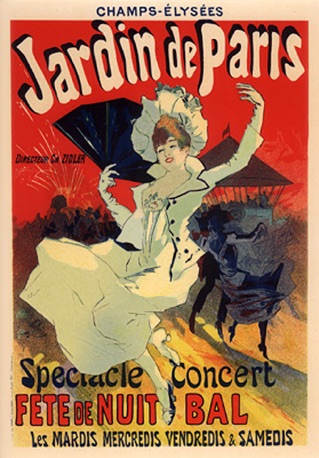}
		\subcaption{220}
	\end{subfigure}%
	\begin{subfigure}{.3\columnwidth}
		\includegraphics[width=\columnwidth]{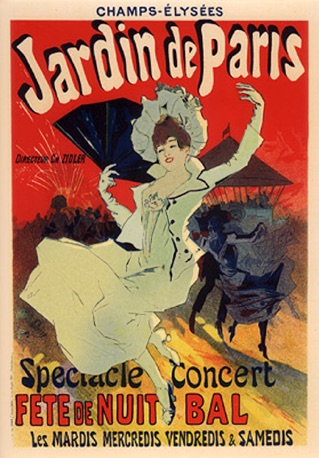}
		\subcaption{230}
	\end{subfigure}%
	\begin{subfigure}{.3\columnwidth}
		\includegraphics[width=\columnwidth]{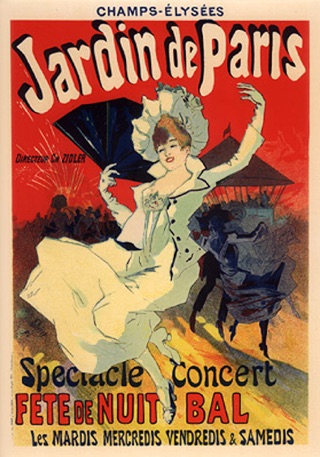}
		\subcaption{810}
	\end{subfigure}%
	\begin{subfigure}{.3\columnwidth}
		\includegraphics[width=\columnwidth]{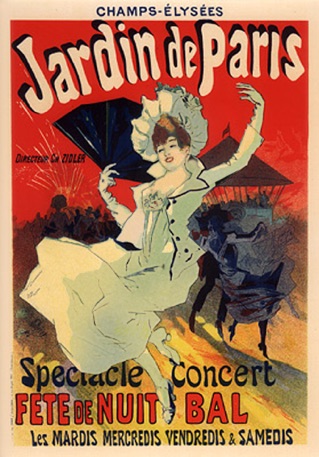}
		\subcaption{820}
	\end{subfigure}%
	\begin{subfigure}{.3\columnwidth}
		\includegraphics[width=\columnwidth]{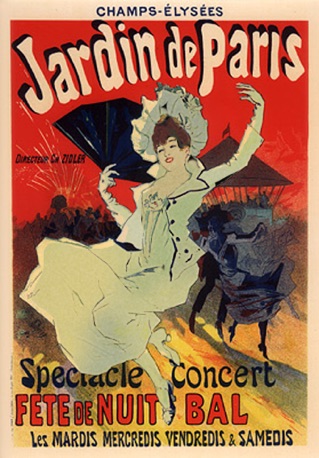}
		\subcaption{830}
	\end{subfigure}
	
	\begin{subfigure}{.3\columnwidth}
		\includegraphics[width=\columnwidth]{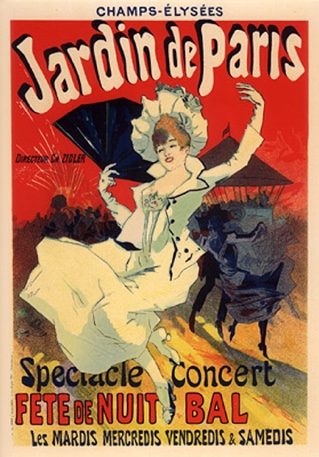}
		\subcaption{110}
	\end{subfigure}%
	\begin{subfigure}{.3\columnwidth}
		\includegraphics[width=\columnwidth]{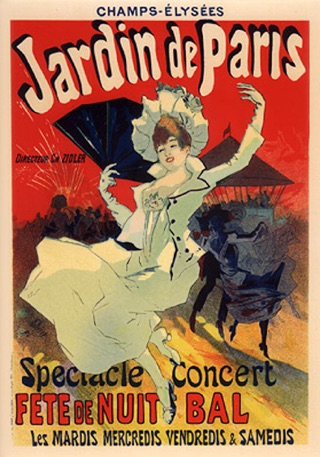}
		\subcaption{120}
	\end{subfigure}%
	\begin{subfigure}{.3\columnwidth}
		\includegraphics[width=\columnwidth]{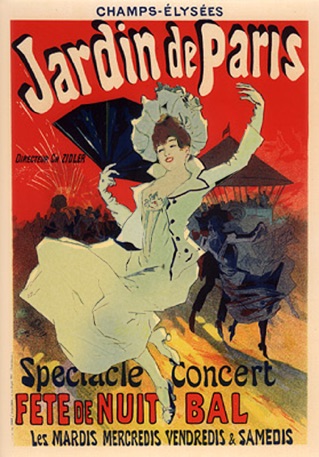}
		\subcaption{130}
	\end{subfigure}%
	\begin{subfigure}{.3\columnwidth}
		\includegraphics[width=\columnwidth]{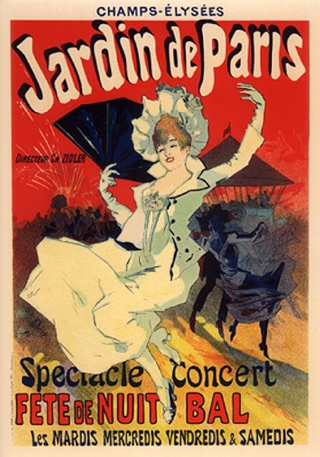}
		\subcaption{710}
	\end{subfigure}%
	\begin{subfigure}{.3\columnwidth}
		\includegraphics[width=\columnwidth]{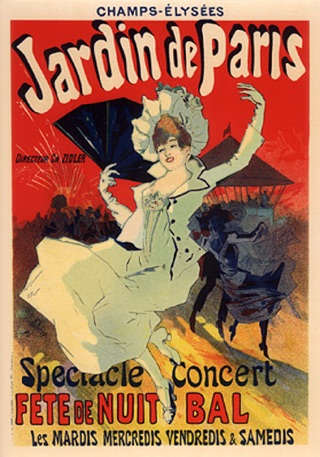}
		\subcaption{720}
	\end{subfigure}%
	\begin{subfigure}{.3\columnwidth}
		\includegraphics[width=\columnwidth]{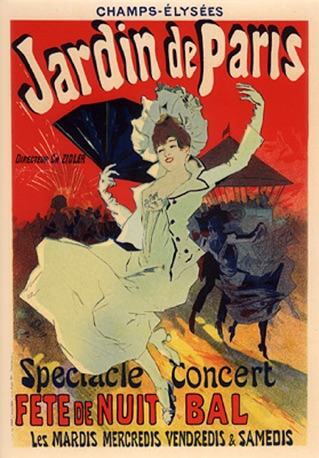}
		\subcaption{730}
	\end{subfigure}
	
	\caption{Examples of our shadowing system applying to poster (Jardin de Paris, Fête de Nuit Bal, illustrated by Jules Cheret. Public domain.). (a): original poster. (b): remove the shadows in the human. (c): extract line drawing from (a). (d)-(r): composites our shadows in various lighting directions with (b). Assuming the artists draw artwork with digital tools, they can rapidly try different shadows with our shadowing system.}
	\label{fig:art3}
\end{figure*}

\begin{figure*}
	\centering
	\begin{subfigure}{.4\columnwidth}
		\includegraphics[width=\columnwidth, height=\columnwidth]{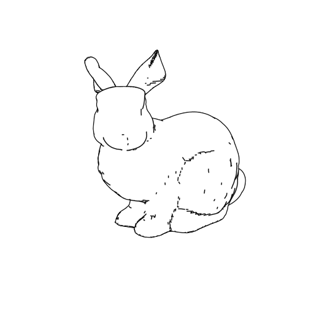}
		\subcaption{line drawing (w/o p.)}
	\end{subfigure}%
	\begin{subfigure}{.4\columnwidth}
		\includegraphics[width=\columnwidth, height=\columnwidth]{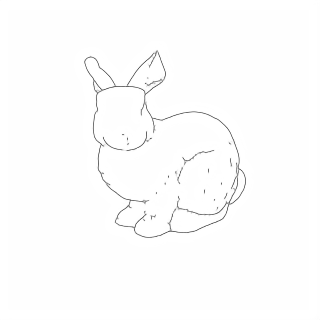}
		\subcaption{(a) (w/ p.)}
	\end{subfigure}%
	\begin{subfigure}{.4\columnwidth}
		\includegraphics[width=\columnwidth, height=\columnwidth]{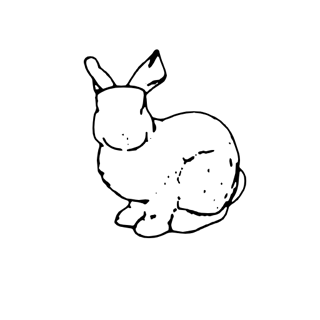}
		\subcaption{stylized (a) (w/o p.)}
	\end{subfigure}%
	\begin{subfigure}{.4\columnwidth}
		\includegraphics[width=\columnwidth, height=\columnwidth]{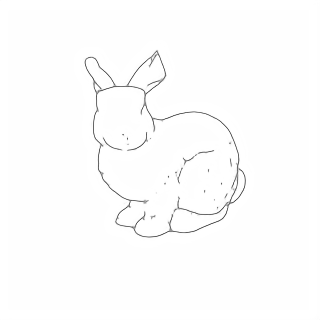}
		\subcaption{(c) (w/ p.)}
	\end{subfigure}
	
	\begin{subfigure}{.4\columnwidth}
		\includegraphics[width=\columnwidth, height=\columnwidth]{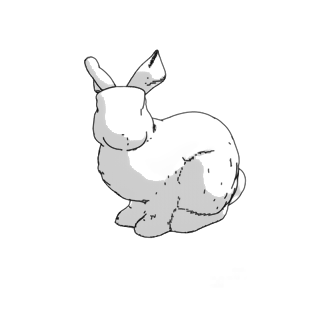}
		\subcaption{shadow from (a)}
	\end{subfigure}%
	\begin{subfigure}{.4\columnwidth}
		\includegraphics[width=\columnwidth, height=\columnwidth]{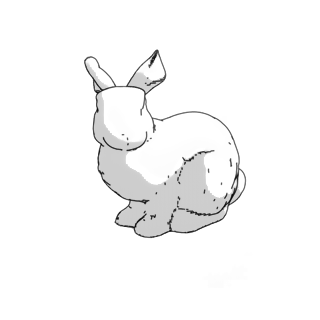}
		\subcaption{shadow from (b)}
	\end{subfigure}%
	\begin{subfigure}{.4\columnwidth}
		\includegraphics[width=\columnwidth, height=\columnwidth]{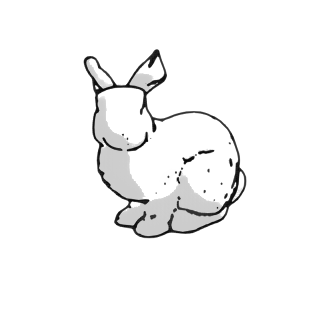}
		\subcaption{shadow from (c)}
	\end{subfigure}%
	\begin{subfigure}{.4\columnwidth}
		\includegraphics[width=\columnwidth, height=\columnwidth]{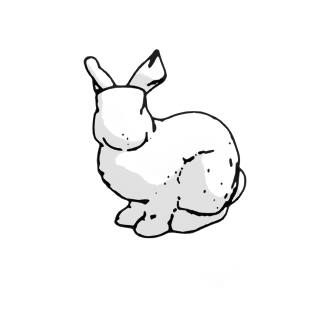}
		\subcaption{shadow from (d)}
	\end{subfigure}
	
	\begin{subfigure}{.4\columnwidth}
		\includegraphics[width=\columnwidth, height=\columnwidth]{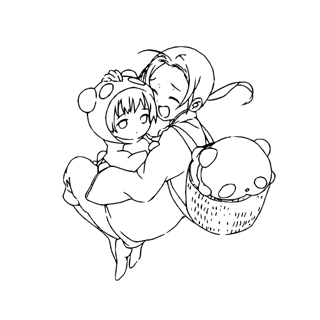}
		\subcaption{line drawing (w/o p.)}
	\end{subfigure}%
	\begin{subfigure}{.4\columnwidth}
		\includegraphics[width=\columnwidth, height=\columnwidth]{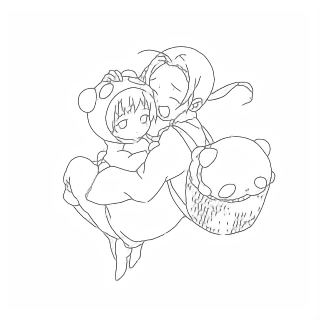}
		\subcaption{(i) (w/ p.)}
	\end{subfigure}%
	\begin{subfigure}{.4\columnwidth}
		\includegraphics[width=\columnwidth, height=\columnwidth]{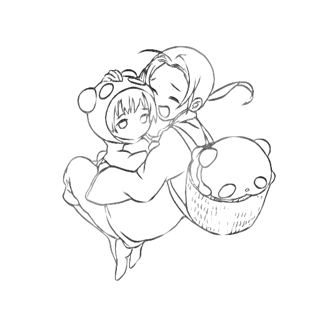}
		\subcaption{stylized (i) (w/o p.)}
	\end{subfigure}%
	\begin{subfigure}{.4\columnwidth}
		\includegraphics[width=\columnwidth, height=\columnwidth]{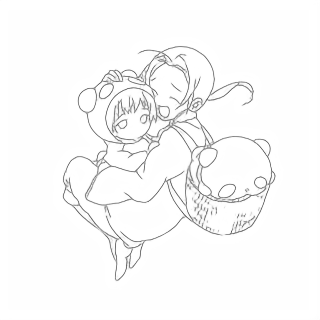}
		\subcaption{(k) (w/ p.)}
	\end{subfigure}
	
	\begin{subfigure}{.4\columnwidth}
		\includegraphics[width=\columnwidth, height=\columnwidth]{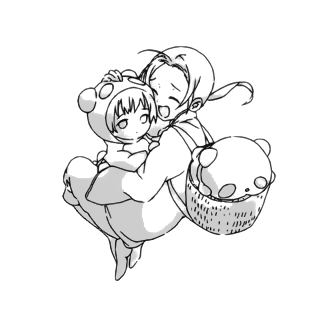}
		\subcaption{shadow from (i)}
	\end{subfigure}%
	\begin{subfigure}{.4\columnwidth}
		\includegraphics[width=\columnwidth, height=\columnwidth]{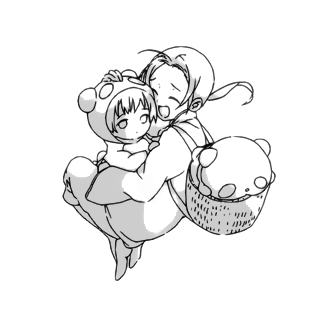}
		\subcaption{shadow from (j)}
	\end{subfigure}%
	\begin{subfigure}{.4\columnwidth}
		\includegraphics[width=\columnwidth, height=\columnwidth]{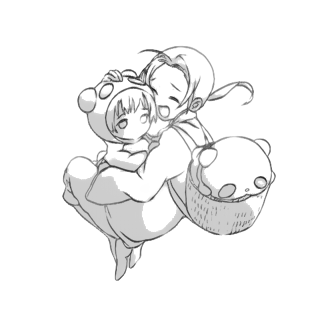}
		\subcaption{shadow from (k)}
	\end{subfigure}%
	\begin{subfigure}{.4\columnwidth}
		\includegraphics[width=\columnwidth, height=\columnwidth]{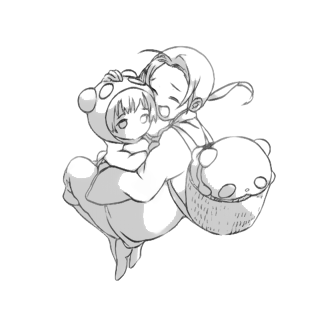}
		\subcaption{shadow from (l)}
	\end{subfigure}
	
	\caption{The comparisons of our shadowing system with and without pre-processing (denoted as w/ p. and w/o p.). (a), (c), (i), (k) are line drawings without our pre-processing. (b), (d), (j), (l) are line drawings after our pre-processing. We test the robustness of our pre-processing system with stylized lines (c) and (k) which have different line width, line transparency, and line strokes.}
	\label{fig:norm}
\end{figure*}

\begin{figure*}[h]
	\centering
	\begin{subfigure}{.33\columnwidth}
		\includegraphics[width=\columnwidth, height=\columnwidth]{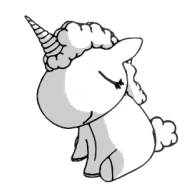}
		\subcaption*{310}
	\end{subfigure}%
	\begin{subfigure}{.33\columnwidth}
		\includegraphics[width=\columnwidth, height=\columnwidth]{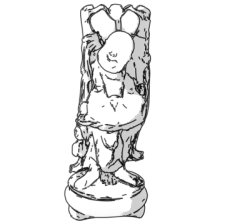}
		\subcaption*{810}
	\end{subfigure}%
	\begin{subfigure}{.33\columnwidth}
		\includegraphics[width=\columnwidth, height=\columnwidth]{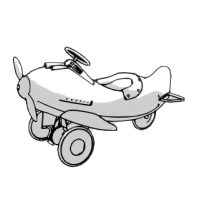}
		\subcaption*{220}
	\end{subfigure}%
	\begin{subfigure}{.33\columnwidth}
		\includegraphics[width=\columnwidth, height=\columnwidth]{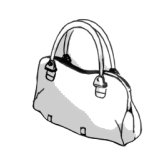}
		\subcaption*{210}
	\end{subfigure}%
	\begin{subfigure}{.33\columnwidth}
		\includegraphics[width=\columnwidth, height=\columnwidth]{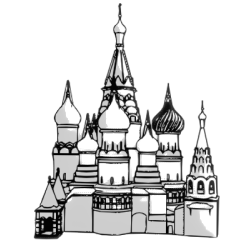}
		\subcaption*{220}
	\end{subfigure}%
	\begin{subfigure}{.33\columnwidth}
		\includegraphics[width=\columnwidth, height=\columnwidth]{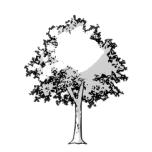}
		\subcaption*{820}
	\end{subfigure}
	
	\begin{subfigure}{.33\columnwidth}
		\includegraphics[width=\columnwidth, height=\columnwidth]{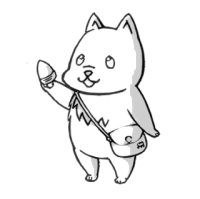}
		\subcaption*{710}
	\end{subfigure}%
	\begin{subfigure}{.33\columnwidth}
		\includegraphics[width=\columnwidth, height=\columnwidth]{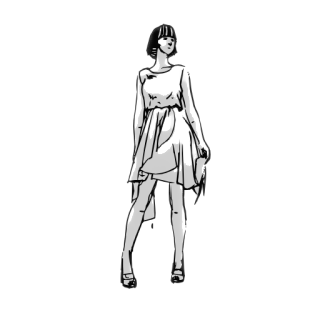}
		\subcaption*{120}
	\end{subfigure}%
	\begin{subfigure}{.33\columnwidth}
		\includegraphics[width=\columnwidth, height=\columnwidth]{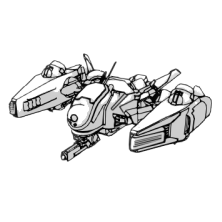}
		\subcaption*{820}
	\end{subfigure}%
	\begin{subfigure}{.33\columnwidth}
		\includegraphics[width=\columnwidth, height=\columnwidth]{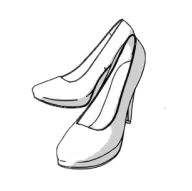}
		\subcaption*{810}
	\end{subfigure}%
	\begin{subfigure}{.33\columnwidth}
		\includegraphics[width=\columnwidth, height=\columnwidth]{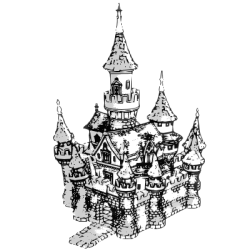}
		\subcaption*{320}
	\end{subfigure}%
	\begin{subfigure}{.33\columnwidth}
		\includegraphics[width=\columnwidth, height=\columnwidth]{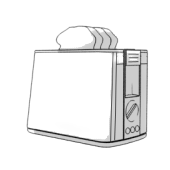}
		\subcaption*{810}
	\end{subfigure}

	\caption{Evaluations on various categories of sketch (e.g. sculpture, bags, shoes, toys, sketchy cloth, buildings and etc.). This demonstrates that our work has generalization ability.}
	\label{fig:generalization}
\end{figure*}

\begin{figure*}[h]
	\centering
	\begin{subfigure}{.33\columnwidth}
		\includegraphics[width=\columnwidth, height=\columnwidth]{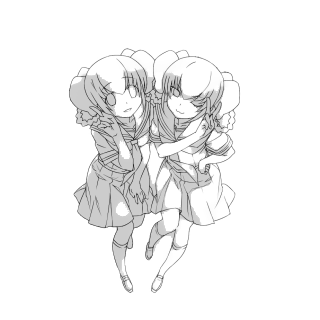}
		\subcaption{size-$320^{2}$}
	\end{subfigure}%
	\begin{subfigure}{.33\columnwidth}
		\includegraphics[width=\columnwidth, height=\columnwidth]{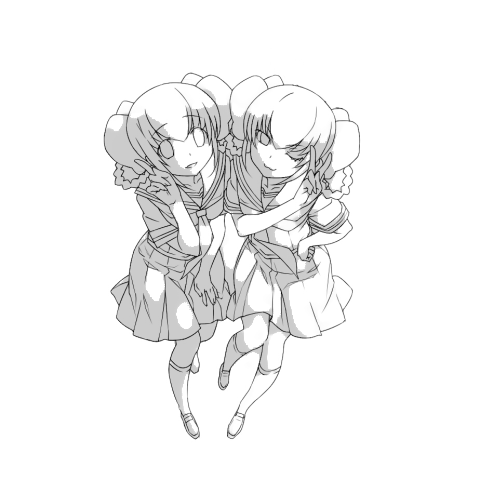}
		\subcaption{size-$480^{2}$}
	\end{subfigure}%
	\begin{subfigure}{.33\columnwidth}
		\includegraphics[width=\columnwidth, height=\columnwidth]{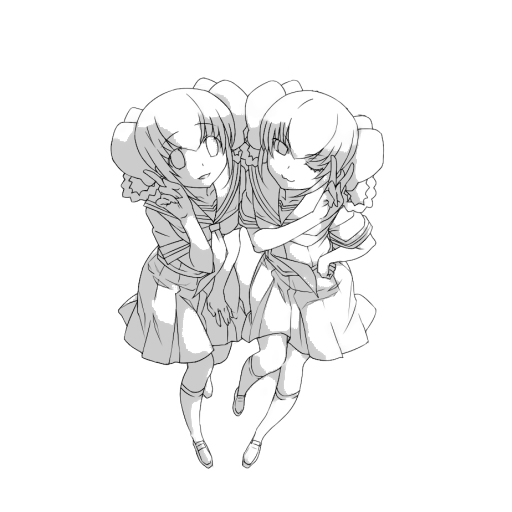}
		\subcaption{size-$512^{2}$}
	\end{subfigure}%
	\begin{subfigure}{.33\columnwidth}
		\includegraphics[width=\columnwidth, height=\columnwidth]{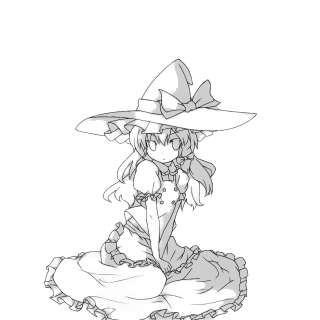}
		\subcaption{size-$320^{2}$}
	\end{subfigure}%
	\begin{subfigure}{.33\columnwidth}
		\includegraphics[width=\columnwidth, height=\columnwidth]{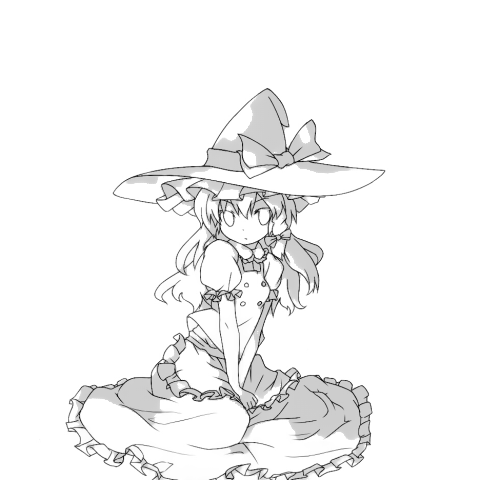}
		\subcaption{size-$480^{2}$}
	\end{subfigure}%
	\begin{subfigure}{.33\columnwidth}
		\includegraphics[width=\columnwidth, height=\columnwidth]{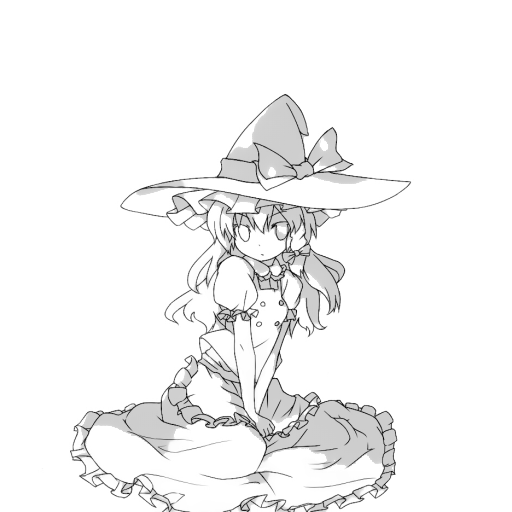}
		\subcaption{size-$512^{2}$}
	\end{subfigure}
	
	\begin{subfigure}{.33\columnwidth}
		\includegraphics[width=0.8\columnwidth, height=0.8\columnwidth]{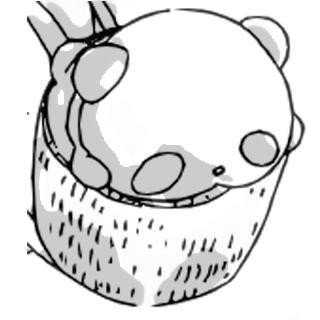}
		\subcaption{210}
	\end{subfigure}%
	\begin{subfigure}{.33\columnwidth}
		\includegraphics[width=0.8\columnwidth, height=0.8\columnwidth]{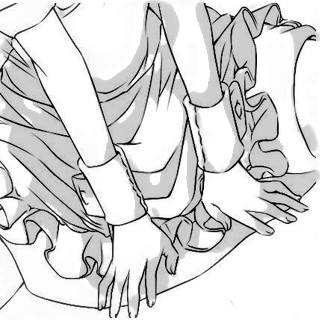}
		\subcaption{710}
	\end{subfigure}%
	\begin{subfigure}{.33\columnwidth}
		\includegraphics[width=0.8\columnwidth, height=0.8\columnwidth]{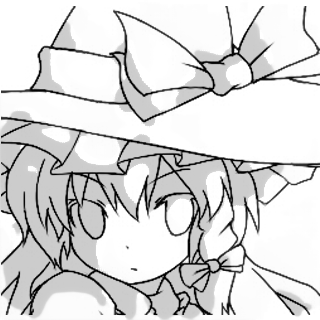}
		\subcaption{310}
	\end{subfigure}%
	\begin{subfigure}{.33\columnwidth}
		\includegraphics[width=0.8\columnwidth, height=0.8\columnwidth]{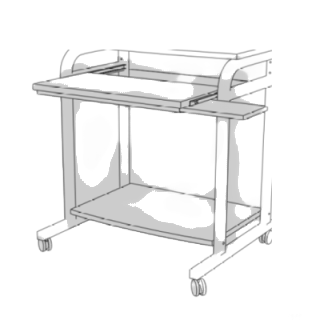}
		\subcaption{210}
	\end{subfigure}%
	\begin{subfigure}{.33\columnwidth}
		\includegraphics[width=0.8\columnwidth, height=0.8\columnwidth]{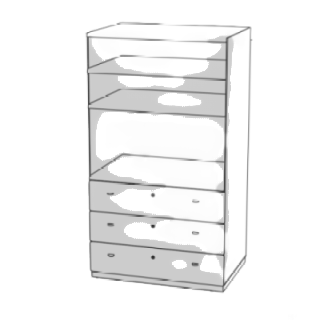}
		\subcaption{210}
	\end{subfigure}%
	\begin{subfigure}{.33\columnwidth}
		\includegraphics[width=0.8\columnwidth, height=0.8\columnwidth]{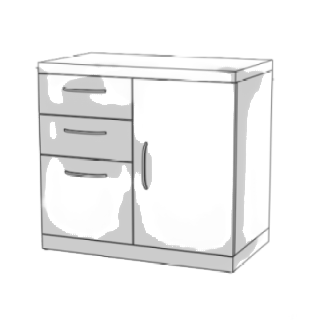}
		\subcaption{210}
	\end{subfigure}
	
	\caption{Limitation examples of the Future Work section. (a)-(c) and (d)-(f): invariant performance of shadows in different input size under the same lighting direction. (g)-(i): results of the local parts of line drawings being inputted. (j)-(l): unrealistic shadows in complex hard surface object.}
	\label{fig:failure_cases}
\end{figure*}

Figure~\ref{fig:compare1_normal}, ~\ref{fig:compare2_normal}, ~\ref{fig:depthcompare2}, ~\ref{fig:cat}, ~\ref{human}, ~\ref{fig:nips}, ~\ref{fig:normal} : more comparisons with related work.
Figure~\ref{contin}, ~\ref{fig:all1}, ~\ref{fig:all2} : our results in more lighting directions.
Figure~\ref{fig:art1}, ~\ref{fig:art2}, ~\ref{fig:art3} : examples of our shadowing system applied to artistic line drawings.
Figure~\ref{fig:norm} : our results with and without pre-processing, and the robustness of our results in the wild.
Figure~\ref{fig:generalization}: generalization ability. Figure~\ref{fig:failure_cases} : Failure cases.

Table~\ref{table:userstudy1} and ~\ref{table:userstudy2} shows the statistically significance report of our user study in Likert scores. We deploy levene's test for the equality of variance, and Fisher's Least Significant Difference (LSD) for the analysis of variance.

\begin{table}[!htb]
	\begin{center}
		\begin{tabular}{l c c c c c c}
			\toprule
			& GT & Our & \cite{hudon2018deep} & \cite{su2018interactive} & pix2pix & U-net\\
			\midrule
			\textbf{Mean} & 6.37 & 6.70 & 5.78 & 3.35 & 3.78 & 3.06\\
			\textbf{SD} & 1.55 & 1.33 & 1.27 & 1.30 & 1.31 & 2.11\\
			\midrule
			\textbf{Mean} & 6.50 & 6.66 & 5.69 & 3.44 & 3.91 & 3.03\\
			\textbf{SD} & 1.40 & 1.15 & 1.02 & 1.12 & 1.10 & 2.10\\
			\bottomrule
		\end{tabular}
	\end{center}
	\caption{Mean and standard deviation of user study (on Likert score). Top: total results. Bottom: results of people with drawing experience.}
	\label{table:userstudy1}
\end{table}

\begin{table}[!htb]
	\begin{center}
		\begin{tabular}{l c c c c c}
			\toprule
			& \textbf{MeanDiff} & \textbf{T-value} & \textbf{P-value} & \textbf{Alpha}\\
			\midrule
			Our / GT                                         & 0.33 & 1.18 & 0.24 & 0.05\\
			\cite{hudon2018deep} / GT  & -0.59 & -2.16 & 0.03 & 0.05\\
			\cite{hudon2018deep} / Our & -0.92 & -3.34 & $9.23E^{-4}$ & 0.05\\
			\cite{su2018interactive} / GT & -3.02 & -10.98 & $2.41E^{-24}$ & 0.05\\
			\cite{su2018interactive} / Our & -3.35 & -12.16 & $1.11E^{-28}$ & 0.05\\
			\cite{su2018interactive} / \cite{hudon2018deep} & -2.43 & -8.82 & $5.21E^{-17}$ & 0.05\\
			pix2pix / GT & -2.60 & -9.45 & $4.83E^{-19}$ & 0.05\\
			pix2pix / Our & -2.93 & -10.63 & $4.36E^{-23}$ & 0.05\\
			pix2pix / \cite{hudon2018deep} & -2.00 & -7.29 & $2.07E^{-12}$ & 0.05\\
			pix2pix / \cite{su2018interactive} & 0.42 & 1.53 & 0.126 & 0.05\\
			U-net / GT & -3.31 & -12.02 & $3.76E^{-28}$ & 0.05\\
			U-net / Our & -3.63 & -13.20 & $1.22E^{-32}$ & 0.05\\
			U-net / \cite{hudon2018deep}   & -2.71 & -9.87 & $1.98E^{-20}$ & 0.05\\
			U-net / \cite{su2018interactive} & -0.29 & -1.04 & 0.30 & 0.05\\
			U-net/pix2pix & -0.71 & -2.57 & 0.01 & 0.05\\
			\midrule
			Our / GT & 0.16 & 0.52 & 0.60 & 0.05\\
			\cite{hudon2018deep} / GT  & -0.82 & -2.67 & $8.07E^{-3}$ & 0.05\\
			\cite{hudon2018deep} / Our & -0.98 & -3.19 & $1.62E^{-3}$ & 0.05\\
			\cite{su2018interactive} / GT & -3.06 & -10.02 & $6.79E^{-20}$ & 0.05\\
			\cite{su2018interactive} / Our & -3.22 & -10.54 & $1.67E^{-21}$ & 0.05\\
			\cite{su2018interactive} / \cite{hudon2018deep} & -2.25 & -7.34 & $3.33E^{-12}$ & 0.05\\
			pix2pix / GT & -2.60 & -8.49 & $2.35E^{-15}$ & 0.05\\
			pix2pix / Our & -2.75 & -9.01 & $7.39E^{-17}$ & 0.05\\
			pix2pix / \cite{hudon2018deep} & -1.78 & -5.82 & $1.91E^{-8}$ & 0.05\\
			pix2pix / \cite{su2018interactive} & 0.47 & 1.53 & 0.128 & 0.05\\
			U-net / GT & -3.48 & -11.37 & $3.87E^{-24}$ & 0.05\\
			U-net / Our & -3.63 & -11.89 & $8.28E^{-26}$ & 0.05\\
			U-net / \cite{hudon2018deep}   & -2.65 & -8.70 & $6.06E^{-16}$ & 0.05\\
			U-net / \cite{su2018interactive} & -0.41 & -1.35 & 0.178 & 0.05\\
			U-net/pix2pix & -0.87 & -2.88 & $4.40E^{-3}$ & 0.05\\
			\bottomrule
		\end{tabular}
	\end{center}
	\caption{Statistical significance report of user study (on Likert score). Top: total results. Bottom: results of people with drawing experience.}
	\label{table:userstudy2}
\end{table}

\section{Dataset Samples}
\begin{figure*}
	\centering
	\includegraphics[width=2\columnwidth]{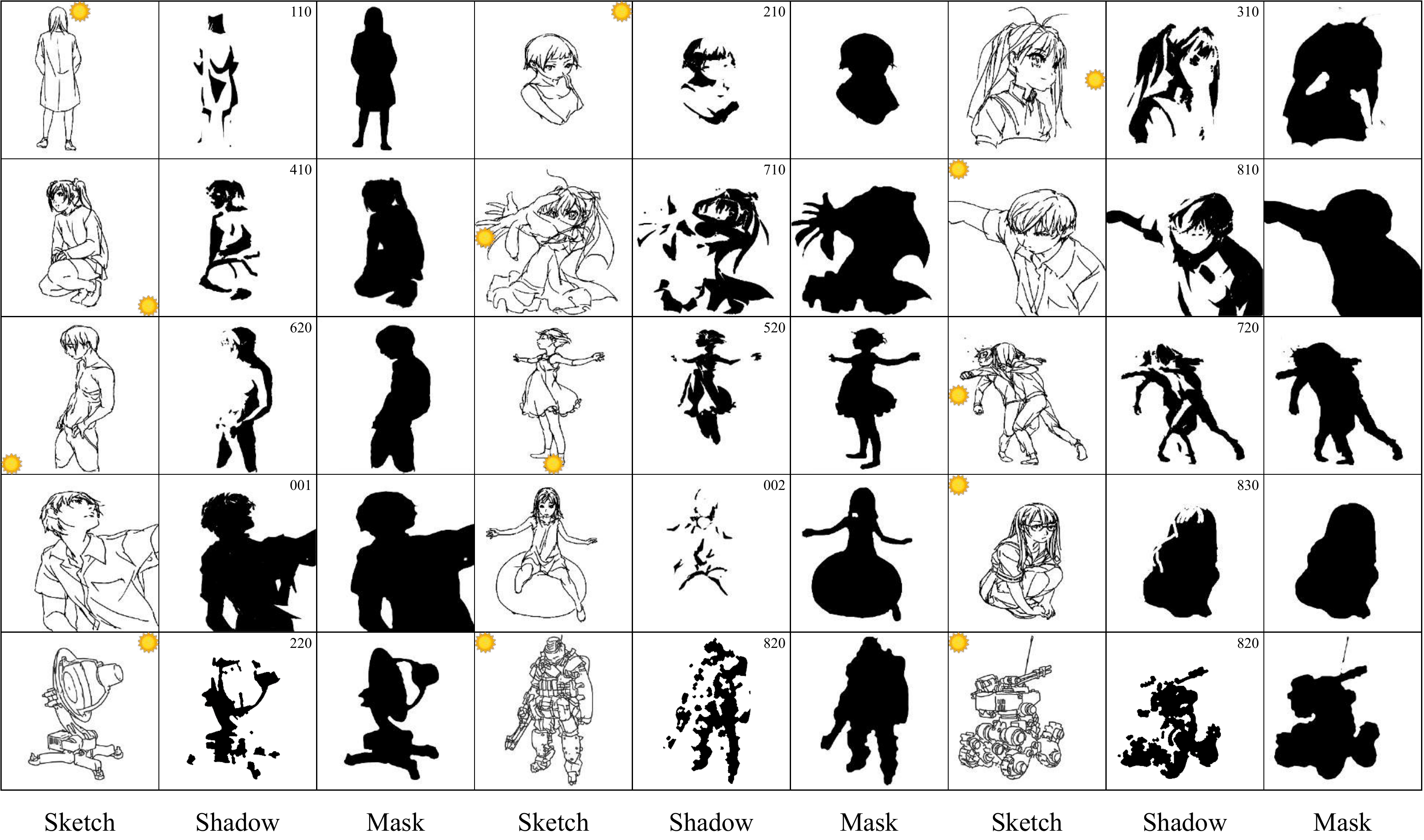}
	
	\caption{Sketch/shadow/mask pairs from our dataset. Our dataset contains alpha masks for the line drawings, but we did not need to use these masks in this paper.}
	\label{fig:dataset}
\end{figure*}

Figure~\ref{fig:dataset} shows \{sketch, light direction, shadow, mask\} sample pairs from our dataset. Our dataset comprises 1,160 sketch/shadow pairs and includes a variety of lighting directions and subjects. Specifically, 372 front-lighting, 506 side-lighting, 111 back-lighting, 85 center-back, and 86 center-front. With regard to subjects there are 867 single-person, 56 multi-person, 177 body-part, and 60 mecha.

\end{document}